\documentclass[10pt,twocolumn,letterpaper]{article}

\usepackage[pagenumbers]{cvpr} 

\usepackage[dvipsnames]{xcolor}

\definecolor{cvprblue}{rgb}{0.21,0.49,0.74}
\usepackage[pagebackref,breaklinks,colorlinks,citecolor=cvprblue]{hyperref}
\usepackage{multirow}
\usepackage{tikz}
\usepackage{siunitx}
\usepackage{color, colortbl}
\usepackage{float}
\usetikzlibrary{shapes.arrows}

\def\paperID{8457} 
\def\confName{CVPR}
\def\confYear{2024}

\title{Compressed 3D Gaussian Splatting for Accelerated Novel View Synthesis}

\author{Simon Niedermayr\\
{\tt\small simon.niedermayr@tum.de}
\and
Josef Stumpfegger
\\
{\tt\small ga87tux@mytum.de}
\and
Rüdiger Westermann
\\
{\tt\small westermann@tum.de} \\
\and
{Technical University of Munich} 
}

\begin{document}
\maketitle
\begin{abstract}
Recently, high-fidelity scene reconstruction with an optimized 3D Gaussian splat representation has been introduced for novel view synthesis from sparse image sets. 
Making such representations suitable for applications like network streaming and rendering on low-power devices requires significantly reduced memory consumption as well as improved rendering efficiency.
We propose a compressed 3D Gaussian splat representation that utilizes sensitivity-aware vector clustering with quantization-aware training to compress directional colors and Gaussian parameters. The learned codebooks have low bitrates and achieve a compression rate of up to $31\times$ on real-world scenes with only minimal degradation of visual quality.
We demonstrate that the compressed splat representation can be efficiently rendered with hardware rasterization on lightweight GPUs at up to $4\times$ higher framerates than reported via an optimized GPU compute pipeline.
Extensive experiments across multiple datasets demonstrate the robustness and rendering speed of the proposed approach. 

\end{abstract}    

\section{Introduction}
\label{sec:intro}

Novel view synthesis 
aims to generate new views
of a 3D scene or object by interpolating from a sparse set of images with known camera parameters. 
NeRF \cite{mildenhall_nerf_2021} and its variants have proposed the use of direct volume rendering to learn a volumetric radiance field from which novel views can be rendered. However, expensive neural network evaluations 
prohibit efficient training and rendering.
Recent research utilizes explicit scene representations such as voxel-based \cite{sun_direct_2022} or point-based structures \cite{xu_point-nerf_2022} to enhance rendering efficiency. The use of 3D voxel grids on the GPU in combination with a multiresolution hash encoding of the input \cite{muller_instant_2022} significantly reduces the operations needed and permits real-time performance.

While achieving excellent reconstruction quality and speed, many NeRF-style approaches require exhaustive memory resources. This affects both the training and rendering times and often prohibits the use of such representations in applications like network streaming and mobile rendering. To overcome these limitations, dedicated compression schemes for the learned parametrizations on regular grids have been proposed, including vector quantized feature encoding \cite{li_compressing_2023}, learned tensor decomposition \cite{chen_tensorf_2022} or frequency domain transformations\cite{zhao_tinynerf_2023,rho_masked_2023}. 



\begin{figure}[t!]
    \centering
    \begin{tikzpicture}[font=\sffamily]
        \node[anchor=north west,inner ysep=0, inner xsep=0.5cm] at (0,0) {\includegraphics[width=0.45\linewidth]{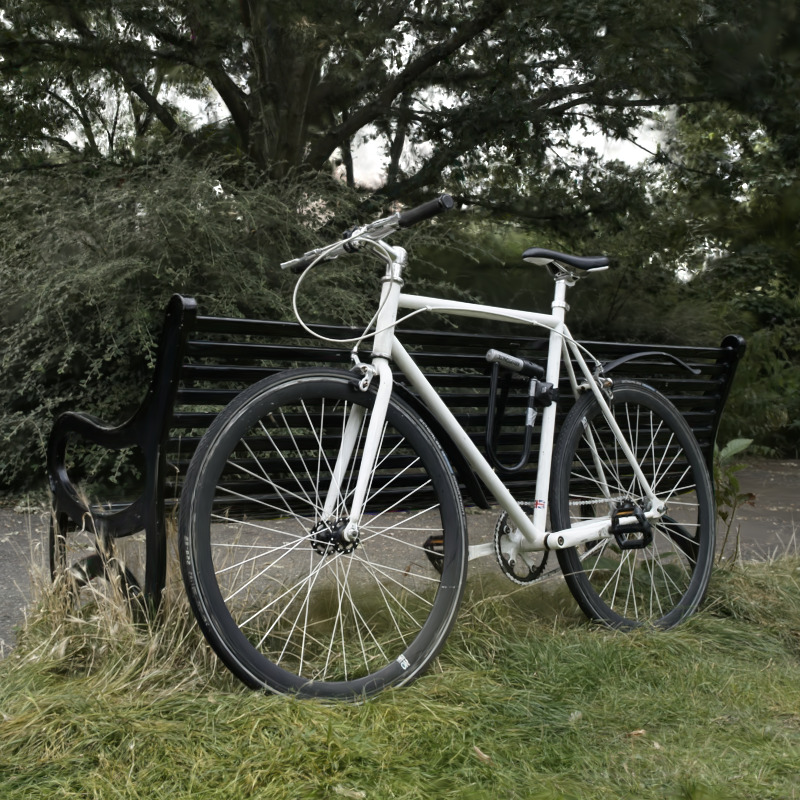}};
        \node[anchor=north east,inner sep=0] at (0,0) {\includegraphics[width=0.45\linewidth]{img/bike.jpg}};
        \node[text=white, anchor=east,fill=black] at (-0.05,-0.3) {\textbf{93 FPS}};
        \node[text=gray, anchor=east,fill=black] at (-2.3,-0.3) {\textbf{54 FPS}};
        
        \node[text=white, anchor=east,fill=black] at (4.2,-0.3) {\textbf{321 FPS}};
        \node[text=gray, anchor=east,fill=black] at (2.15,-0.3) {\textbf{211 FPS}};

        \node[rectangle,text=white,fill=black,anchor=north east,minimum width=0.45\linewidth, inner ysep=0.2cm] at (0,-4) {25.2 PSNR / \color{red}{\textbf{1.5 GB}}};
        
        \node[rectangle,text=white,fill=black,anchor=north west,minimum width=0.45\linewidth,outer ysep=0, outer xsep=0.5cm, inner ysep=0.2cm] at (0,-4) {25.0 PSNR / \color{red}{\textbf{47 MB}}};
        \node [fill=white,draw=black,single arrow, anchor=center] at (0.3,-2) {31$\times$ Compression};
    \end{tikzpicture}
    \caption{Our method achieves a $31\times$ compression at indiscernible loss in image quality and greatly improves rendering speed compared to \cite{kerbl_3d_2023}. Framerates in grey and white, respectively, are taken on NVIDIA's RTX 3070M and RTX A5000 at 1080p resolution.}
    \label{fig:preview}
\end{figure}

Recently, differentiable 3D Gaussian splatting \cite{kerbl_3d_2023} has been introduced to generate a sparse adaptive scene representation that can be rendered at high speed on the GPU.
The scene is modeled as a set of 3D Gaussians with shape and appearance parameters, which are optimized via differentiable rendering to match a set of recorded images.
The optimized scenes usually consist of millions of Gaussians and require up to several gigabytes of storage and memory.  
This makes rendering difficult or even impossible on low-end devices with limited video memory, such as handhelds or head-mounted displays.
Gaussians are rendered using a specialized compute pipeline, which shows real-time performance on high-end GPUs.
This pipeline, however, cannot be seamlessly integrated into VR/AR environments or games to work in tandem with hardware rasterization of polygon models. 

We address the storage and rendering issue of 3D Gaussian splatting by compressing the reconstructed Scene parameters and rendering the compressed representation via GPU rasterization. 
To compress the scenes, we first analyze its components and observe that the SH coefficients and the multivariate Gaussian parameters take up the majority of storage space and are highly redundant.
Inspired by previous work on volumetric radiance field compression\cite{li_compressing_2023,takikawa_variable_2022} and deep network weight quantization, we derive a compression scheme that reduces the storage requirements of typical scenes by up to a factor of $31\times$.
Our compression scheme consists of three main steps:

\begin{itemize}
    \item Sensitivity-aware clustering: We derive a sensitivity measure for each scene parameter by calculating its contribution to the training images. Color information and Gaussian parameters are encoded into compact codebooks via sensitivity-aware vector quantization. 
    \item Quantization-aware fine-tuning: To regain information that is lost during clustering we fine-tune the scene parameters at reduced bit-rates using quantization-aware training.
    \item Entropy encoding: 3D Gaussians are linearized along a space-filling curve to exploit the spatial coherence of scene parameters with entropy and run-length encoding.
\end{itemize}


Further, we propose a renderer for the compressed scenes using GPU sorting and rasterization.
It enables novel view synthesis in real-time, even on low-end devices, and can be easily integrated into applications rendering polygonal scene representations.
Due to the reduced memory bandwidth requirements of the compressed representation and the use of hardware rasterization, a significant speed-up is achieved over the compute pipeline by Kerbl et al. \cite{kerbl_3d_2023}. 

We show the state-of-the-art quality of novel view rendering on benchmark datasets at significantly reduced memory consumption and greatly improved rendering performance (\cref{fig:preview}).
The compressed scenes 
can be used in applications requiring network streaming, 
and they can be rendered on low-end devices with limited video memory and bandwidth capacities. 
We perform a number of experiments on benchmark datasets to empirically validate our method across different scenarios. The contribution of each individual step is demonstrated with an ablation study.

\begin{figure*}[t!]
    \centering
    \includegraphics[width=\textwidth]{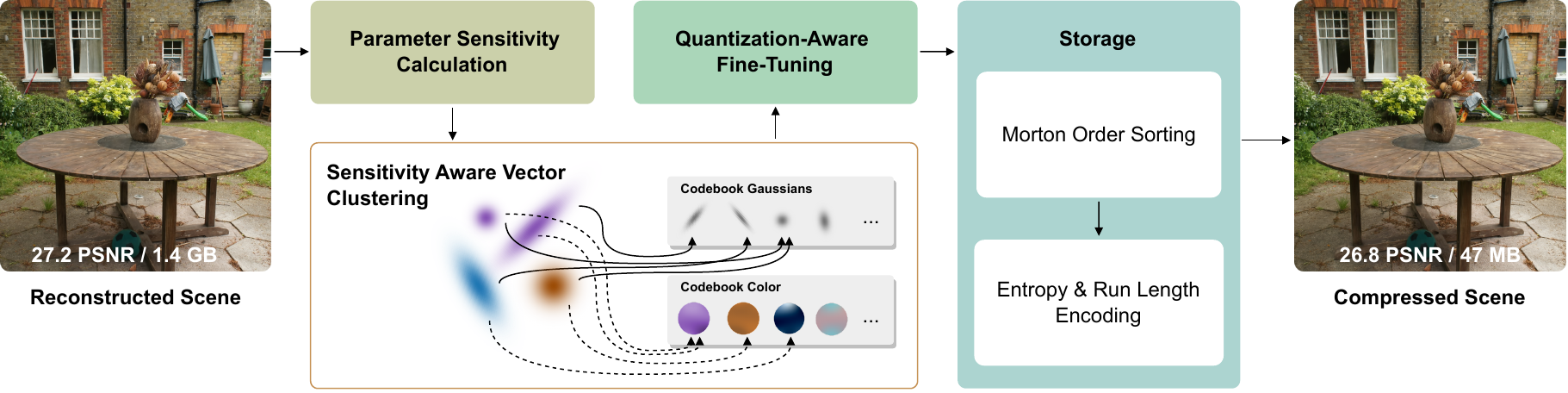}
    \caption{Proposed compression pipeline. Input is an optimized 3D Gaussian scene representation. First, a sensitivity measure is computed for the Gaussian parameters, and color and shape information is compressed into separate codebooks using sensitivity-aware and scale-invariant vector clustering. Next, the compressed scene is fine-tuned on the training images to recover lost information. Finally, the Gaussians are sorted in Morton order and further compressed using entropy and run-length encoding. The shown scene is from \cite{barron_mip-nerf_2022}.
    }
    \label{fig:pipeline}
\end{figure*}

\section{Related Work}
\label{sec:related-work}
Our work builds upon previous works in novel view synthesis via differentiable rendering and scene compression.

{\bf Novel View Synthesis} 
Neural Radiance Fields (NeRF) \cite{mildenhall_nerf_2021} use neural networks to model a 3D scene. They represent the scene as a density field with direction-dependent colors that are rendered with volume rendering. 
The field is reconstructed from a set of images with known camera parameters using gradient-based optimization of the volume rendering process. 

To speed up training and rendering efficiency, a number of different scene models have been proposed.  
Most often, structured space discretizations like voxel grids \cite{hedman_baking_2021,sun_direct_2022,weiss_differentiable_2022}, octrees \cite{fridovich-keil_plenoxels_2022} or hash grids \cite{muller_instant_2022} are used to represent the scene. 
To avoid encoding empty space, point-based representations have been proposed. 
Xu \etal\cite{xu_point-nerf_2022} perform nearest neighbor search in a point cloud to aggregate local features, and Rückert \etal \cite{ruckert_adop_2022} render a point cloud with deep features and use deferred neural rendering to generate the final image.

More recently, differentiable splatting  \cite{kerbl_3d_2023,gao_surfelnerf_2023} has been positioned as a powerful alternative to NeRF-like approaches for novel view synthesis. In particular, 3D Gaussian Splatting \cite{kerbl_3d_2023} offers state-of-the-art scene reconstruction, by using a scene model consisting of an optimized set of 3D Gaussian kernels that can be rendered efficiently.  
Differentiable rendering on a set of training images is used to adaptively refine an initial set of Gaussian kernels and optimize their parameters. 

{\bf NeRF Compression}
While grid-based NeRF variants achieve high rendering performance due to GPU ray-marching, in particular, the use of full spatial grids introduces considerable storage costs. 
Tensor decomposition \cite{chen_tensorf_2022,tang_compressible-composable_2022}, frequency domain transformation \cite{zhao_tinynerf_2023,rho_masked_2023} and voxel pruning \cite{deng_compressing_2023} have been proposed to reduce the memory consumption of grid-based NeRFs.
Takikawa \etal \cite{takikawa_variable_2022} perform vector quantization during training with a learnable index operation. Li \etal\cite{li_compressing_2023} compress grid-based radiance fields by up to a factor of $100\times$ using post-training vector quantization. The use of a hash encoding on the GPU in combination with vector quantization of latent features reduces the required memory and permits high rendering performance \cite{muller_instant_2022} 

A number of works have especially addressed memory reduction during inference, to make grid-based scene representations more suitable for low-end devices with limited video memory  \cite{wadhwani_squeezenerf_2022,reiser_merf_2023}.  
To our knowledge, our approach is the first that aims at the compression of point-based radiance fields to enable high-quality novel view synthesis at interactive frame rates on such devices. 


\textbf{Quantization-Aware Training}
Rastegari et al. \cite{rastegari_xnor-net_2016} simulate weight quantization during training to reduce quantization errors when using low-precision weights for inference. 
The use of quantization-aware training has been explored for neural scene representations \cite{gordon_quantizing_2023} and voxel-based NeRFs \cite{kang_ternarynerf_2022}, demonstrating effective weight quantization with negligible loss in rendering quality.  

To reduce the size and latency of neural networks, various approaches for weight quantization have been explored \cite{rastegari_xnor-net_2016,gordon_quantizing_2023,kang_ternarynerf_2022,jacob_quantization_2018}. 
These methods rely on the observation that in most cases a lower weight precision is required for model inference than for training (e.g., 8-bit instead of 32-bit). 
In post-training quantization, the model weights are reduced to a lower bit representation after training. 
In quantization-aware training, the quantization is simulated during training while operations are performed at full precision to obtain numerically stable gradients.
For storage and inference, the low precision weights can then be used with minor effects on the output. 


\section{Differentiable Gaussian Splatting} 
\label{sec:prerequisites}

Differentiable Gaussian splatting \cite{kerbl_3d_2023} builds upon EWA volume splatting \cite{zwicker_ewa_2001} to efficiently compute the projections of 3D Gaussian kernels onto the 2D image plane. On top of that, differentiable rendering is used to optimize the number and parameters of the Gaussian kernels that are used to model the scene.

The final scene representation comprises a set of 3D Gaussians, each  described by a covariance matrix $\Sigma \in\mathbb{R}^{3\times3}$ centered at location $x\in\mathbb{R}^3$.
The covariance matrix can be parameterized by a rotation matrix $R$ and a scaling matrix $S$.
For independent optimization of $R$ and $S$, Kerbl \etal \cite{kerbl_3d_2023} represent the rotation with a quaternion $q$ and scaling with a vector $s$, 
both of which can be converted into their respective matrices. 
In addition, each Gaussian has its own opacity $\alpha \in [0,1]$ and a set of spherical harmonics (SH) coefficients to reconstruct a view-dependent color. 

The 2D projection of a 3D Gaussian is again a Gaussian with covariance \begin{equation}
    \Sigma' = JW\Sigma W^TJ^T,
\end{equation}
where $W$ is the view transformation matrix and $J$ is the Jacobian of the affine approximation of the projective transformation. This allows to evaluate the 2D color and opacity footprint of each projected Gaussian. A pixel's color $C$ is then computed by blending all $N$ 2D Gaussians contributing to this pixel in sorted order: 
\begin{equation}
    C = \sum_{i\in N} c_i \alpha_i \prod_{j=1}^{i-1}(1-\alpha_j).
\end{equation}
Here, $c_i$ and $\alpha_i$, respectively, are the view-dependent color of a Gaussian and its opacity, modulated by
the exponential falloff from the projected Gaussian's center point. 

The position $x$, rotation $q$, scaling $s$, opacity $\alpha$, and SH coefficients of each 3D Gaussian are optimized so that the rendered 2D Gaussians match the training images. 
For more details on the reconstruction process, we refer to the original paper by Kerbl \etal. \cite{kerbl_3d_2023}.

\section{Sensitivity-Aware Scene Compression}
We compress a set of optimized 3D Gaussian kernels as follows: 
First, sensitivity-aware vector clustering is used to cluster the Gaussian appearance and shape parameters into compact codebooks (\cref{sec:clustering}).
Second, the clustered and other scene parameters are fine-tuned on the training images to recover information lost due to clustering.  
We use quantization-aware training in this step to reduce the scene parameters to a lower bit-rate representation (\cref{sec:qa-training}). By linearizing the set of 3D Gaussians along a space-filling curve, entropy and run-length encoding can exploit the spatial coherence of Gaussian parameters to further compress the scene (\cref{sec:entropy-encoding}). An overview of the proposed compression pipeline is shown in Fig. \ref{fig:pipeline}.

\subsection{Sensitivity-Aware Vector Clustering}
\label{sec:clustering}

Inspired by volumetric NeRF compression \cite{li_compressing_2023,takikawa_variable_2022}, we utilize vector clustering for compressing 3D Gaussian kernels. 
We use clustering to encode SH coefficients and Gaussian shape features (scale and rotation) into two separate codebooks.
As a result, each Gaussian can be compactly encoded via two indices into the codebooks stored alongside.  

{\bf Parameter Sensitivity}: The sensitivity of the reconstruction quality to changes of the Gaussian parameters is not consistent. 
While a slight change in one parameter of a Gaussian can cause a significant difference in the rendered image, a similar change in another parameter or the same parameter of another Gaussian can have low or no effect.  

We define the sensitivity $S$ of image quality to changes in parameter $p$ with respect to the training images as 
\begin{align}\label{eq:sensitivity}
    S(p) &= \frac{1}{\sum_{i=1}^N P_i} \sum_{i=1}^N \big{|} \frac{\partial E_i}{\partial p} \big|.
\end{align}
$N$ is the number of images in the training set used for scene reconstruction, and $P_i$ is the number of pixels in image $i$.
$E$ is the total image energy, i.e., the sum of the RGB components over all pixels. The sensitivity of $E$ to changes in $p$ is considered via the gradient of $E$ with respect to $p$, i.e., a large gradient magnitude indicates high sensitivity to changes in the respective parameter.  
With this formulation, the sensitivity to every parameter can be computed with a single backward pass over each of the training images.

{\bf Sensitivity-aware k-Means}: 
Given a vector $\boldsymbol{x}\in\mathbb{R}^D$, we define its sensitivity as the maximum over its component's sensitivity:
\begin{equation}
    S(\boldsymbol{x}) = \max_{d\in [1..D]} S(x_d).
\end{equation}

The sensitivity measure is then used for sensitivity-aware clustering, i.e., to compute codebooks $\mathbf{C}\in \mathbb{R}^{K\times D}$ with $K$ representatives $\boldsymbol{c_k}\in \mathbb{R}^{D}$ (so-called centroids). 


We define the weighted distance between a vector $\boldsymbol{x}$ and a centroid $\boldsymbol{c_k}$
as 
\begin{equation}
    \mathcal{D}(\boldsymbol{x},\boldsymbol{c}_k) = S(\boldsymbol{x})  \lVert \boldsymbol{x}-\boldsymbol{c}_k \rVert_2^2.
\end{equation}
A codebook is then obtained by using k-Means clustering with $\mathcal{D}$ as a similarity measure.
The codebooks are initialized randomly with a uniform distribution within the minimum and maximum values of each parameter.
The centroids are computed with 
an iterative update strategy: 
In each step, the pairwise weighted distances between the vectors $\boldsymbol{x}$ and the codebook vectors $\boldsymbol{c}_k$ are calculated, and each vector is assigned to the centroid to which it has the minimum distance.
Each centroid is then updated as 

\begin{equation}
    \boldsymbol{c}_k = \frac{1}{\underset{\boldsymbol{x}_i\in A(k)}{\sum} S(\boldsymbol{x}_i)} \sum_{\boldsymbol{x}_i\in A(k)} S(\boldsymbol{x}_i) \boldsymbol{x}_i
\end{equation}
Where $A(k)$ is the set of vectors assigned to centroid $c_k$.

For performance reasons, a batched clustering strategy is used \cite{sculley_web-scale_2010}.   
In each update step, a random subset of vectors is picked and used to compute the update step. Then, the centroids are updated using the moving average with a decay factor $\lambda_d$.


{\bf Color Compression}:
Each Gaussian stores 
SH coefficients to represent the direction-dependent RGB color (e.g., 48 coefficients in \cite{kerbl_3d_2023}).
We treat SH coefficients as vectors and compress them into a codebook using sensitivity-aware vector clustering. 

\begin{figure}[t]
    \centering
    \includegraphics[width=\linewidth]{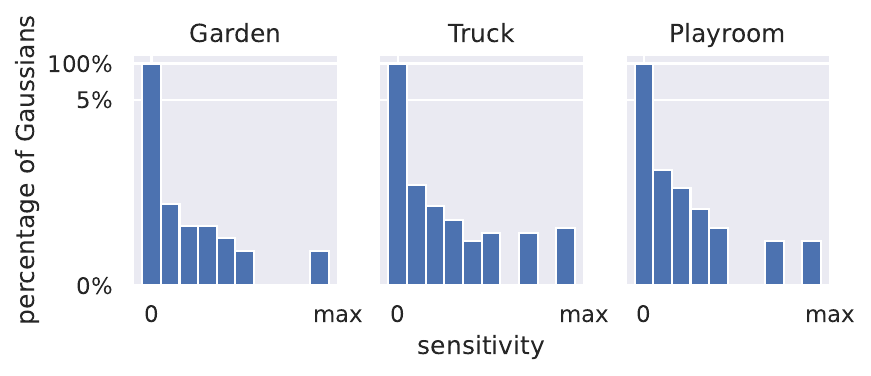}
    \caption{Histograms of maximum sensitivity to changes of SH coefficients for different scenes. Only SH coefficients of a tiny fraction of all Gaussians strongly affect image quality.}
    \label{fig:sensitivity-plot}
\end{figure}

For volumetric NeRF models, Li~\etal~\cite{li_compressing_2023} have shown that only a small number of voxels contribute significantly to the training images. 
Thus, they propose to keep the color features that contribute the most and only compress the remaining features with vector clustering. 
We observe a similar behavior for 3D Gaussians, as shown for some benchmark scenes in \cref{fig:sensitivity-plot}.
For a small percentage of all SH coefficients ($<5\%$), the sensitivity measure indicates a high sensitivity towards the image quality.
Thus, to keep the introduced rendering error low, we do not consider the SH vectors of Gaussians with a sensitivity higher than a threshold $\beta_c$ in the clustering process. 
These vectors are added to the codebook after clustering. 

{\bf Gaussian Shape Compression}:
A 3D Gaussian kernel  
can be parameterized with a rotation matrix $R$ and a scaling vector $\mathbf{s}$.
We observe that for typical scenes, the shapes of the Gaussians are highly redundant up to a scaling factor. 
Thus, we re-parameterize the scaling vector $\mathbf{s} = \eta \hat{\mathbf{s}}$,
where $\eta=\lVert\mathbf{s}\rVert_2$ is the scalar scaling factor and $\hat{\mathbf{s}}=\eta^{-1}\mathbf{s}$ is the normalized scaling vector.
With $\hat{S}=\mathrm{diag}(\hat{\mathbf{s}})$, the normalized covariance matrix is
\begin{equation}
    \hat{\Sigma} = (R\hat{S})(R\hat{S})^T = \frac{1}{\eta^2} \Sigma.
\end{equation}

Clustering is then performed using the normalized covariance matrices, and each Gaussian stores, in addition to a codebook index, the scalar scaling factor $\eta$. 
We compute the sensitivity to each of the matrix entries and perform sensitivity-aware vector quantization to compress them into a codebook. The sensitivity plots for Gaussian shape parameters 
look mostly similar to the SH plots shown in \cref{fig:sensitivity-plot}. As for SH coefficients, Gaussians with a maximum sensitivity over a threshold $\beta_g$ are not considered for clustering and are added to the codebook. 

Note that k-Means clustering of normalized covariance matrices results in covariance matrices that are again normalized.
However, due to floating point errors, clustering can lead to non-unit scaling vectors.
To counteract this problem, we re-normalize each codebook vector after each update step by dividing it through the trace of the covariance metric. In the appendix, we prove both the normalization preserving properties of k-Means and re-normalization.

After clustering, each codebook entry is decomposed into a rotation and scale parameter using an eigenvalue decomposition.
This is required for quantization-aware training since direct optimization of the matrix is not possible\cite{kerbl_3d_2023}. 
In the final codebook, each matrix's rotation and scaling parameters are encoded via 4 (quaternion) plus 3 (scaling) scalar values. 




\subsection{Quantization-Aware Fine-Tuning}
\label{sec:qa-training}

To regain information that is lost due to parameter quantization, the parameters can be fine-tuned on the training images after compression  \cite{li_compressing_2023,yu_plenoctrees_2021}. 
To do so, we use the training setup described by Kerbl \etal \cite{kerbl_3d_2023}.
We optimize for the position, opacity, and scaling factor of each Gaussian as well as the color and Gaussian codebook entries. 
For the two codebooks, the incoming gradients for each entry are accumulated and then used to update the codebook parameters. 

For fine-tuning, we utilize quantization-aware training with Min-Max quantization (k-bit Quantization \cite{rastegari_xnor-net_2016}) to represent the scene parameters with fewer bits.
In the forward pass, the quantization of a parameter $p$ is simulated using a rounding operation considering the number of bits and the moving average of each parameter's minimum and maximum values.
The backward pass ignores the simulated quantization and calculates the gradient w.r.t. $p$ as without quantization.
After training, the parameters can be stored with only $b$-bit precision (e.g., 8-bit), while the minimum and maximum values required for re-scaling are stored at full precision (e.g., 32-bit float).

Quantization of opacity is applied after the sigmoid activation function.
Quantization of the scaling and rotation vector is applied before the respective normalization step.
For the scale factor parameter, the quantization is applied before the activation (exponential function) to allow for a fine-grained representation of small Gaussians without losing the ability to model large ones. 
We quantize all Gaussian parameters despite position to an 8-bit representation with the Min-Max scheme. 16-bit float quantization is used for position, as a further reduction decreases the reconstruction quality considerably. 

\subsection{Entropy Encoding}
\label{sec:entropy-encoding}

After quantization-aware fine-tuning, the compressed scene representation consists of a set of Gaussians and the codebooks storing SH coefficients and shape parameters.
Indices into the codebooks are stored as 32-bit unsigned integers. 

The data is then compressed using DEFLATE~\cite{deutsch_deflate_1996}, which utilizes a combination of the LZ77~\cite{ziv_universal_1977} algorithm and Huffman coding.
In the reconstructed scenes, many features, such as color, scaling factor, and position, are spatially coherent.
By ordering the Gaussians according to their positions along a Z-order curve in Morton order, the coherence can be exploited and the effectivity of run-length encoding (LZ77) can be improved.
The effect on the compressed file size is analyzed in the ablation study in \cref{sec:ablation}.
Note that entropy encoding reduces the two codebook indices to their required bit-length according to the codebook sizes. 



\section{Novel View Rendering}

Kerbl~\etal~\cite{kerbl_3d_2023} propose a software rasterizer for differentiable rendering and novel view synthesis. 
To render 3D Gaussian scenes fast especially on low-power GPUs, our novel view renderer utilizes hardware rasterization.

\textbf{Preprocess}:
In a compute pre-pass, 
Gaussians whose $99\%$ confidence interval does not intersect the view frustum after projection are discarded. For the remaining Gaussians, the direction-dependent color is computed with the SH coefficients.
The color, the Gaussian's opacity, projected screen-space position, and covariance values are stored in an atomic linear-append buffer. 
The covariance values indicate the orientation and size of the 2D Gaussian into which a 3D Gaussian projects under the current viewing transformation \cite{zwicker_ewa_2001}. 
As in \cite{kerbl_3d_2023}, Gaussians are then depth-sorted to enable order-dependent blending. We use the Onesweep sorting algorithm by Adinets and Merrill \cite{adinets_onesweep_2022} to sort the Gaussians directly on the GPU. 
Due to its consistent performance, the implementation is well suited for embedding into real-time applications. 

\begin{table*}[ht]
\resizebox{\linewidth}{!}{
\begin{tabular}{l|SSSS|SSSS|S}
\hline
Method         & \multicolumn{4}{c|}{3D Gaussian Splatting} & \multicolumn{4}{c|}{Ours} &                \\
Dataset        & {PSNR $\uparrow$ }     & {SSIM $\uparrow$} & {LPIPS $\downarrow$} & {SIZE $\downarrow$} & {PSNR $\uparrow$ }    & {SSIM $\uparrow$} & {LPIPS $\downarrow$} & {SIZE $\downarrow$} & {Compression Ratio $\uparrow$}\\ \hline
Synthetic-NeRF~\cite{mildenhall_nerf_2021} & 33.21      & 0.969	& 0.031	& 69.89  & 32.936  & 0.967 & 0.033  & 3.68 	& 19.17 \\
Mip-NeRF360~\cite{barron_mip-nerf_2022}    & 27.21      & 0.815 & 0.214 & 795.26 & 26.981  & 0.801 & 0.238  & 28.80 & 26.23 \\
Tanks\&Temples~\cite{knapitsch_tanks_2017} & 23.36      & 0.841 & 0.183 & 421.90 & 23.324  & 0.832 & 0.194	& 17.28 & 23.26 \\
Deep Blending~\cite{hedman_deep_2018}  & 29.41      & 0.903 & 0.243 & 703.77 & 29.381  & 0.898 & 0.253	& 25.30 & 27.81 \\ \hline
average*       & 26.58      & 0.853	& 0.213	& 640.31 & 26.560  & 0.844 & 0.238	& 23.73 & 25.77 \\ \hline

\end{tabular}
}
\caption{Quantitative comparison to 3D Gaussian Splatting. Size is measured in Megabytes. *Synthetic scenes are excluded.}
\label{tab:dataset-metrics}
\end{table*}

\textbf{Rendering}:
Gaussians are finally rendered in sorted order via GPU rasterization. 
For each Gaussian, one planar quad (a so-called splat) consisting of two triangles is rendered.
A vertex shader computes the screen space vertex positions of each splat from the 2D covariance information. The size of a splat is set such that it covers the $99\%$ confidence interval of the projected Gaussian. 
The vertex shader simply outputs the color computed in the pre-pass and the 2D splat center as input to the pixel shader. The pixel shader then discards fragments outside the $99\%$ confidence interval. All remaining fragments use their distance to the splat center to compute the exponential color and opacity falloff and blend their final colors into the framebuffer.



\section{Experiments}

\subsection{Datasets}

We evaluate our compression and rendering method on the \textbf{Mip-Nerf360\cite{barron_mip-nerf_2022}} indoor and outdoor scenes, 
two scenes from the \textbf{Tanks\&Temples\cite{knapitsch_tanks_2017}} and \textbf{Deep Blending \cite{hedman_deep_2018}} dataset, and \textbf{NeRF-Synthetic\cite{mildenhall_nerf_2021}}.
For Mip-Nerf360, Tanks\&Temples and Deep Blending the reconstructions from Kerbl \etal\cite{kerbl_3d_2023} were used.
We generated the 3D Gaussian representation for NeRF-Synthetic ourselves. 

\subsection{Implementation Details}

We use a decay factor $\lambda_d = 0.8$ for batched clustering with 800 update steps for the Gaussians and 100 for the SH coefficients. 
A batch size of $2^{18}$ is used for the color features, and $2^{20}$ for the Gaussian parameters. 
We use 4096 as the default codebook size in all our experiments and set $\beta_c=6\cdot10^{-7}$ and $\beta_g=3\cdot10^{-6}$.
We perform 5000 optimization steps of quantization-aware fine-tuning.

The renderer is implemented with the WebGPU graphics API in the Rust programming language. Thus, it can run in a modern web browser on a large variety of devices.
More details about the implementation can be found in the supplementary material.
The source code is available at \url{https://github.com/KeKsBoTer/c3dgs}.

\subsection{Results}

We use the scenes reconstructed by 3D Gaussian Splatting \cite{kerbl_3d_2023} and compress them using the proposed method.
For all scenes, we evaluate the PSNR, SSIM, and LPIPS \cite{zhang_unreasonable_2018} before and after compression.
\cref{tab:dataset-metrics} shows the results for different datasets. 

\begin{figure}[ht]
    \centering
    \includegraphics[width=\linewidth]{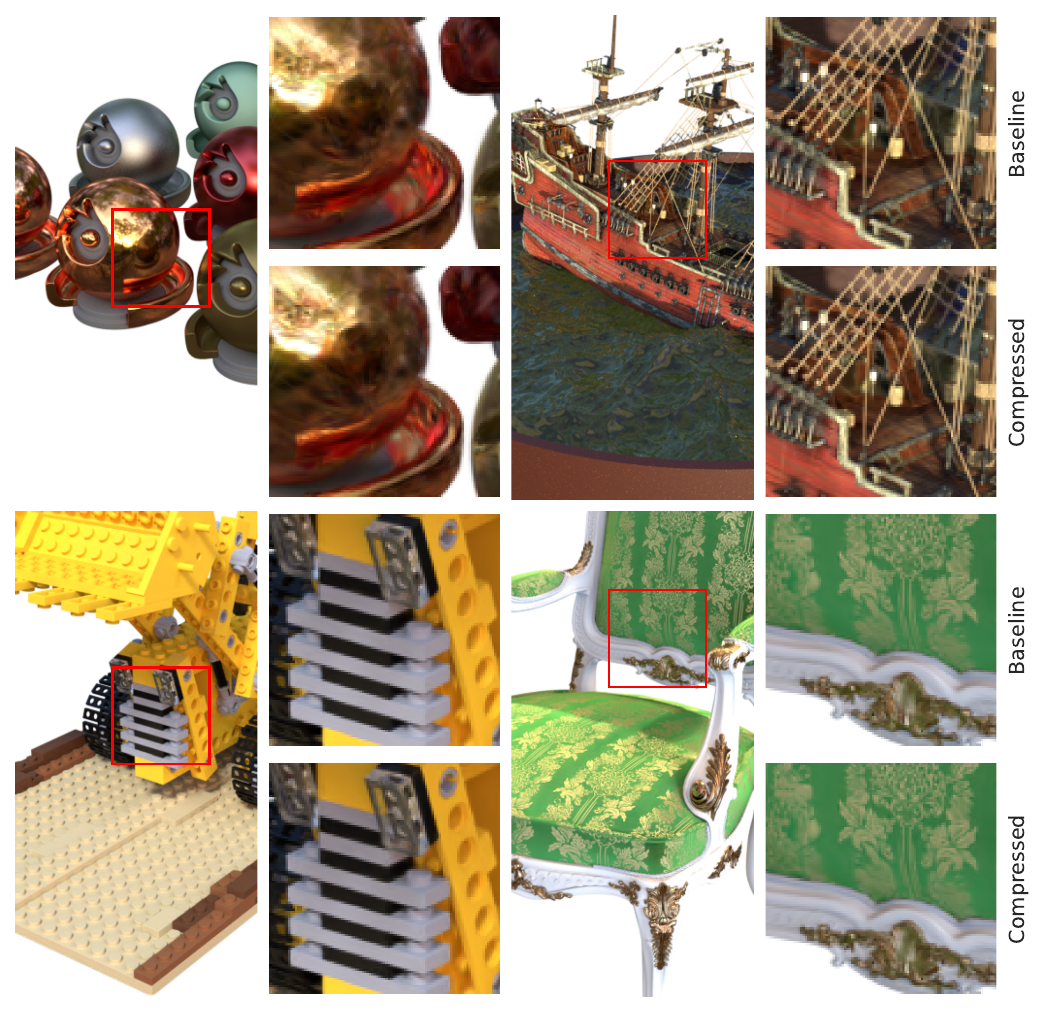}
    \caption{3D Gaussian splatting of synthetic scenes~\cite{mildenhall_nerf_2021}. Uncompressed (Baseline)  vs. compressed scene.}
    \label{fig:split-syn}
\end{figure}

\begin{figure*}
    \centering
    \includegraphics[width=0.85\textwidth]{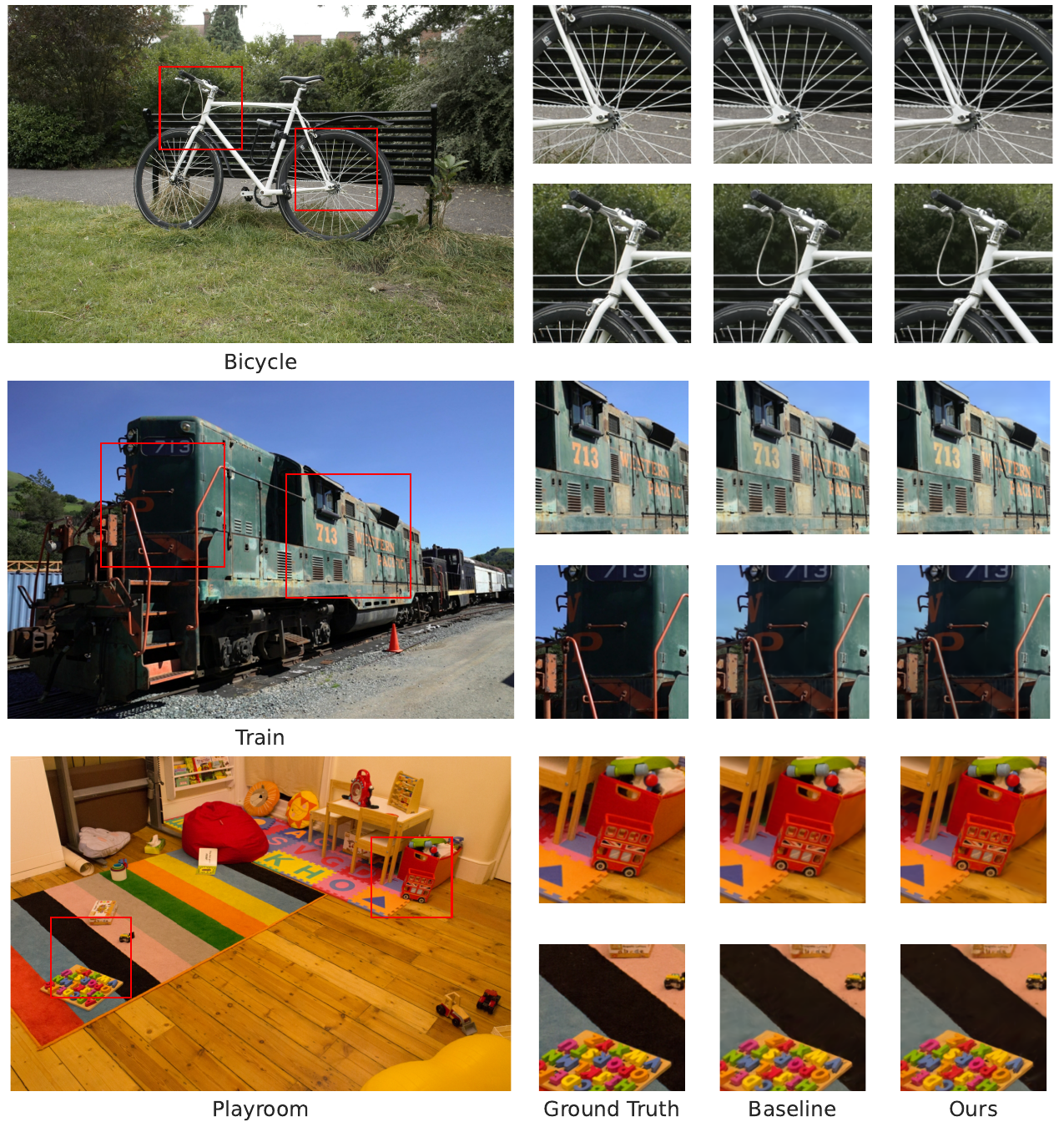}
    \caption{
    Ground truth images from the test set, results of Kerbl et al.~\cite{kerbl_3d_2023} (Baseline), results using the compressed representation (Ours).}
    \label{fig:comparison-crop}
\end{figure*}

Our compression method achieves a compression ratio of up to 31$\times$ with an average of 26$\times$ at the indiscernible loss of quality (0.23 PSNR on average) for real-world scenes. 
Here, it should be noted that a difference of 0.5 PSNR is considered indistinguishable for the human eye \cite{salomon_handbook_2010}.
For some of the scenes, \cref{fig:comparison-crop} compares training images to the renderings of the uncompressed and compressed scenes.
\cref{fig:split-syn} shows close-up views of uncompressed and compressed synthetic scenes. 
More comparisons and results are given in the supplementary material.

\begin{figure*}
    \includegraphics[width=\textwidth]{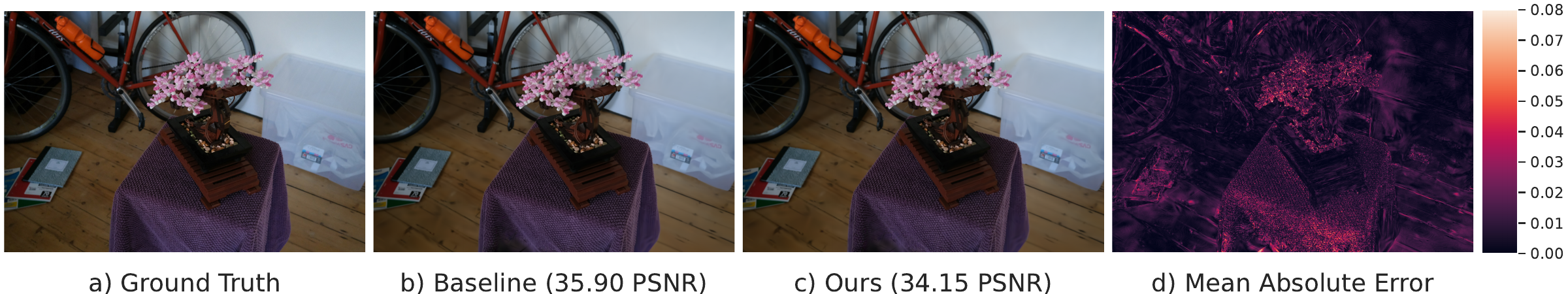}
    \caption{Test image with the highest drop in PSNR in all scenes used in this work. d) Per pixel mean absolute error between Kerbl et al. \cite{kerbl_3d_2023} b) and our approach c).}
    \label{fig:worst}
\end{figure*}

\textbf{Image Quality Loss}
\cref{fig:comparison-crop} shows that it is almost impossible to spot the difference between the uncompressed and the compressed scenes.  
We also analyze the images from all test sets with the largest drop in PSRN. The image which could be reconstructed least accurately is shown in \cref{fig:worst}. We observe that the loss is mainly due to very subtle color shifts below what can be perceived by the human eye. 

\textbf{Compression Runtime}
The compression process takes about 5-6 minutes and increases the reconstruction time by roughly 10\%.
The timings of each individual steps are given in the supplementary material.

\textbf{Rendering Times}
We see a significant increase of up to a factor of $4\times$ in rendering speed (see \cref{tab:rendering-speed}). Roughly a 2x increase can be attributed to the compressed data's reduced bandwidth requirements, hinting at the software rasterizer's memory-bound performance by Kerbl et al. \cite{kerbl_3d_2023}.
The additional speedup is achieved by the hardware rasterization-based renderer, which pays off on low- and high-end GPUs.
Timings of the different rendering stages are given in the supplementary material.
\begin{table}[h]
\resizebox{\linewidth}{!}{%
\begin{tabular}{lr|c|c|c|c}
\hline
 &
  \multicolumn{1}{l|}{} &
  \begin{tabular}[c]{@{}l@{}}NVIDIA \\ RTX A5000\end{tabular} &
  \begin{tabular}[c]{@{}l@{}}NVIDIA \\ RTX 3070M\end{tabular} &
  \begin{tabular}[c]{@{}l@{}}Intel UHD \\ Graphics 11\end{tabular} &
  \begin{tabular}[c]{@{}l@{}}AMD Radeon\\ R9 380\end{tabular} \\ \hline
\vbox{\hbox{\multirow{3}{*}{\rotatebox[origin=c]{90}{Bicycle}}}} & Kerbl et al. \cite{kerbl_3d_2023}     & 93  & 54 & - & -  \\
                         & Ours       & 215 & 134 & 9 & 41 \\
                         & Compressed & 321 & 211 & 16 & 83 \\ \hline
\vbox{\hbox{\multirow{3}{*}{\rotatebox[origin=c]{90}{Bonsai}}}}  & Kerbl et al. \cite{kerbl_3d_2023}      & 184  & 122 & -   & -\\
                         & Ours       & 414 & 296 & 23 & 76 \\
                         & Compressed & 502 & 380 & 28 & 128 \\ \hline
\end{tabular}%
}
\caption{Rendering performance at 1080p resolution in frames per second, averaged over all training images. Bicycle consists of $6.1$ million 3D Gaussians, Bonsai of $1.2$ million 3D Gaussians.}
\label{tab:rendering-speed}
\end{table}

\subsection{Ablation Study}
\label{sec:ablation}

In a number of experiments we evaluate the components of our compression pipeline. This includes a detailed analysis of the influence of the hyper-parameters.

\begin{table}[]
\resizebox{\linewidth}{!}{
\begin{tabular}{l|SSS|S}
\hline
                   &{ PSNR $\uparrow$}   & {SSIM $\uparrow$}   & {LPIPS $\downarrow$}  & {SIZE $\downarrow$}   \\ \hline
baseline                & 27.179 & 0.861 & 0.115  & 1379.99 \\ \hline
+ Pruning             &  27.083 &  0.856 &  0.118 &  1217.25 \\
+ Color Clustering    &  25.941 &  0.818 &  0.178 &   278.41 \\
+ Gaussian Clustering &  25.781 &  0.811 &  0.186 &   164.15 \\
+ QA Finetune         &  26.746 &  0.844 &  0.144 &    86.69 \\
+ Encode              &  26.746 &  0.844 &  0.144 &    58.40 \\
+ Morton Order        &  26.746 &  0.844 &  0.144 &    46.57 \\\hline
\end{tabular}
}
\caption{Losses introduced and regained by individual stages of the compression pipeline. 
Experiments were performed with the garden scene from Mip-Nerf360\cite{barron_mip-nerf_2022}}
\label{tab:ablation}
\end{table}

\textbf{Loss Contribution}
\cref{tab:ablation} indicates that the most significant loss increase comes from the compression of the SH coefficients, which, on the other hand, gives the highest memory reduction. Quantization of shape parameters can additionally reduce the memory by about $60\%$, only introducing a slight loss in image quality. Quantization-aware fine-tuning can regain much of the information that is lost due to quantization and further reduces the memory by about $50\%$. Entropy and run length encoding in combination with Morton order layout saves an additional $50\%$ of the memory.

\textbf{Codebook Sizes}
SH coefficients and Gaussian shape parameters are compressed into codebooks of predefined sizes.
\cref{tab:ablation} shows the effects of different codebook sizes on image quality. Errors were averaged over all test images, with the difference to the maximum error given in brackets.
It can be seen that the codebook size has little effect on the average reconstruction error, independent of the scene.
Nevertheless, larger codebooks reduce the maximum error with only minimal memory overhead.

\begin{table}[]
\resizebox{\linewidth}{!}{
\begin{tabular}{lr|ccc|S}
\hline
                      &       & {PSNR $\uparrow$}  & {SSIM $\uparrow$}  & {LPIPS $\downarrow$}  & {SIZE $\downarrow$}   \\ \hline
\multirow{4}*{Color} 
      & 1024 &  26.95(-0.67) &  0.80(-0.02) &  0.24(+0.03) &  28.47 \\
      & 2048 &  26.95(-0.62) &  0.80(-0.02) &  0.24(+0.03) &  28.65 \\
      & \cellcolor{lightgray}4096 &  \cellcolor{lightgray}26.98(-0.63) &  \cellcolor{lightgray}0.80(-0.02) &  \cellcolor{lightgray}0.24(+0.03) &  \cellcolor{lightgray}28.80 \\
      & 8192 &  27.00(-0.58) &  0.80(-0.02) &  0.24(-0.03) &  28.92 \\\hline
\multirow{4}*{Gaussian} 
         & 1024 &  26.95(-0.79) &  0.80(-0.02) &  0.24(+0.03) &  28.14 \\
         & 2048 &  26.97(-0.80) &  0.80(-0.02) &  0.24(+0.03) &  28.45 \\
         & \cellcolor{lightgray}4096 &  \cellcolor{lightgray}26.98(-0.63) &  \cellcolor{lightgray}0.80(-0.02) &  \cellcolor{lightgray}0.24(+0.03) &  \cellcolor{lightgray}28.80 \\
         & 8192 &  26.97(-0.60) &  0.80(-0.02) &  0.24(+0.03) &  29.06 \\\hline
\end{tabular}
}
\caption{
Average reconstruction error over the test images for different codebook sizes, including the maximum deviation from the baseline ($+/-$). Rows marked grey indicate the default configurations. Experiments were performed on the Mip-Nerf360\cite{barron_mip-nerf_2022} dataset. }

\label{tab:codebook-sizes}
\end{table}

\textbf{Sensitivity Thresholds}
The sensitivity thresholds $\beta^c$ and $\beta^g$ are used to decide whether to consider SH coefficients and shape parameters for clustering. They offer a trade-off between quality and compression rate.
The influence of these values is analyzed in \cref{tab:keep-prune}, showing in particular the sensitivity of image quality to the quantization of SH coefficients.   

\begin{table}[]
\resizebox{\linewidth}{!}{
\begin{tabular}{ll|l|l|l|r}
\hline 
 &           & {PSNR $\uparrow$}   & {SSIM $\uparrow$} & {LPIPS $\downarrow$} & {SIZE $\downarrow$}   \\ \hline
\multicolumn{2}{l|}{baseline} & 26.976 & 0.801 & 0.238 & 28.80 \\\hline
\multirow{6}*{$\beta_c$} & $6.0\cdot10^{-8}$ &  $27.22(-0.25)$ &  $0.81(-0.00)$ &  $0.22(+0.00)$ &  56.50 \\
            & $3.0\cdot10^{-7}$ &  $27.09(-0.41)$ & $0.80(-0.01)$ &  $0.23(+0.02)$ &  33.00  \\
            & \cellcolor{lightgray}$6.0\cdot10^{-7}$ & \cellcolor{lightgray}$26.98(-0.63)$ & \cellcolor{lightgray}$0.80(-0.02)$ &\cellcolor{lightgray}$0.24(+0.03)$& \cellcolor{lightgray}28.80 \\
            & $1.2\cdot10^{-6}$ &  $26.87(-0.75)$ & $0.80(-0.02)$ &  $0.24(+0.04)$ &  27.02 \\
            & $6.0\cdot10^{-6}$ &  $26.74(-0.94)$ & $0.80(-0.02)$ &  $0.25(+0.04)$ &  25.97 \\
            & -                 &  $26.55(-1.47)$ & $0.79(-0.03)$ &  $0.25(+0.05)$ &  25.65 \\\hline
\multirow{6}*{$\beta_g$}   & $3.0\cdot10^{-7}$ &  $27.05(-0.41)$ &  $0.80(-0.01)$ & $0.23(+0.03)$ &  33.90 \\
                           & $1.5\cdot10^{-6}$ &  $27.00(-0.62)$ &  $0.80(-0.02)$ & $0.24(+0.03)$ &  29.95 \\
                           & \cellcolor{lightgray}$3.0\cdot10^{-6}$ &\cellcolor{lightgray}$26.98(-0.63)$ &\cellcolor{lightgray}$0.80(-0.02)$ &\cellcolor{lightgray}$0.24(+0.03)$ &\cellcolor{lightgray}28.80 \\
                           & $6.0\cdot10^{-6}$ &  $26.91(-0.77)$ &  $0.80(-0.02)$ & $0.24(+0.03)$ &  28.08 \\
                           & $3.0\cdot10^{-5}$ &  $26.86(-0.72)$ &  $0.80(-0.02)$ & $0.24(+0.04)$ &  27.30 \\
                           & -                 &  $26.80(-0.83)$ &  $0.80(-0.02)$ & $0.25(+0.04)$ &  27.10 \\\hline
\end{tabular}
}
\caption{Sensitivity threshold ablation study. $\beta^c$ and $\beta^g$ are the sensitivity thresholds for controlling which SH vectors and shape parameters are clustered. The average error and in brackets the maximum deviation from the baseline are reported. The last row shows the results when no threshold is considered. The rows marked grey are the default configurations. Experiments were performed with Mip-Nerf360~\cite{barron_mip-nerf_2022} dataset.}
\label{tab:keep-prune}
\end{table}

\subsection{Limitations}

As the main limitation for making the proposed compression and rendering pipeline even more powerful, we see the current inability to aggressively compress the Gaussians' positions in 3D space. We performed experiments where positions were quantized to a lattice structure, and we even embedded these positional constraints into the Gaussian splatting training process.
Unfortunately, we were not able to further compress the positions without introducing a significant error in the rendering process.  



\section{Conclusion}

We have introduced a novel compression and rendering pipeline for 3D Gaussians with color and shape parameters, 
achieving compression rates of up to 31$\times$ and up to a 4$\times$ increase in rendering speed. Our experiments with different datasets have shown that the compression introduces an indiscernible loss in image quality.  
The compressed data can be streamed over networks and rendered on low-power devices, making it suitable for mobile VR/AR applications and games.
In the future, we aim to explore new approaches for reducing the memory footprint during the training phase, and additionally compressing positional information end-to-end. We also believe that 3D Gaussian splatting has the potential for reconstructing volumetric scenes, and we will investigate advanced options for compressing and rendering the optimized representations.

{
    \small
    \bibliographystyle{ieeenat_fullname}
    \bibliography{main}
}
\clearpage
\maketitlesupplementary

\renewcommand\thesubsection{\Alph{subsection}}
\section{Supplementary}
\subsection{Detailed Scene Analysis}

For all scenes used in the paper, we report the 
PSNR, SSIM, LPIPS, memory consumption, and compression ratio of our approach. See 
\cref{tab:results-mip} for Mip-Nerf360~\cite{barron_mip-nerf_2022}, 
\cref{tab:results-deep} for Deep Blending~\cite{hedman_deep_2018}, 
\cref{tab:results-tnt} for Tanks\&Temples~\cite{knapitsch_tanks_2017}, \cref{tab:results-syn}
for the synthetic scenes from ~\cite{mildenhall_nerf_2021}.

\subsection{Image Quality}

\begin{figure*}
	\centering
\begin{tabular}{rccc}
		\raisebox{1.3cm}[0pt][0pt]{\rotatebox[origin=c]{90}{Truck}} &
		\includegraphics[width=0.30\textwidth]{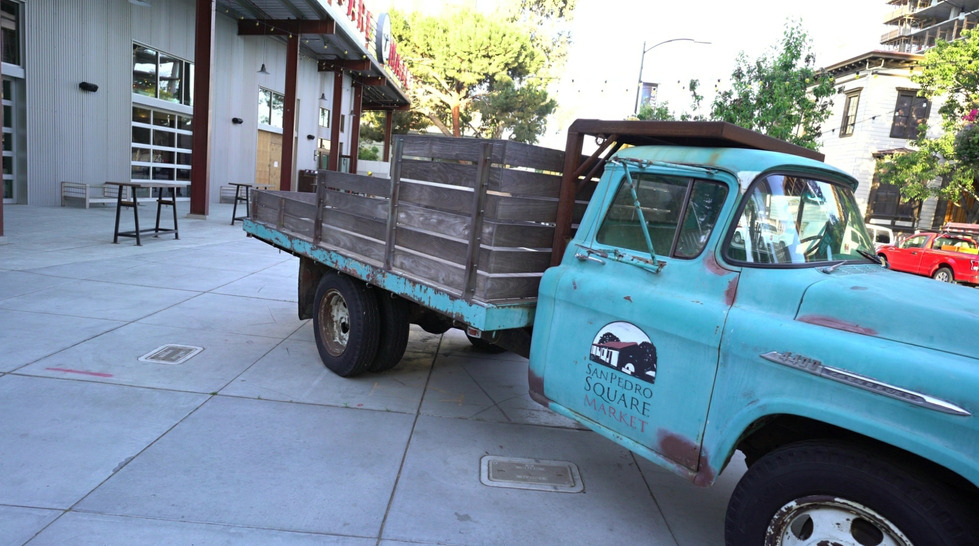}&
		\includegraphics[width=0.30\textwidth]{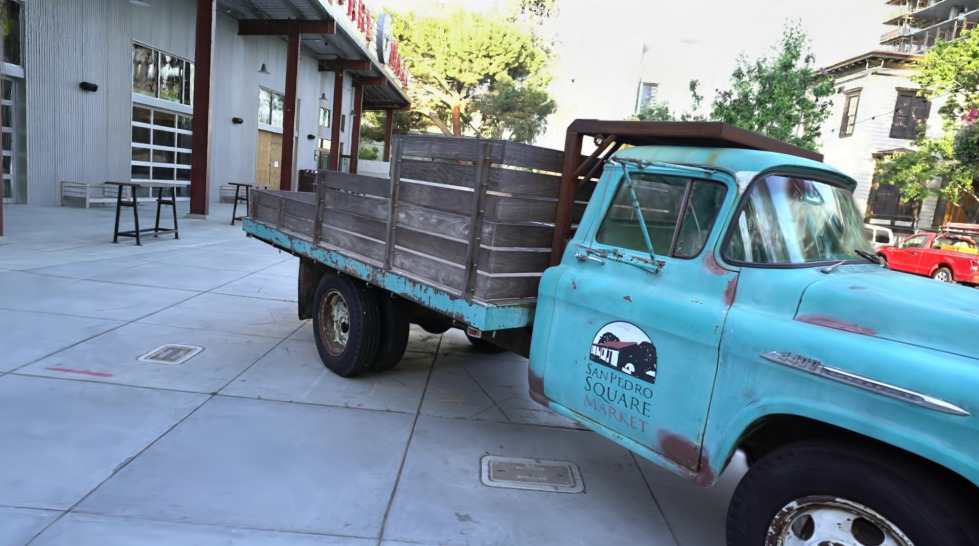}&
		\includegraphics[width=0.30\textwidth]{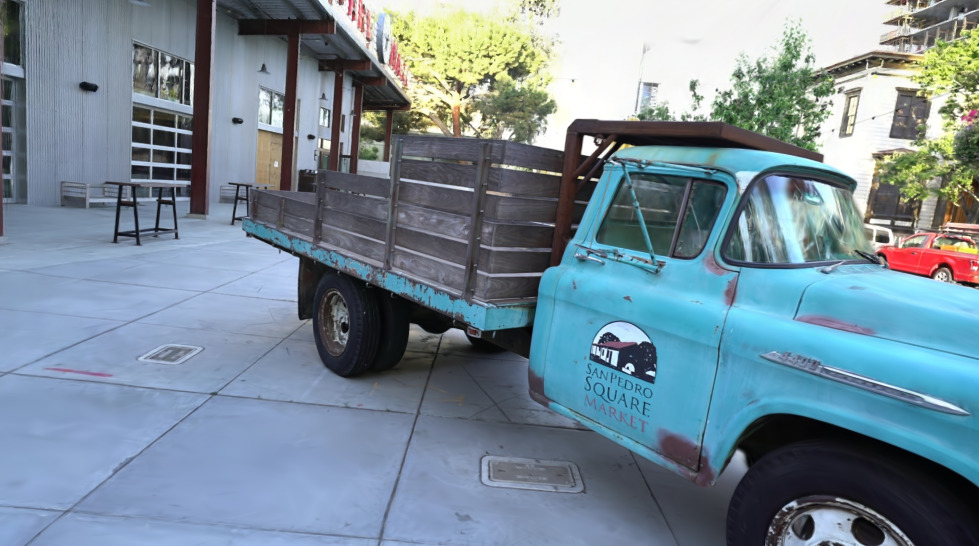}\\
		\raisebox{1.3cm}[0pt][0pt]{\rotatebox[origin=c]{90}{Train}} &
		\includegraphics[width=0.30\textwidth]{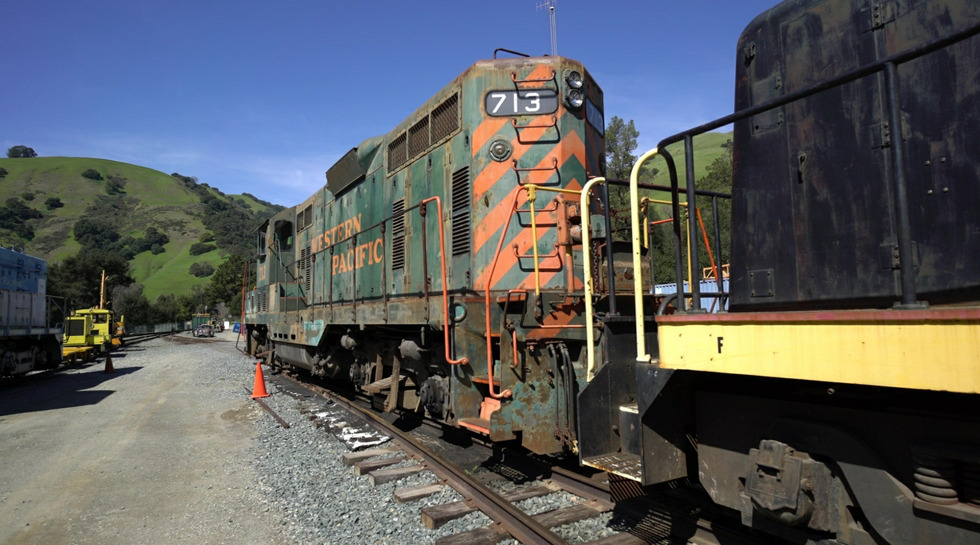}&
		\includegraphics[width=0.30\textwidth]{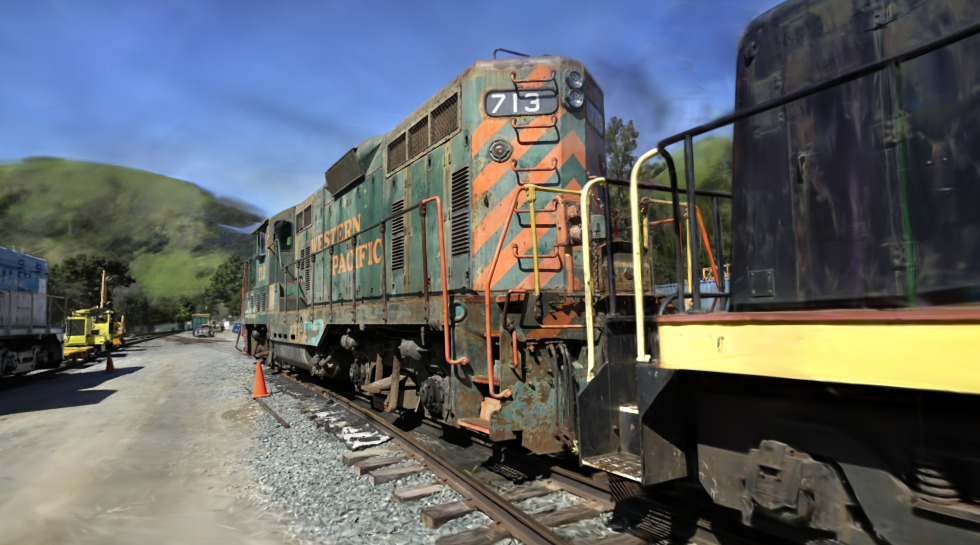}&
		\includegraphics[width=0.30\textwidth]{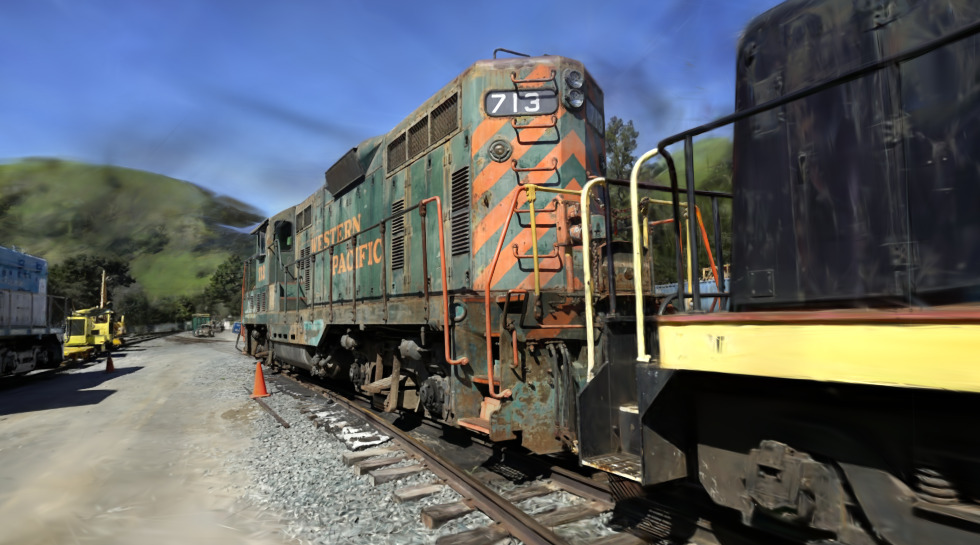}\\
& Ground Truth & Baseline & Compressed
\end{tabular}
    \caption{Random test views for each scene from Tanks\&Temples~\cite{knapitsch_tanks_2017}}
    \label{fig:example-tnt}
\end{figure*}

\begin{figure*}
	\centering
	\begin{tabular}{rccc}
		\raisebox{1.6cm}[0pt][0pt]{\rotatebox[origin=c]{90}{Playroom}} &
		\includegraphics[width=0.30\textwidth]{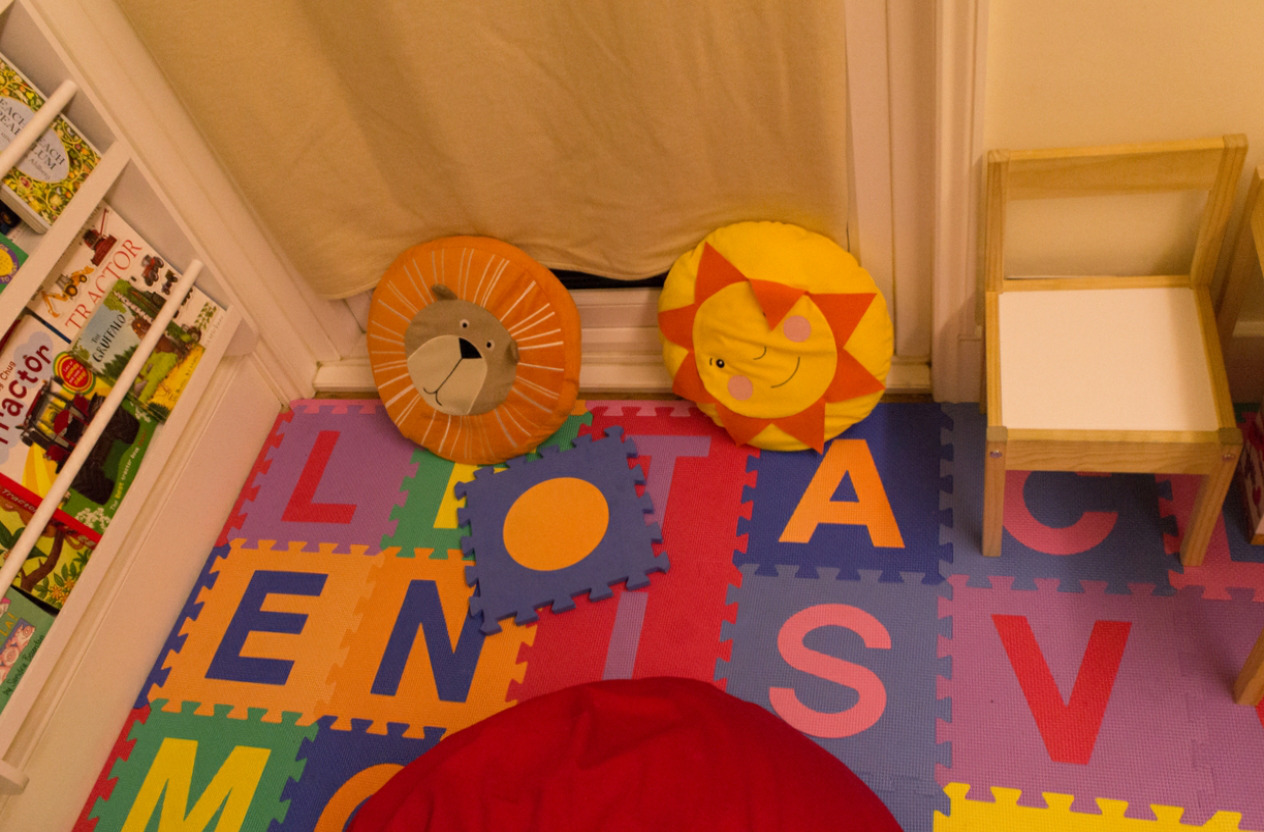}&
		\includegraphics[width=0.30\textwidth]{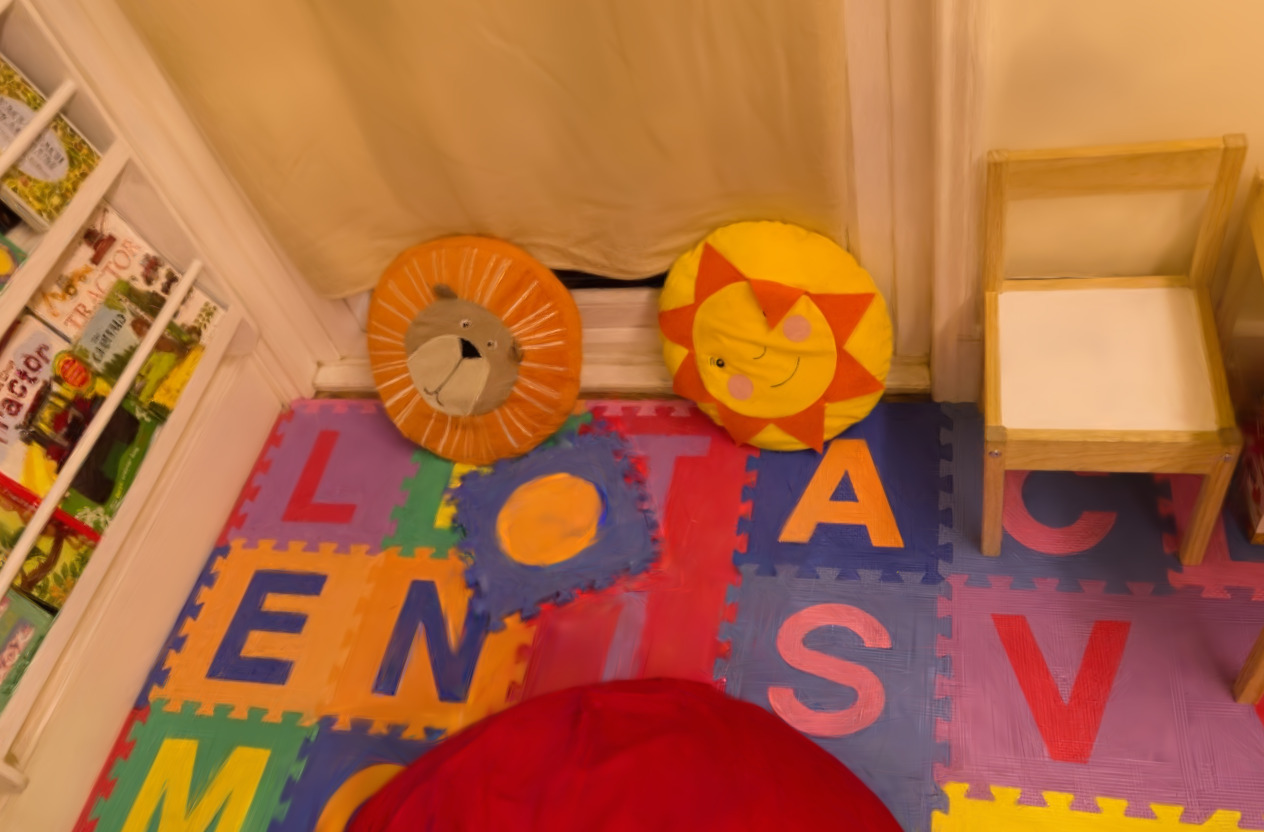}&
		\includegraphics[width=0.30\textwidth]{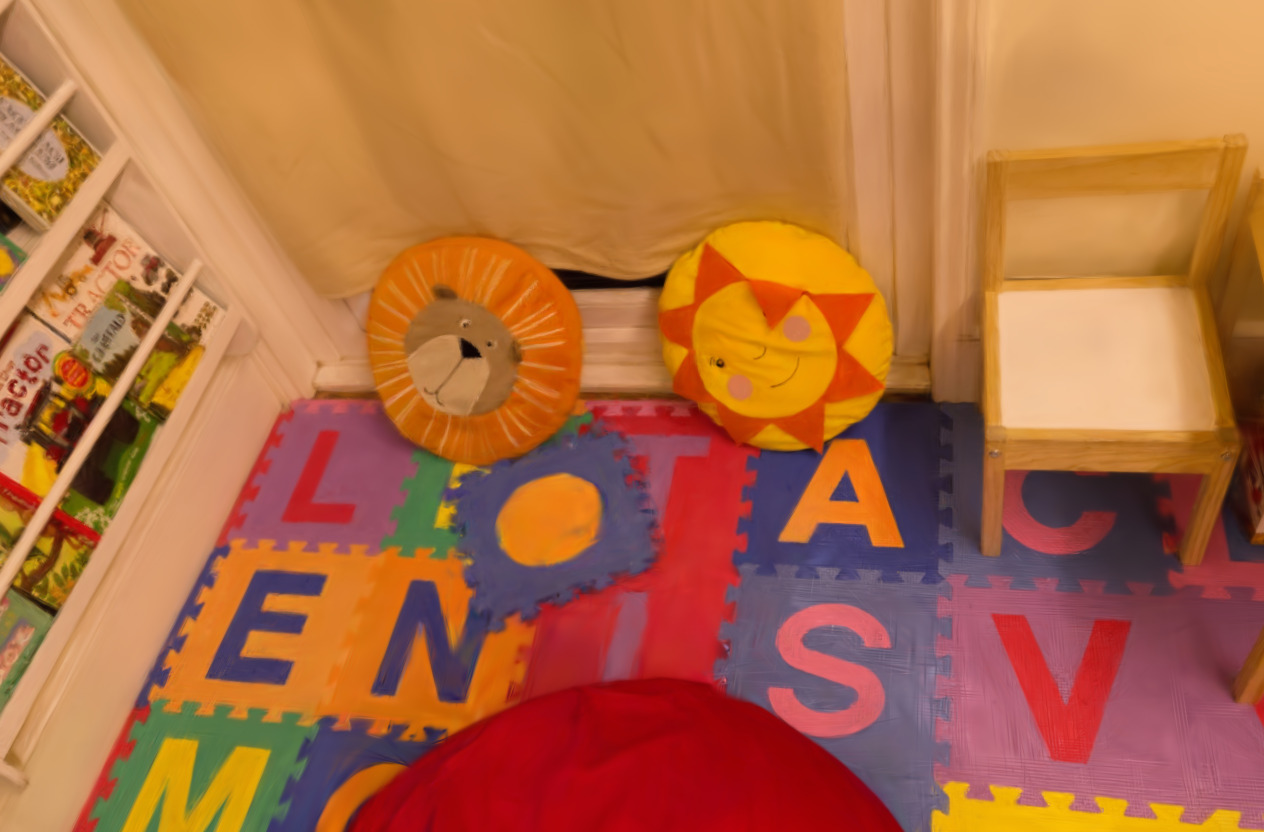}\\
		\raisebox{1.6cm}[0pt][0pt]{\rotatebox[origin=c]{90}{Drjohnson}} &
		\includegraphics[width=0.30\textwidth]{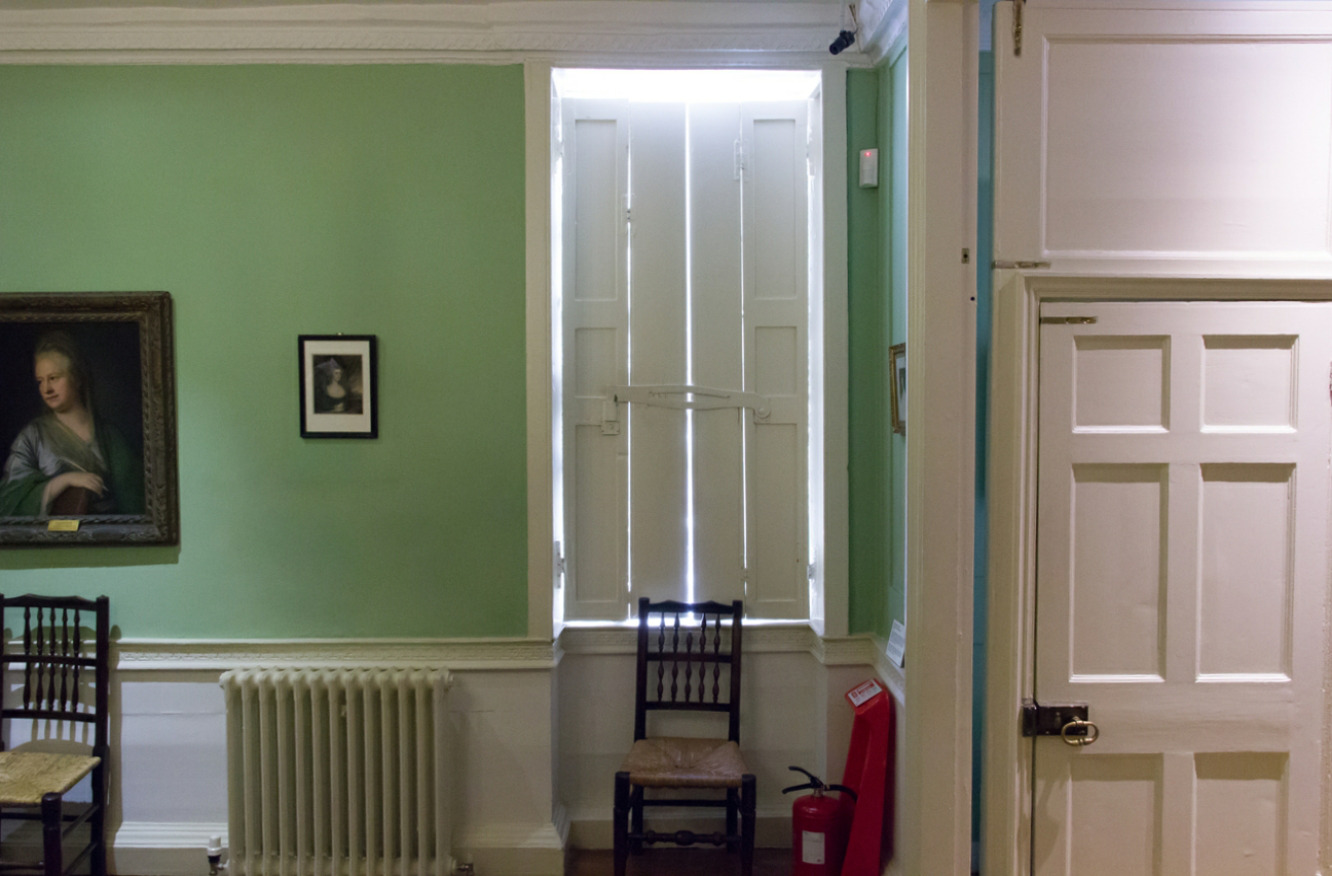}&
		\includegraphics[width=0.30\textwidth]{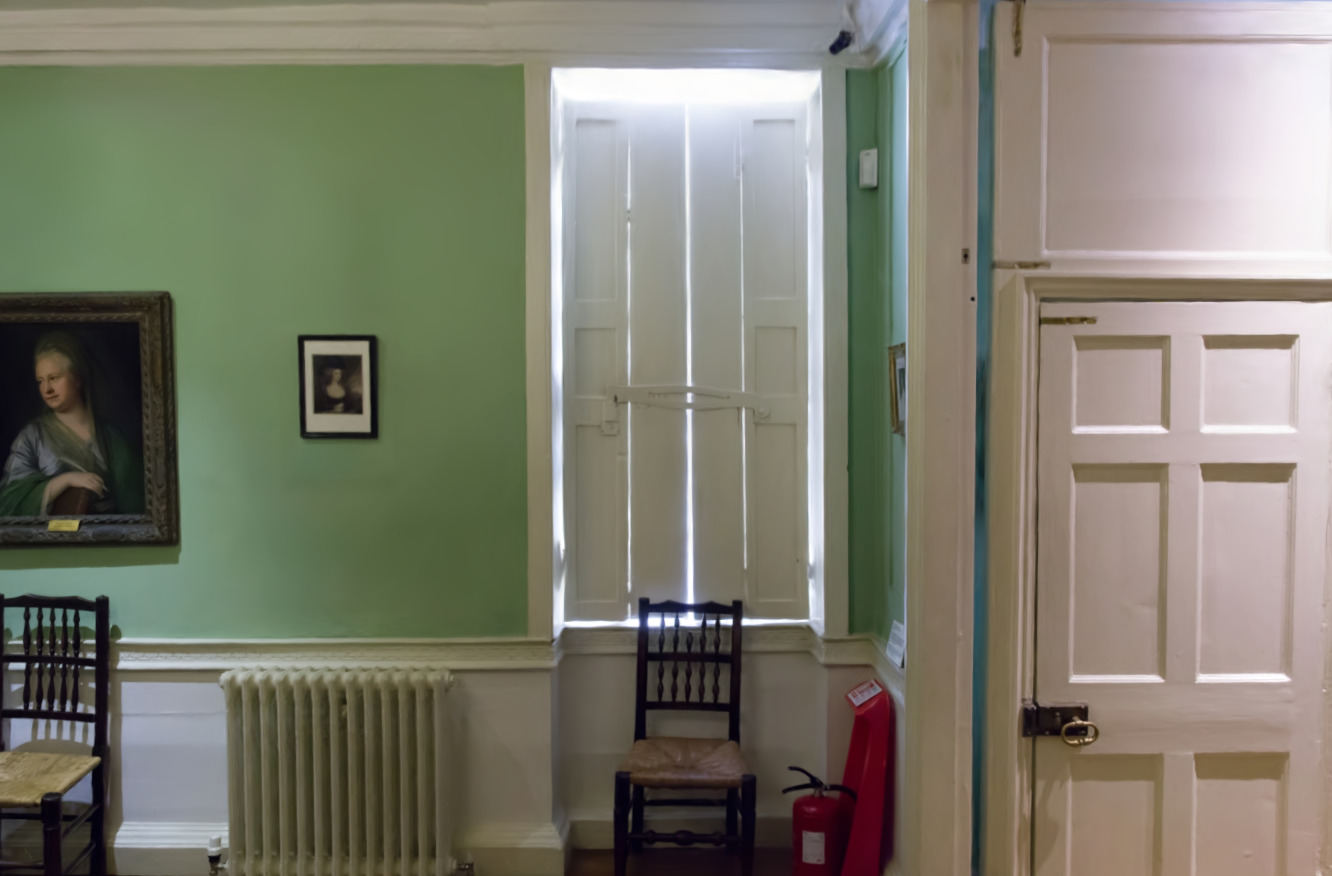}&
		\includegraphics[width=0.30\textwidth]{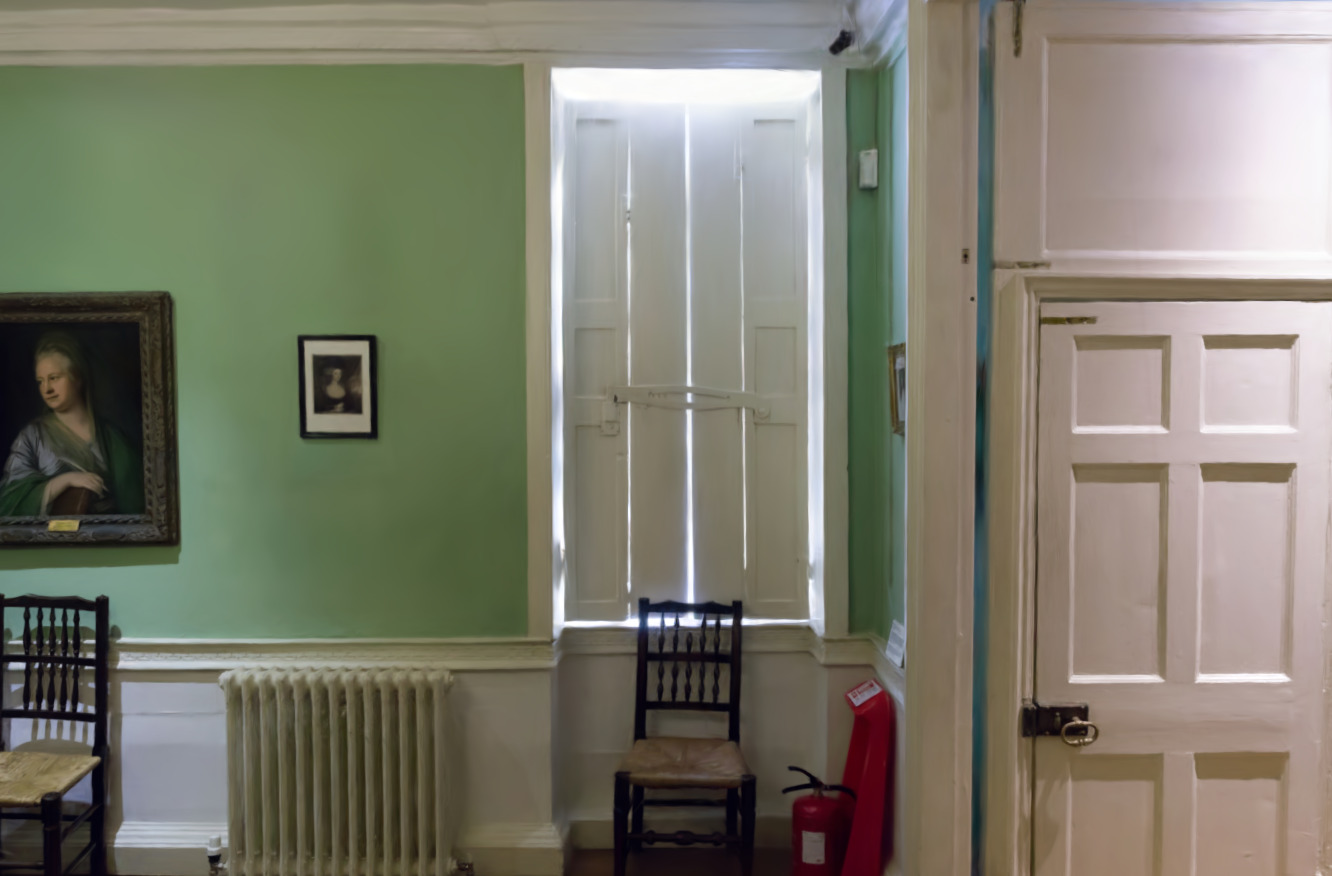}\\
    & Ground Truth & Baseline & Compressed
    \end{tabular}
    \caption{Random test views for each scene from Deep Blending~\cite{hedman_deep_2018}} 
    \label{fig:example-deep}
\end{figure*}

\begin{figure*}
\centering
\begin{tabular}{cccc}
		  \raisebox{1.5cm}[0pt][0pt]{\rotatebox[origin=c]{90}{Flowers}} &
		\includegraphics[width=0.30\textwidth]{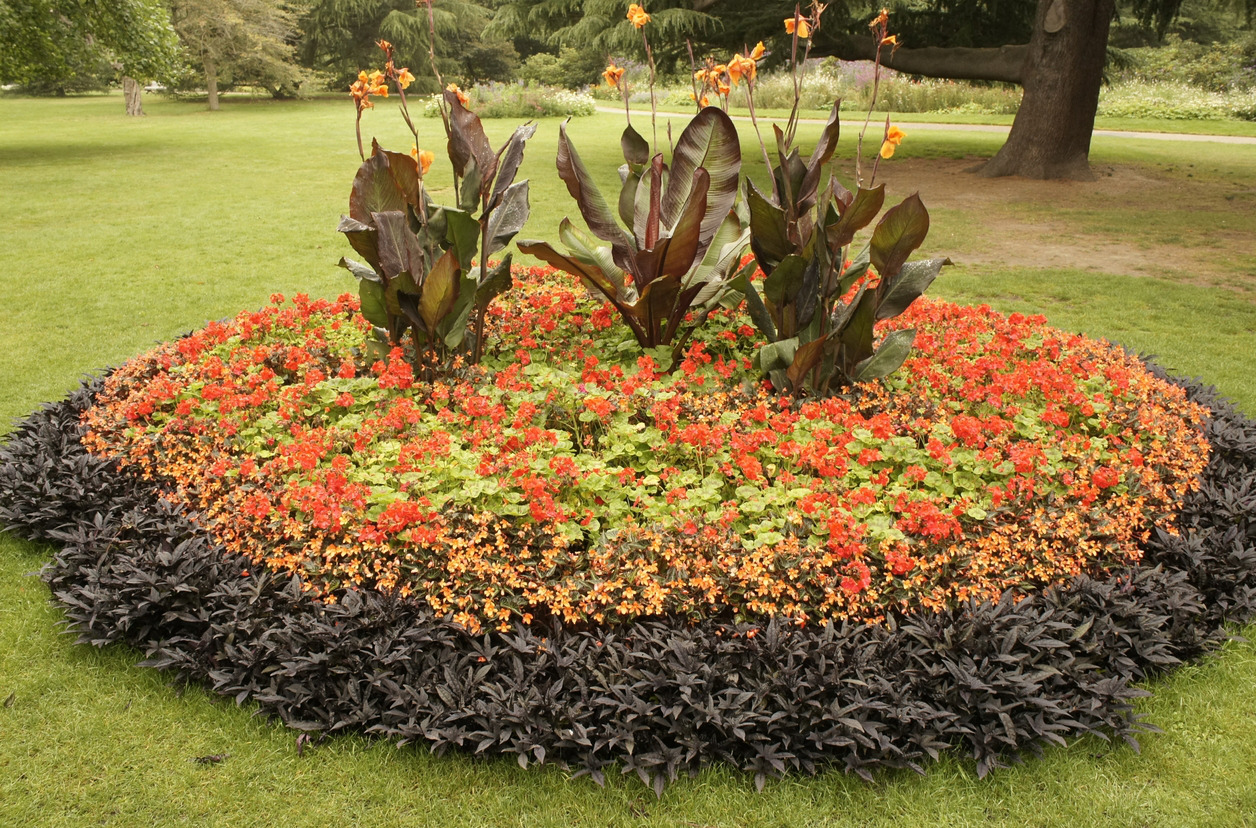}&
		\includegraphics[width=0.30\textwidth]{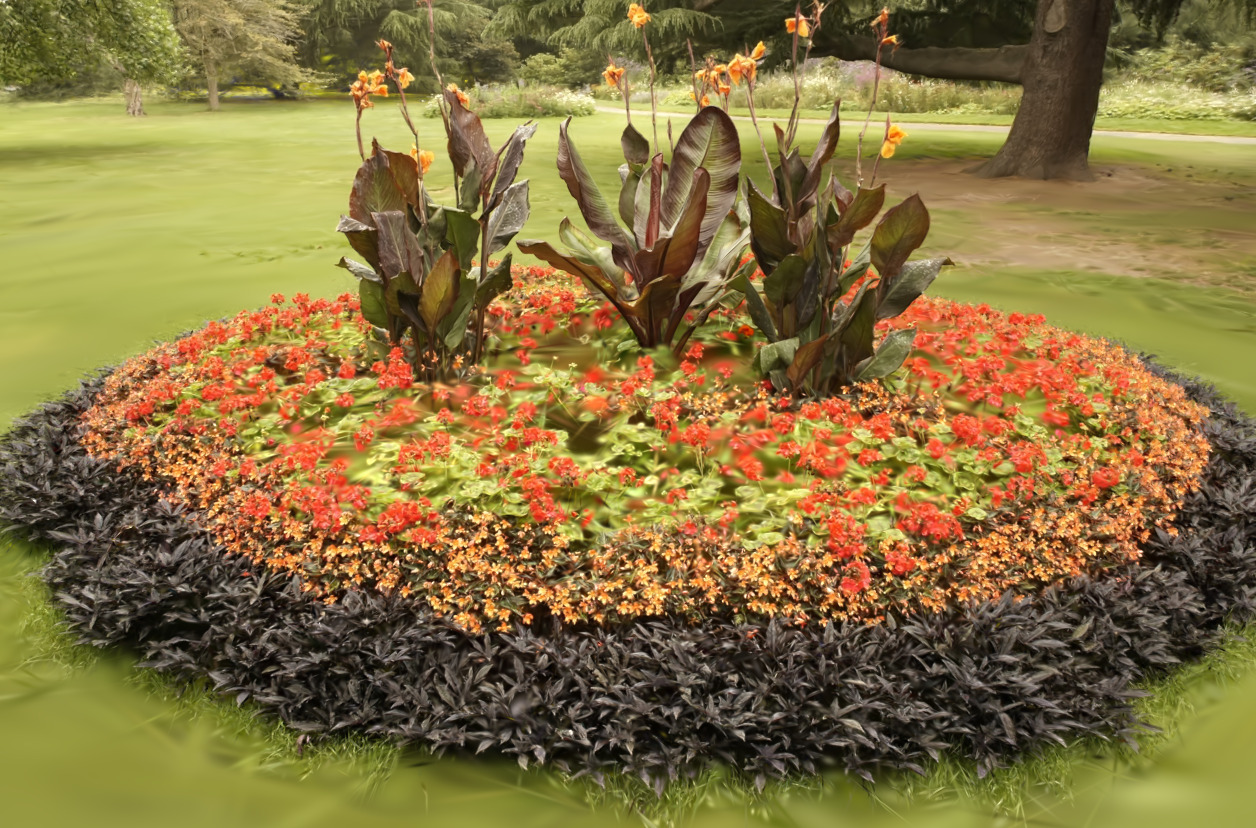}&
		\includegraphics[width=0.30\textwidth]{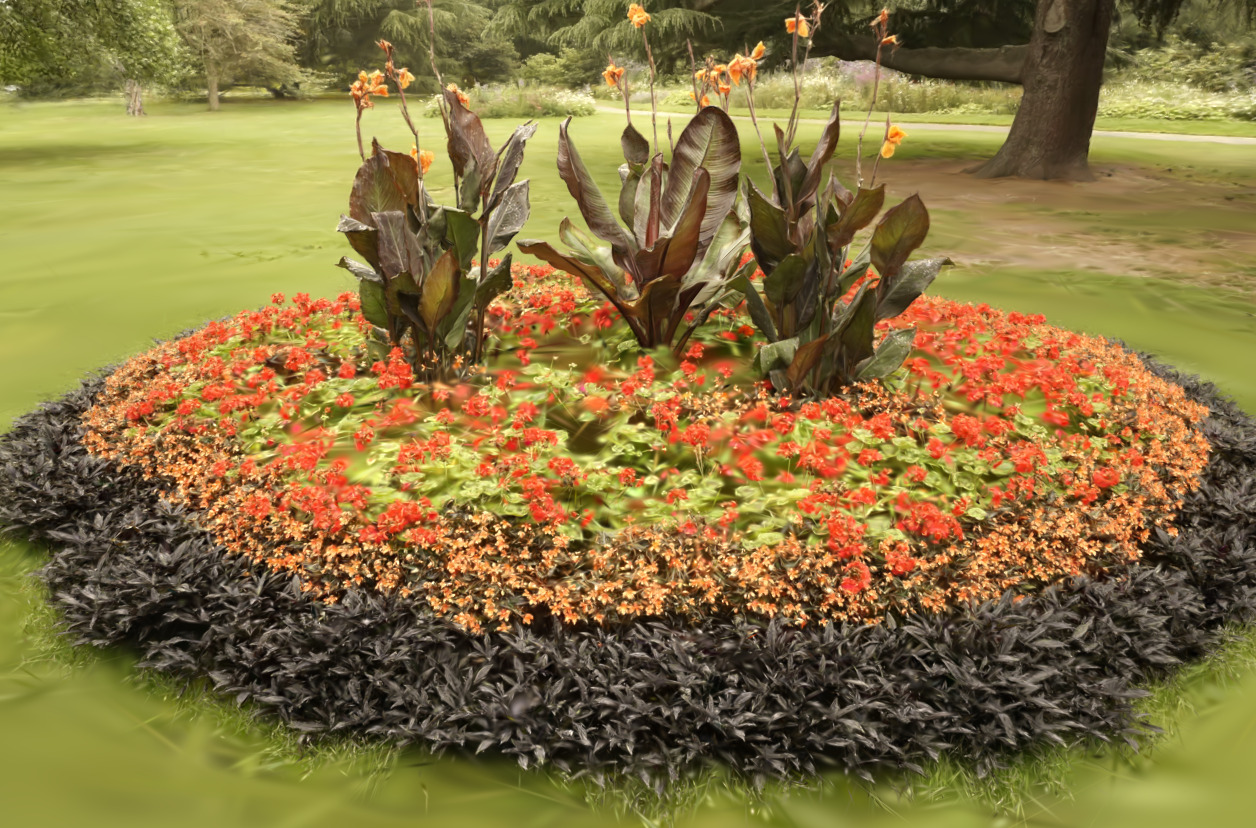}\\
		\raisebox{1.5cm}[0pt][0pt]{\rotatebox[origin=c]{90}{Garden}} &
		\includegraphics[width=0.30\textwidth]{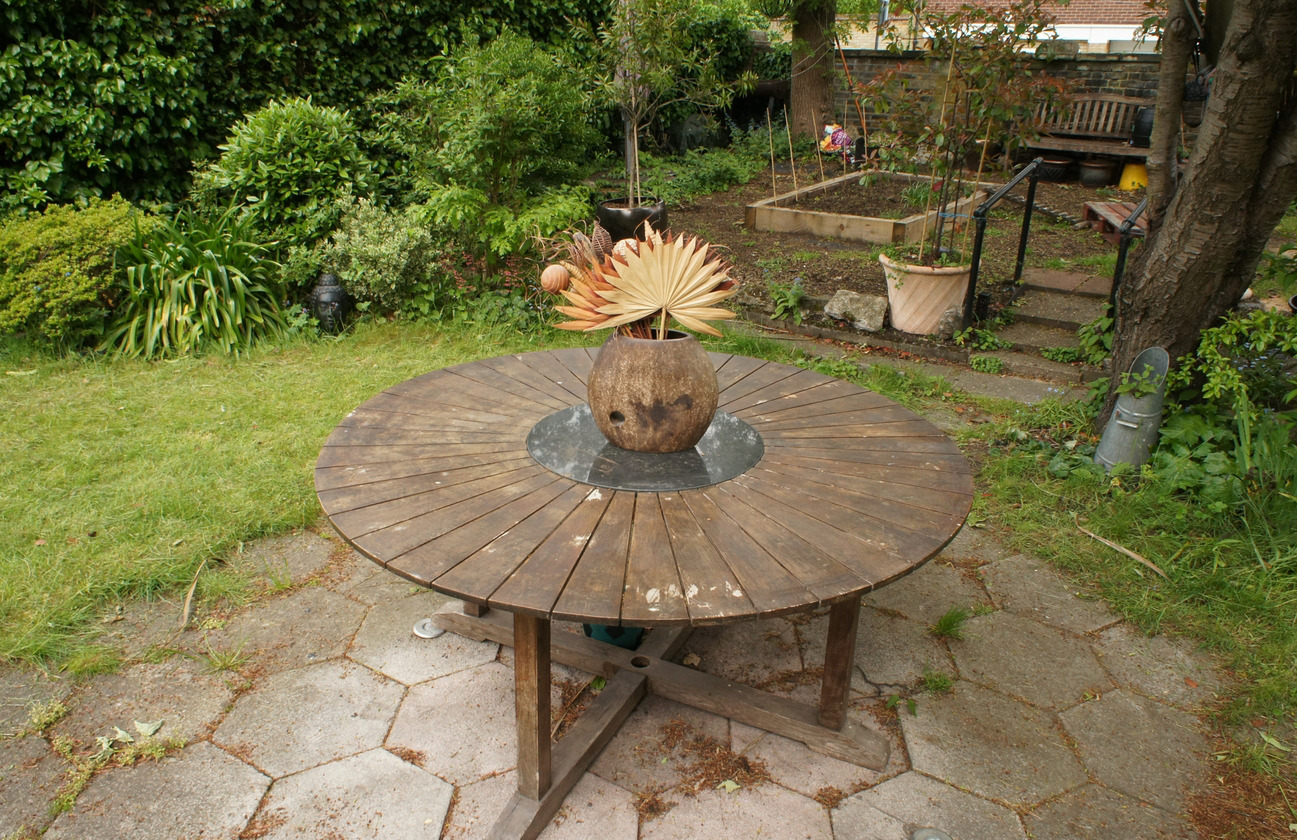}&
		\includegraphics[width=0.30\textwidth]{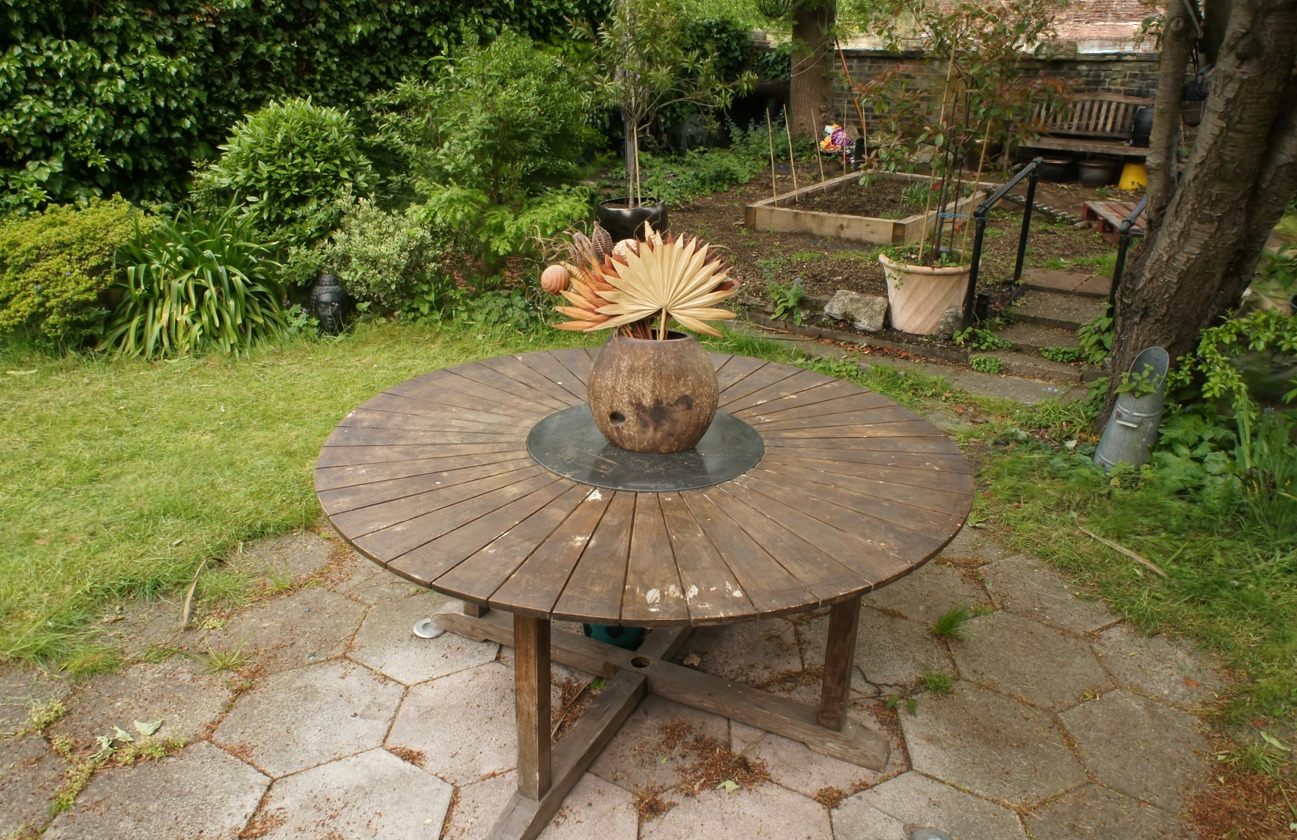}&
		\includegraphics[width=0.30\textwidth]{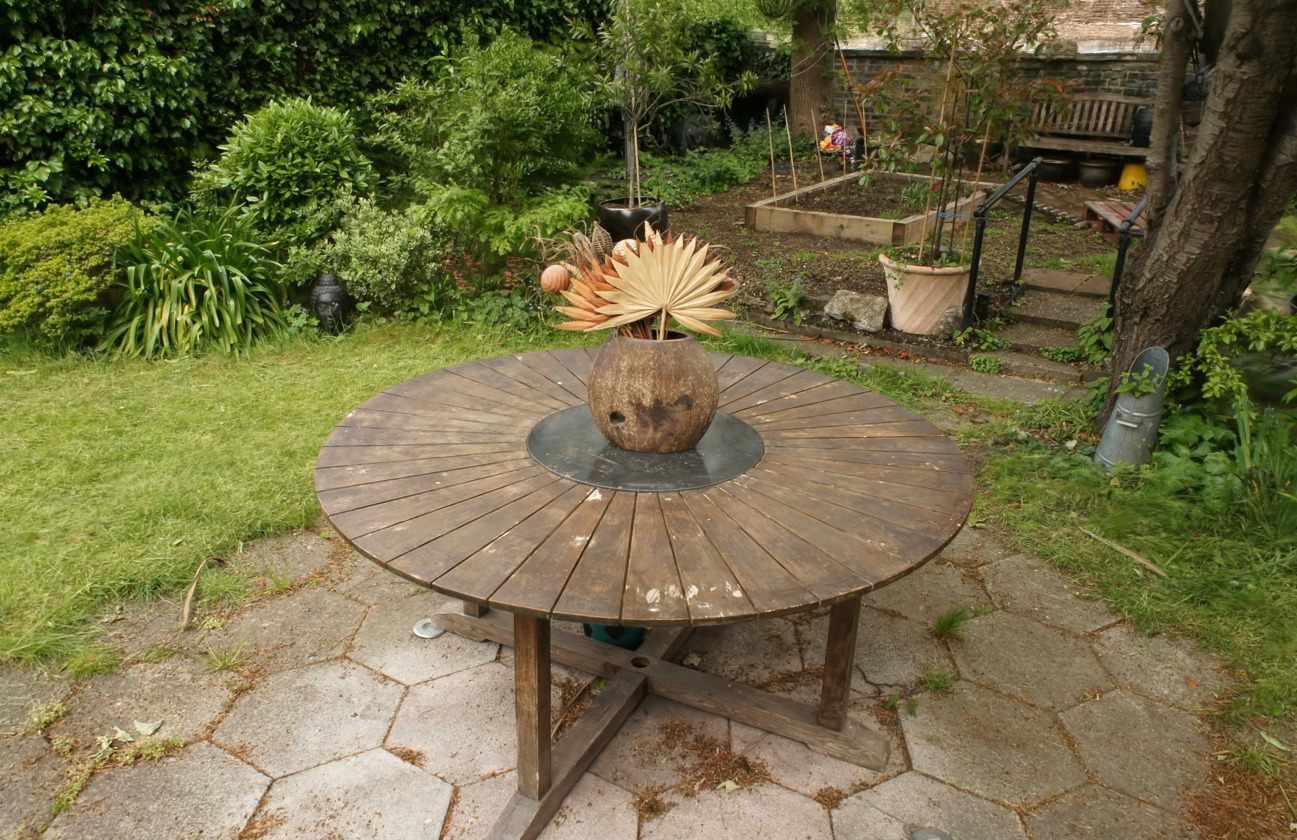}\\
		\raisebox{1.5cm}[0pt][0pt]{\rotatebox[origin=c]{90}{Stump}} &
		\includegraphics[width=0.30\textwidth]{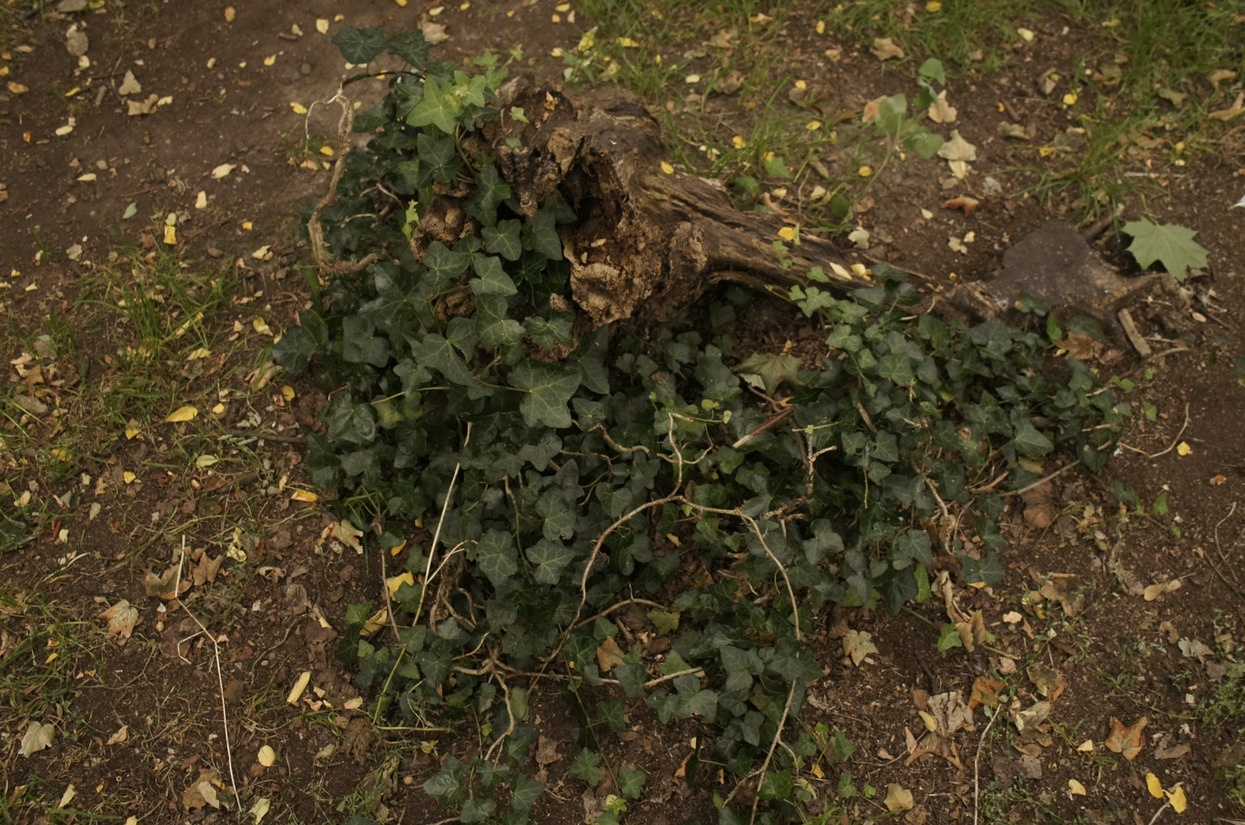}&
		\includegraphics[width=0.30\textwidth]{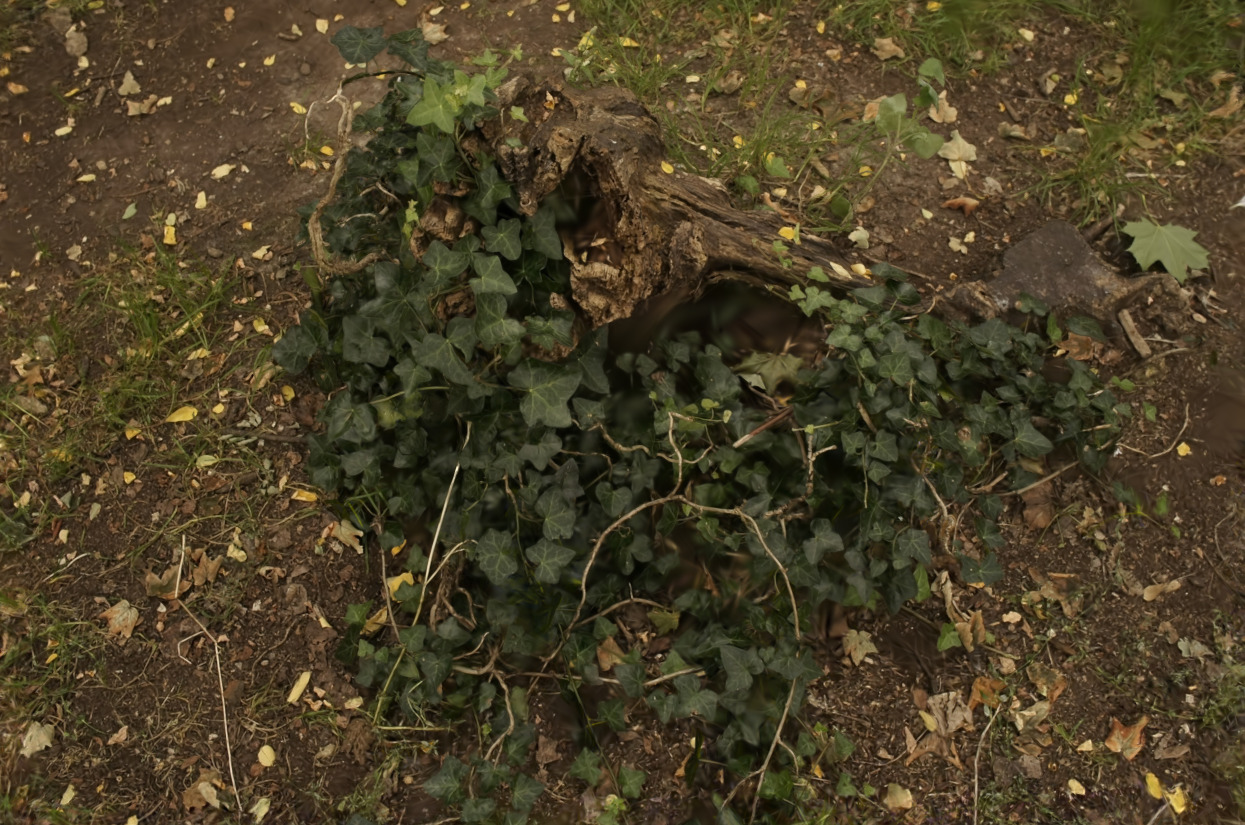}&
		\includegraphics[width=0.30\textwidth]{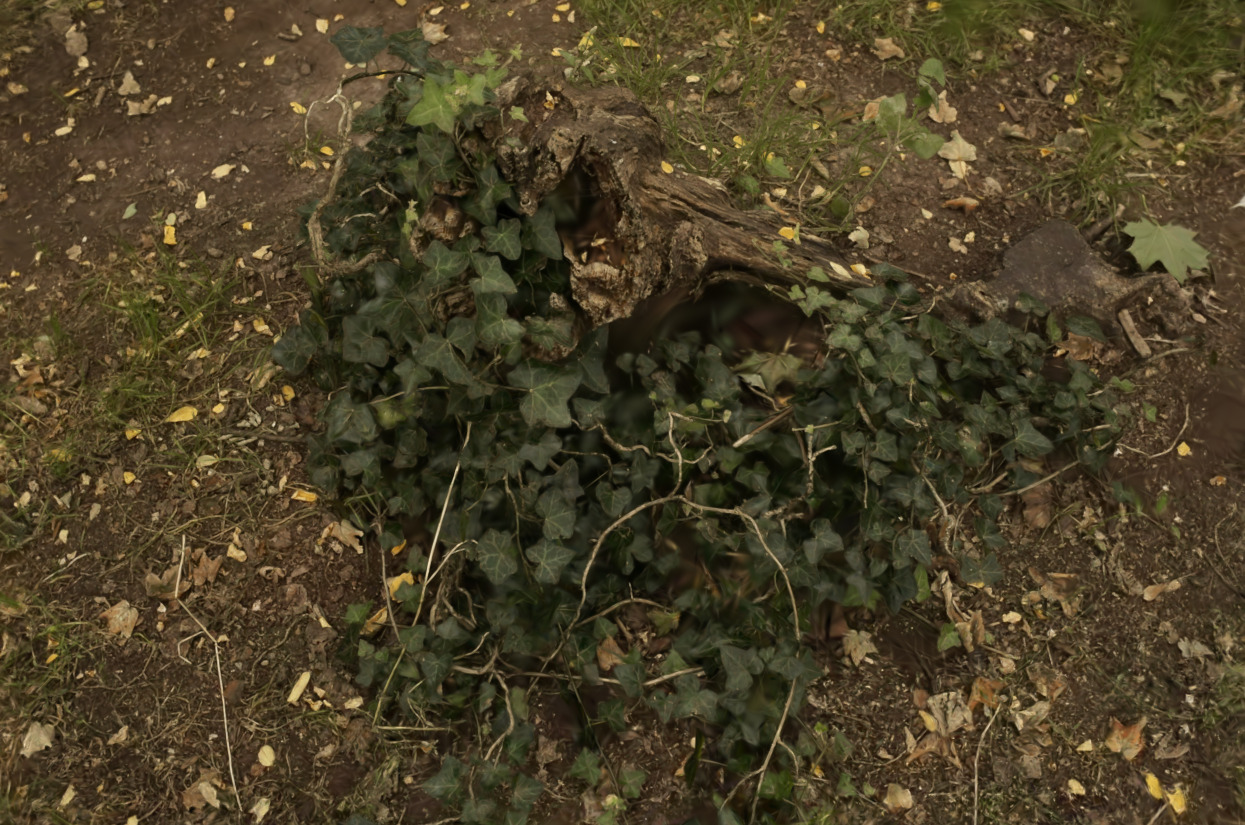}\\
		\raisebox{1.5cm}[0pt][0pt]{\rotatebox[origin=c]{90}{Treehill}} &
		\includegraphics[width=0.30\textwidth]{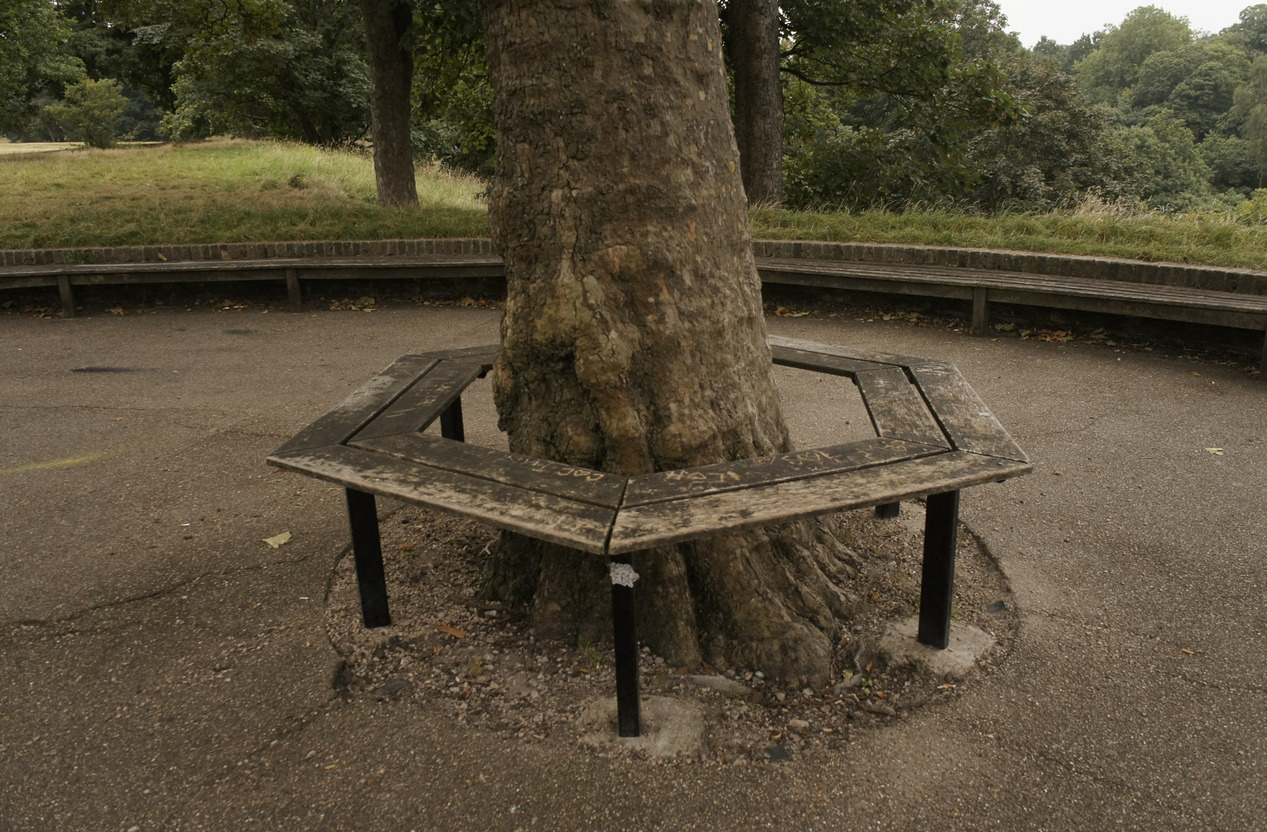}&
		\includegraphics[width=0.30\textwidth]{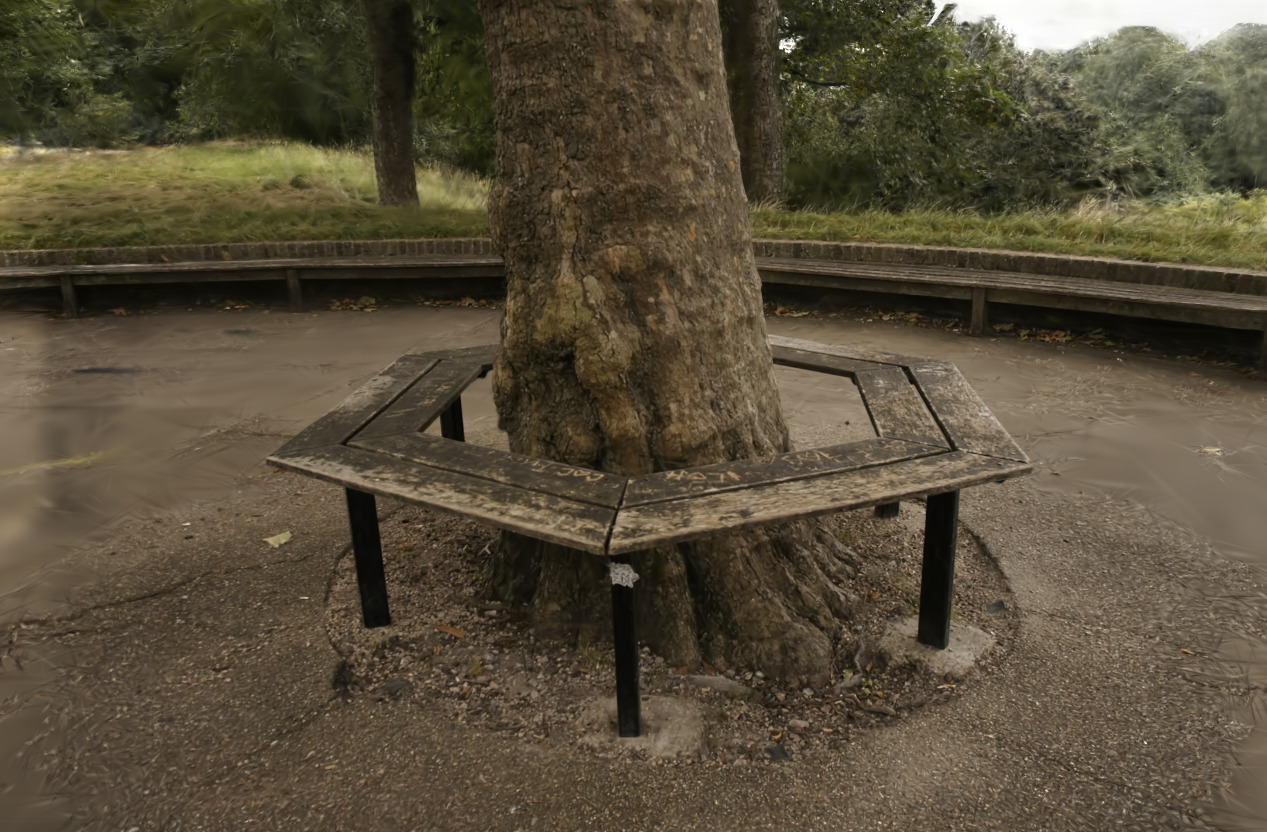}&
		\includegraphics[width=0.30\textwidth]{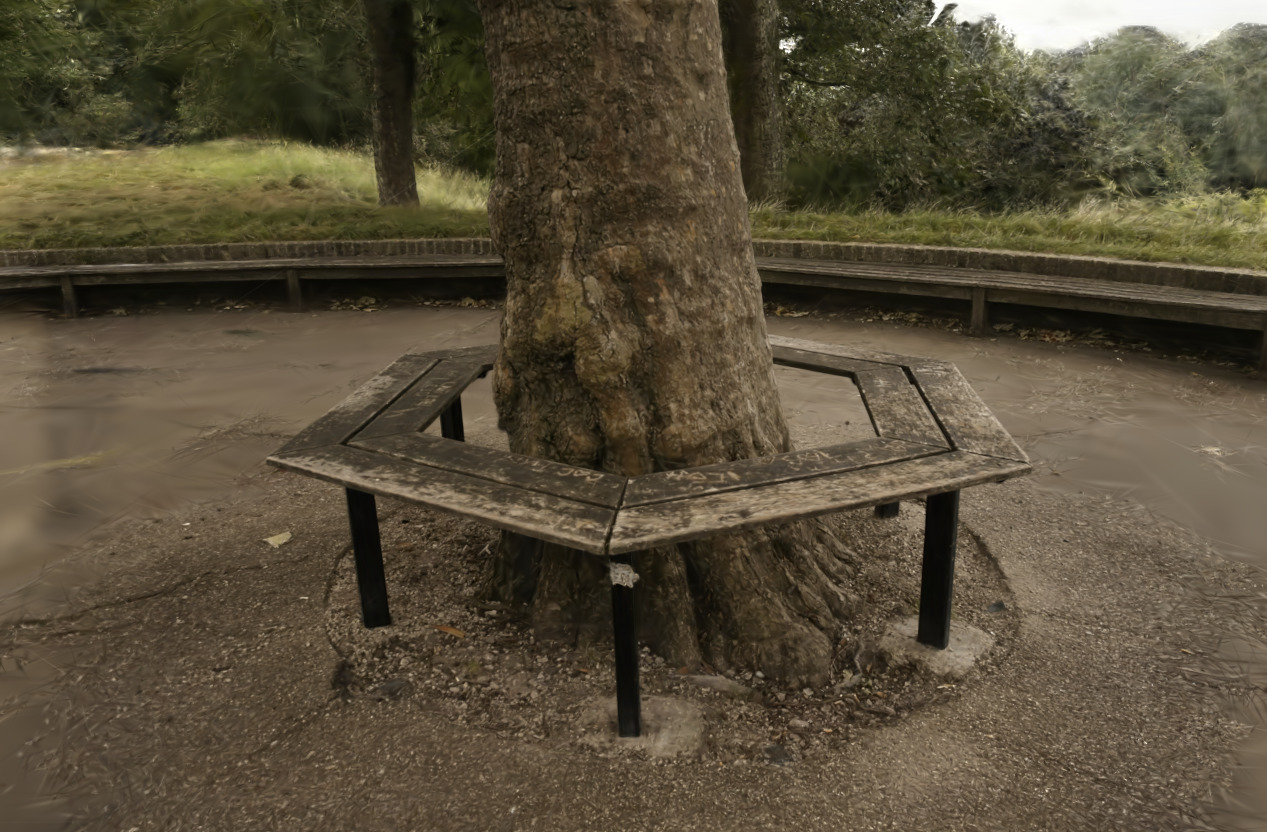}\\
		\raisebox{1.5cm}[0pt][0pt]{\rotatebox[origin=c]{90}{Room}} &
		\includegraphics[width=0.30\textwidth]{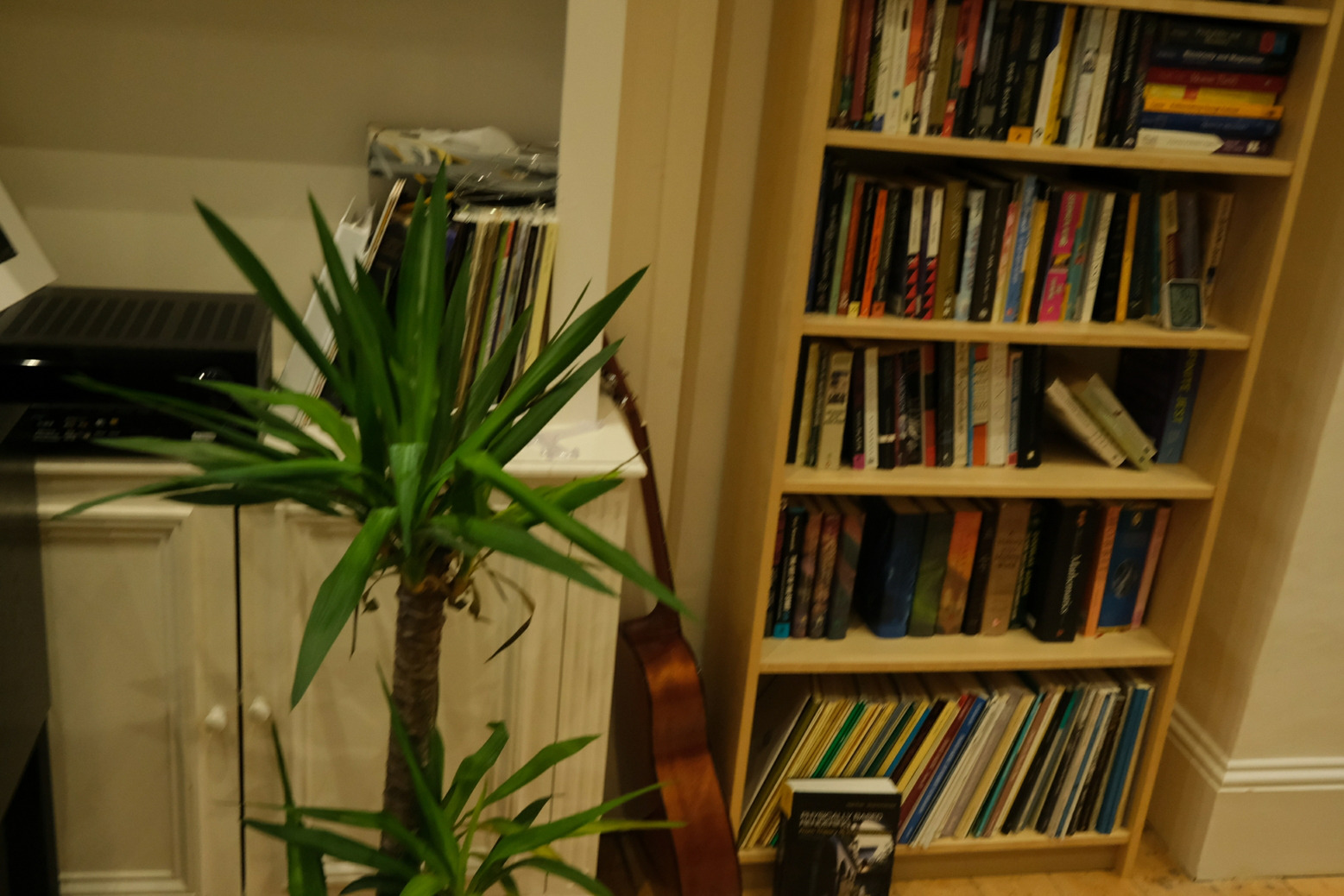}&
		\includegraphics[width=0.30\textwidth]{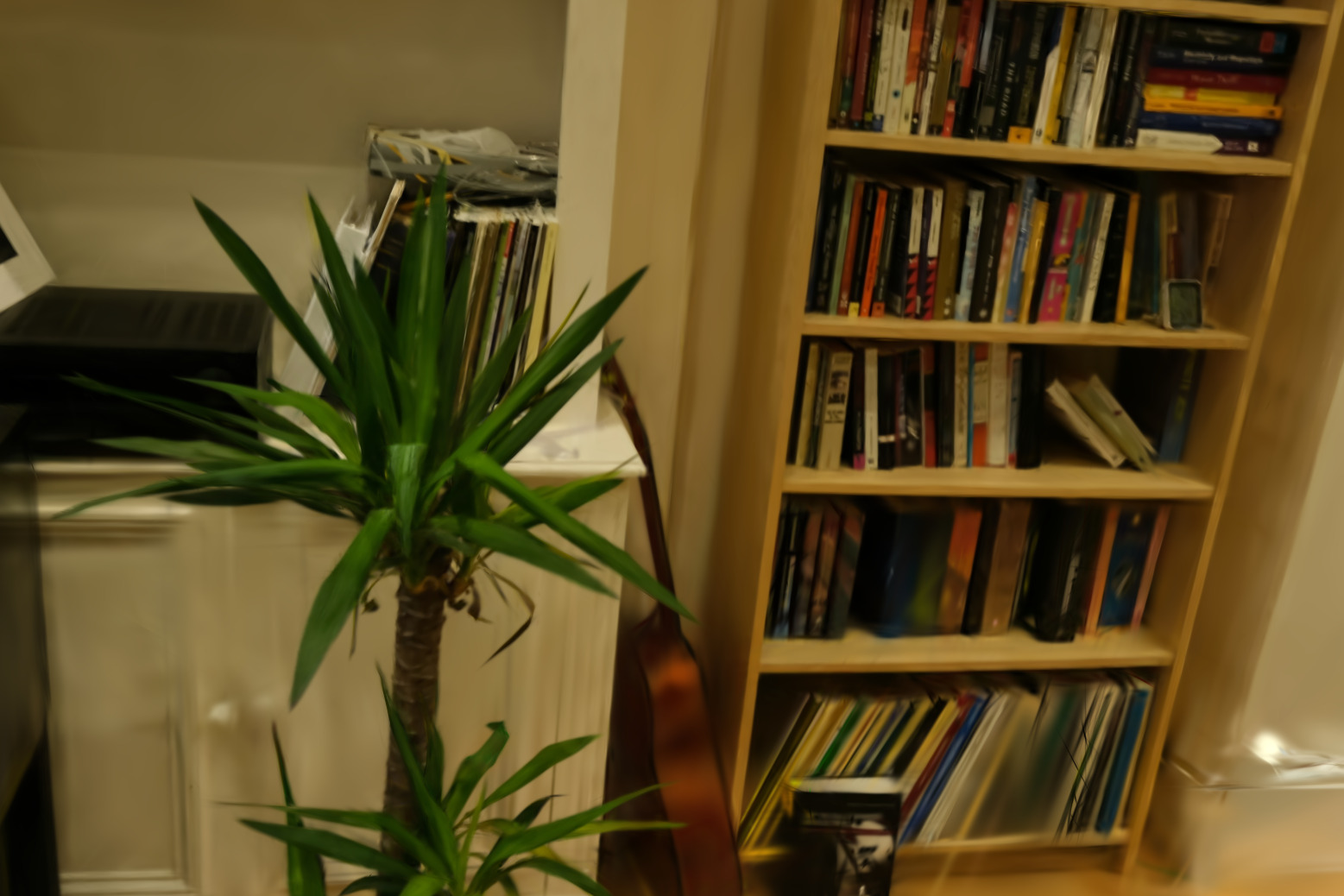}&
		\includegraphics[width=0.30\textwidth]{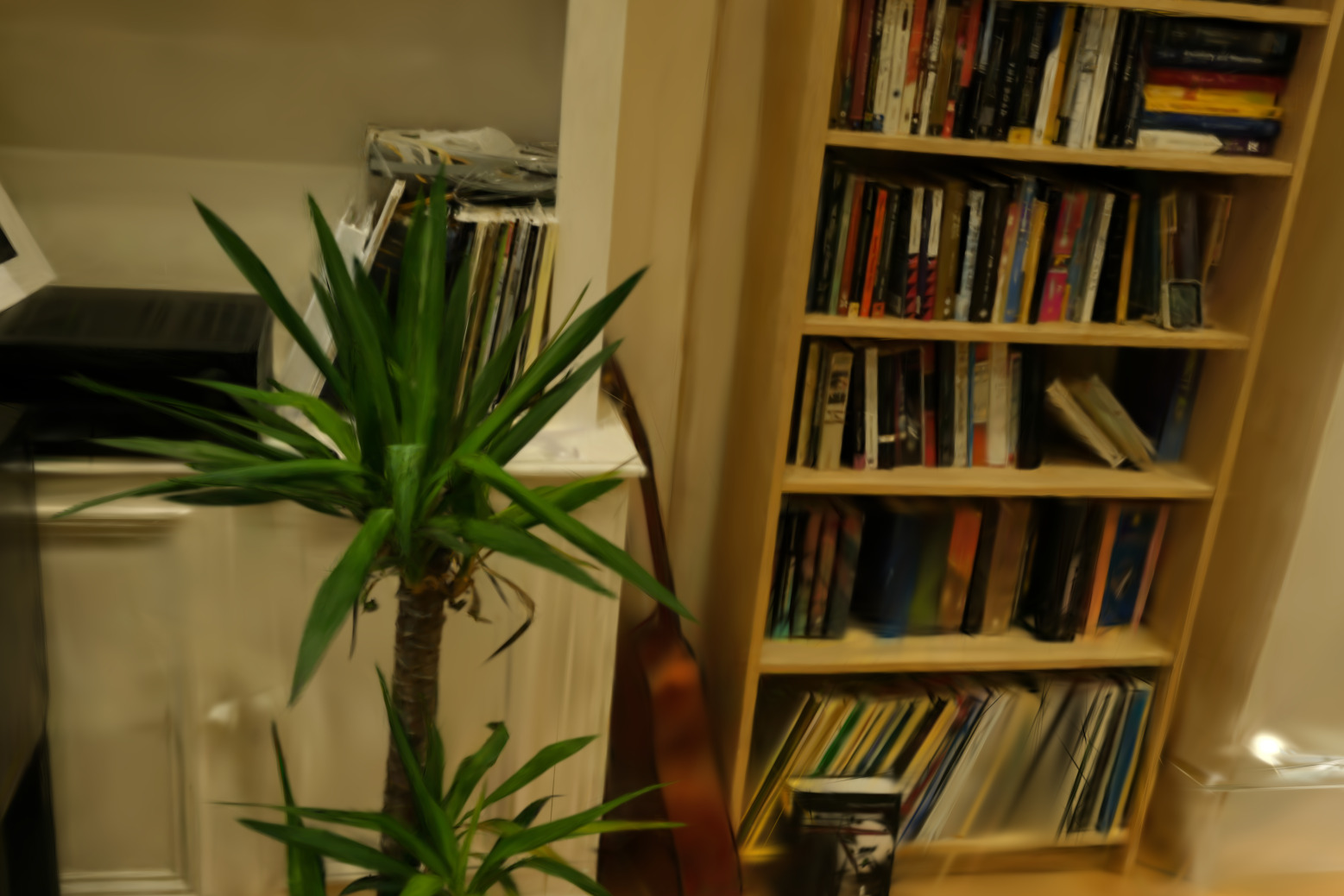}\\
        &Ground Truth & Baseline & Compressed
\end{tabular}
\caption{Random test views for each scene from Mip-NeRF360~\cite{barron_mip-nerf_2022}}
\label{fig:example-mip-1}
\end{figure*}

\begin{figure*}
\centering
\begin{tabular}{cccc}
		\raisebox{1.5cm}[0pt][0pt]{\rotatebox[origin=c]{90}{Counter}} &
		\includegraphics[width=0.30\textwidth]{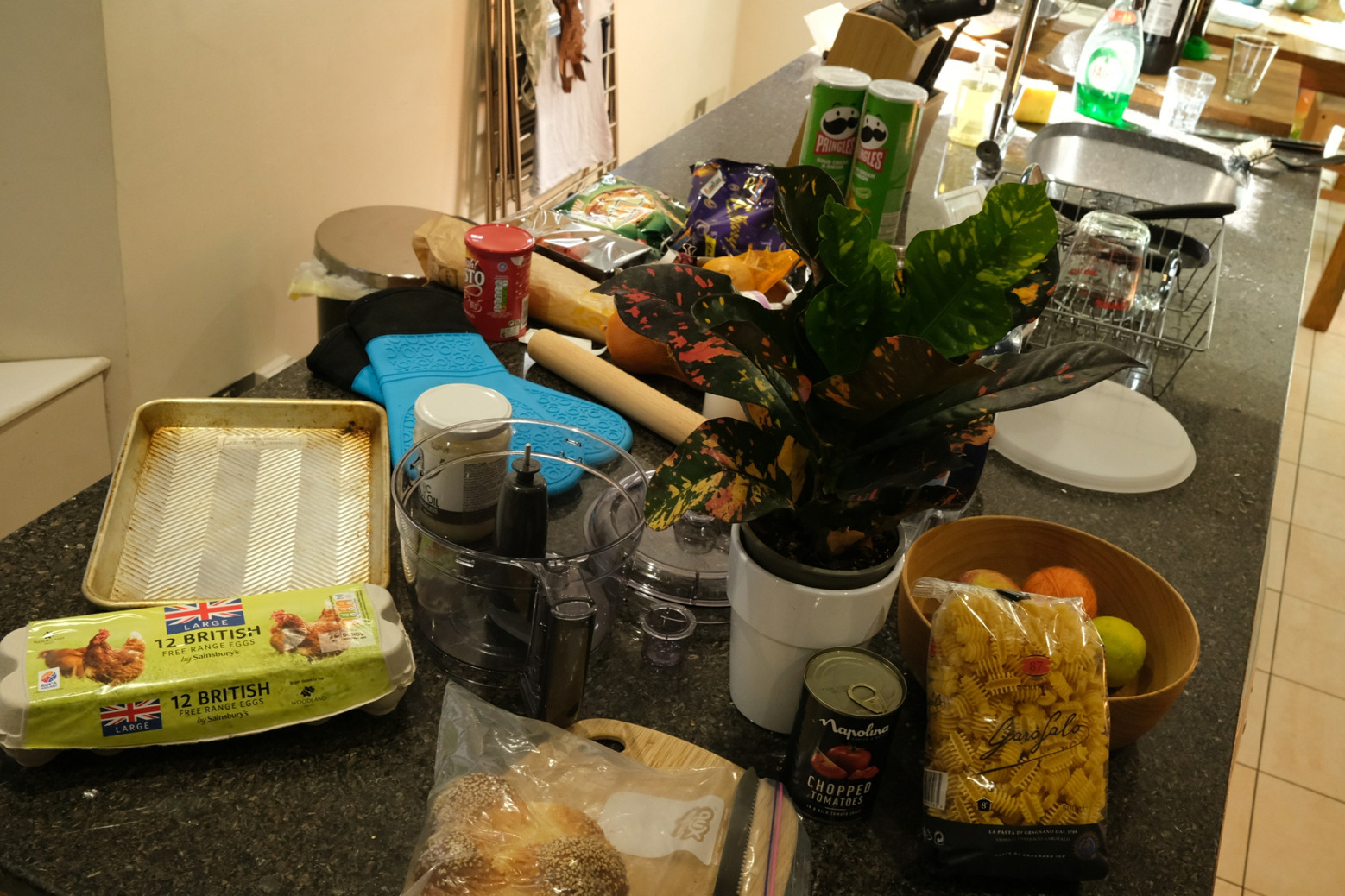}&
		\includegraphics[width=0.30\textwidth]{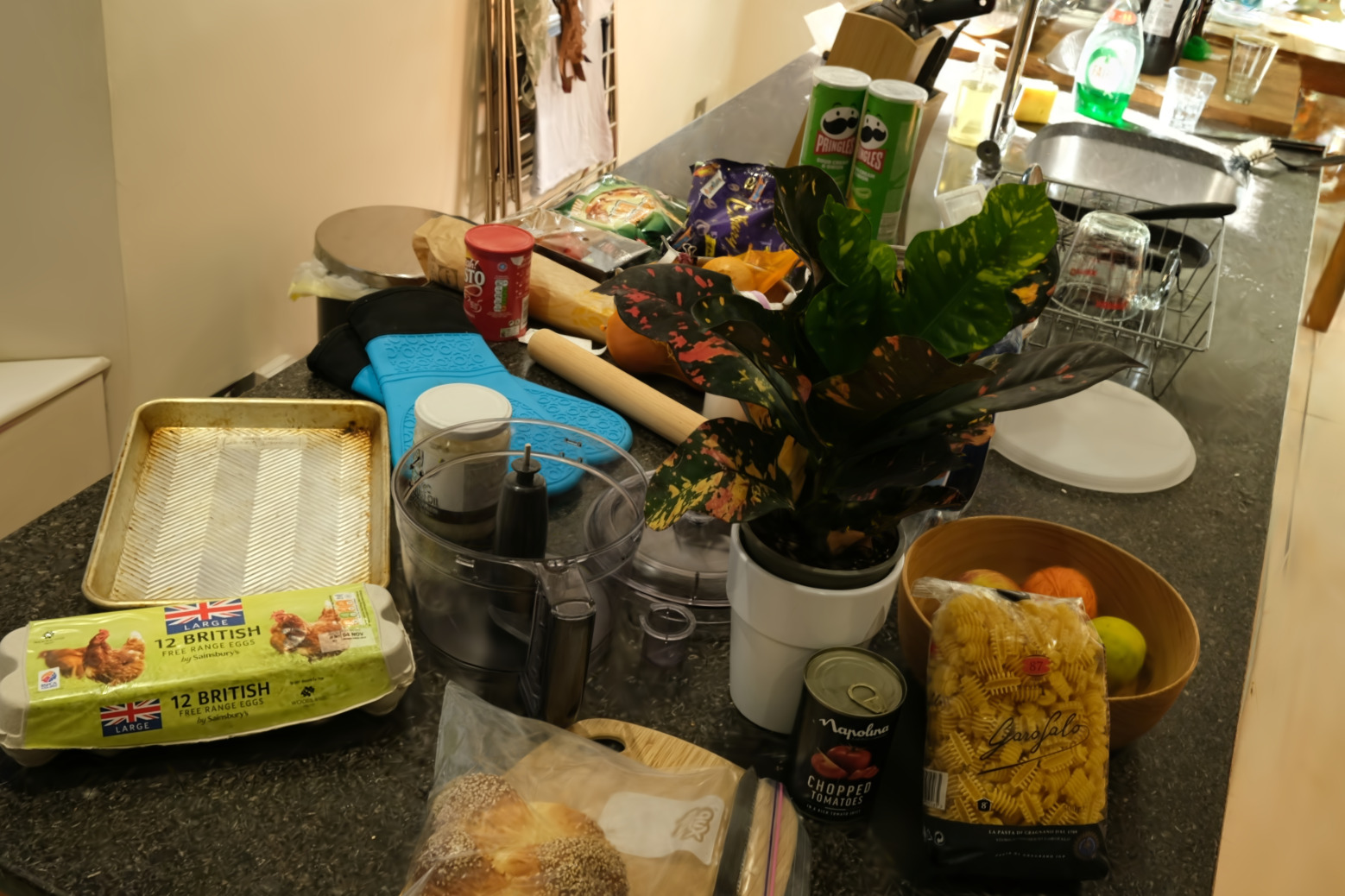}&
		\includegraphics[width=0.30\textwidth]{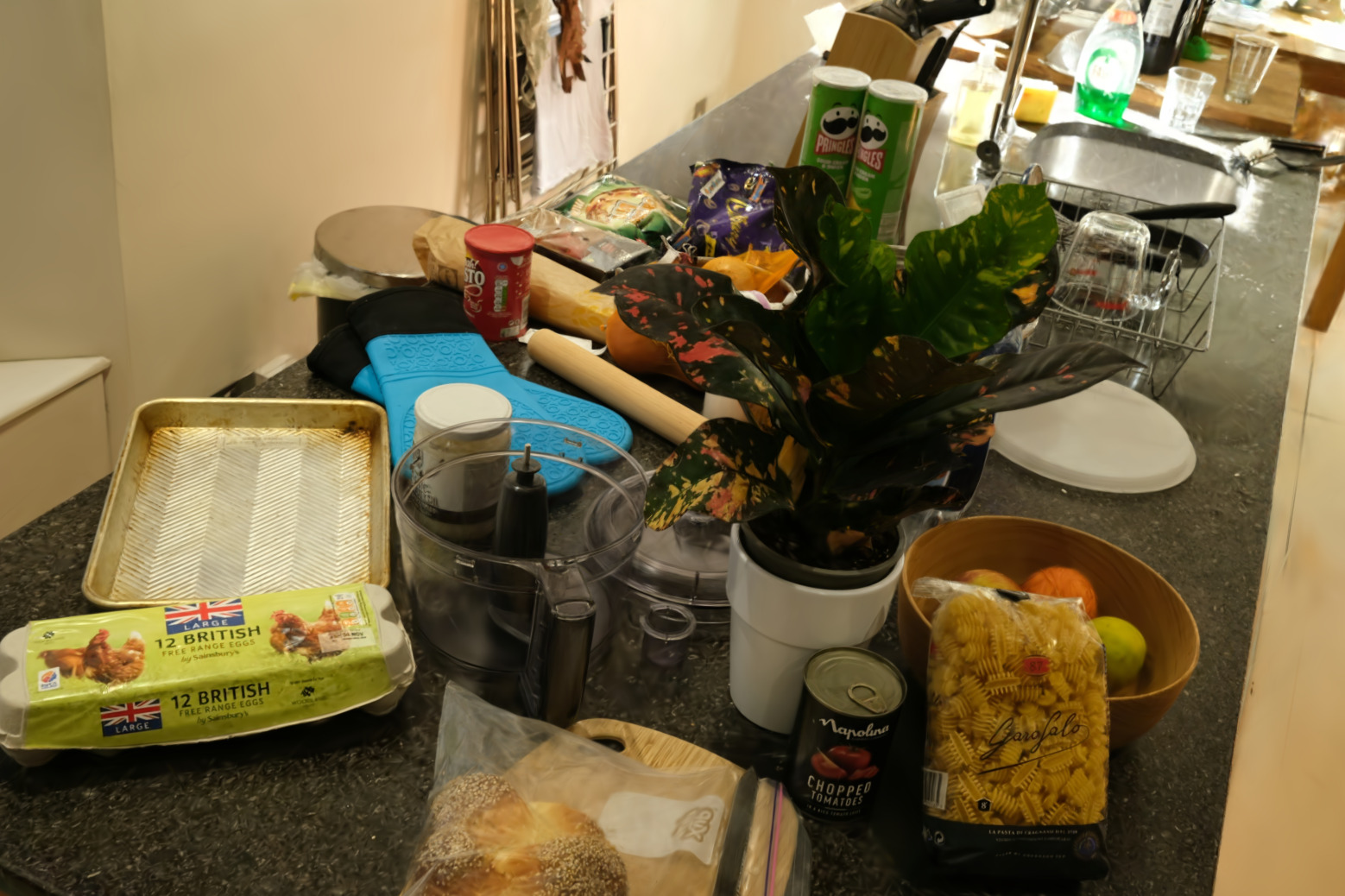}\\
		\raisebox{1.5cm}[0pt][0pt]{\rotatebox[origin=c]{90}{Kitchen}} &
		\includegraphics[width=0.30\textwidth]{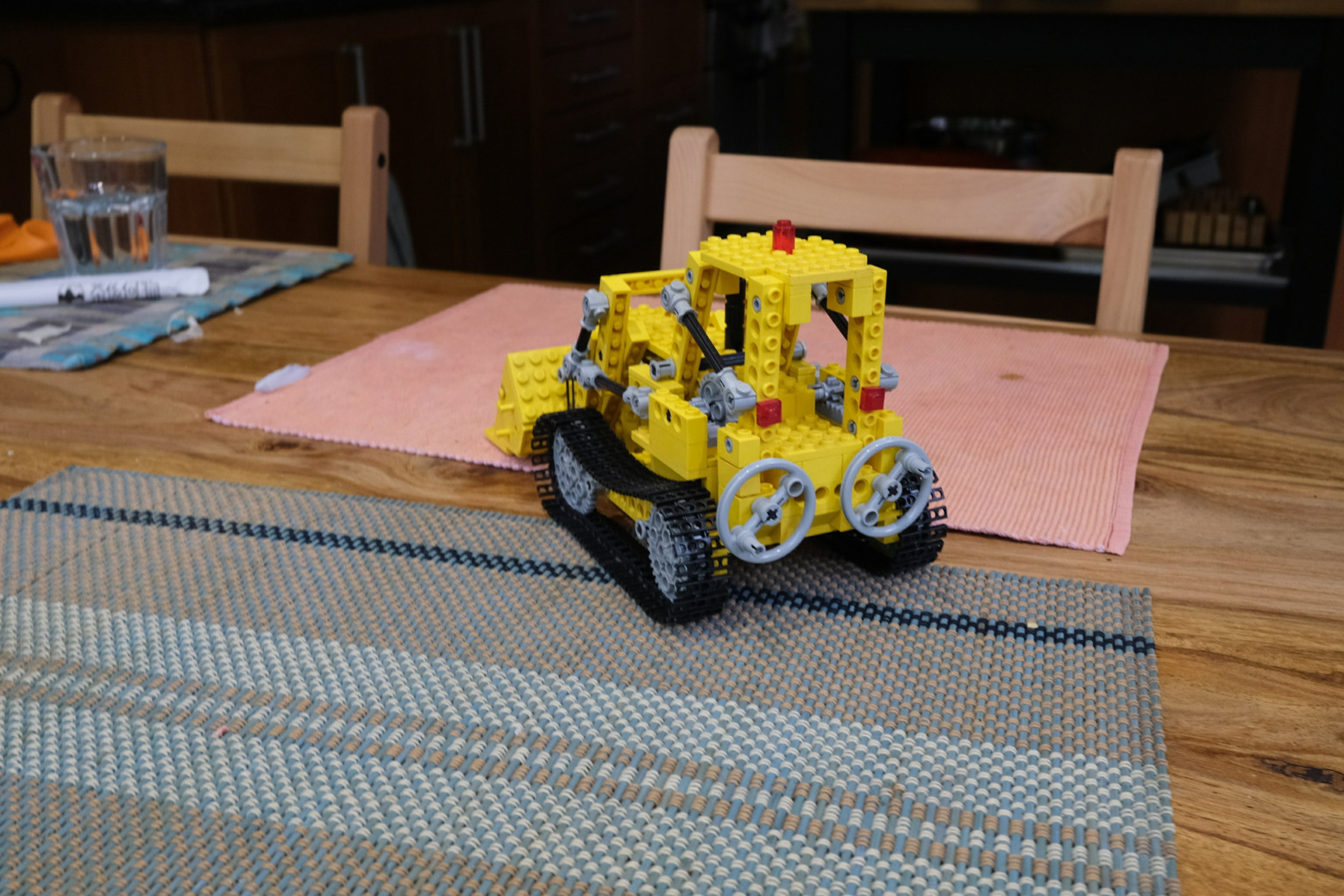}&
		\includegraphics[width=0.30\textwidth]{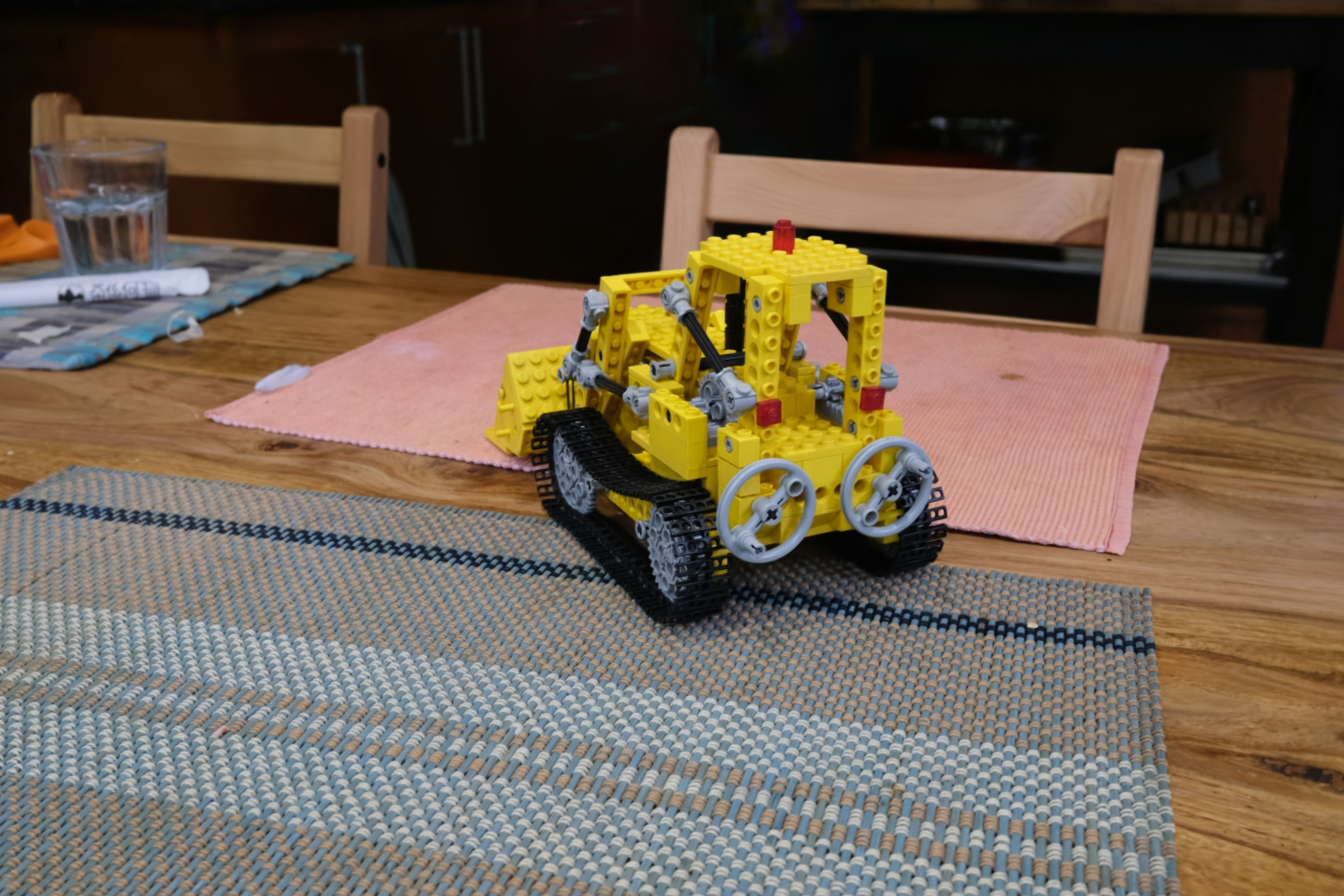}&
		\includegraphics[width=0.30\textwidth]{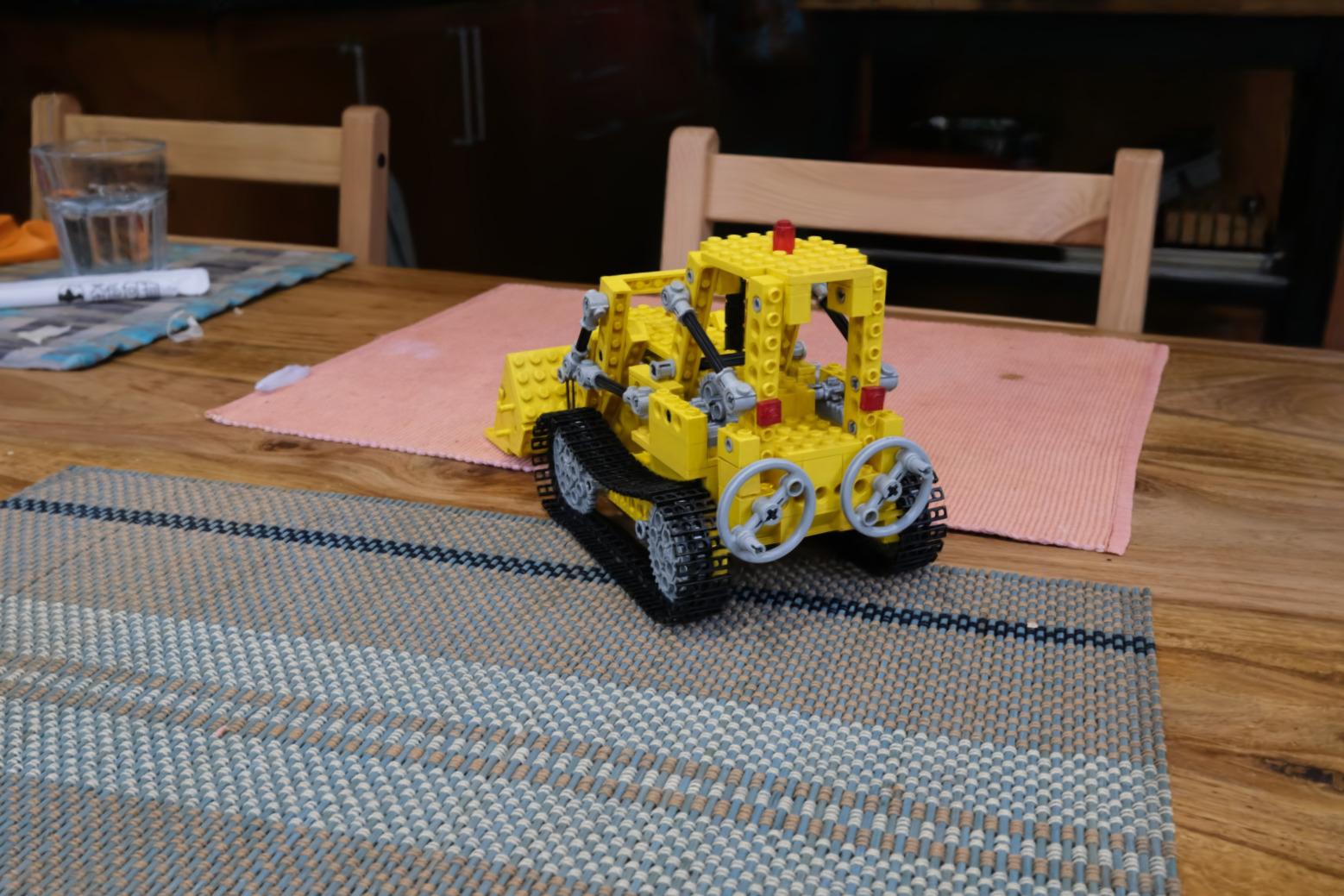}\\
		\raisebox{1.5cm}[0pt][0pt]{\rotatebox[origin=c]{90}{Bonsai}} &
		\includegraphics[width=0.30\textwidth]{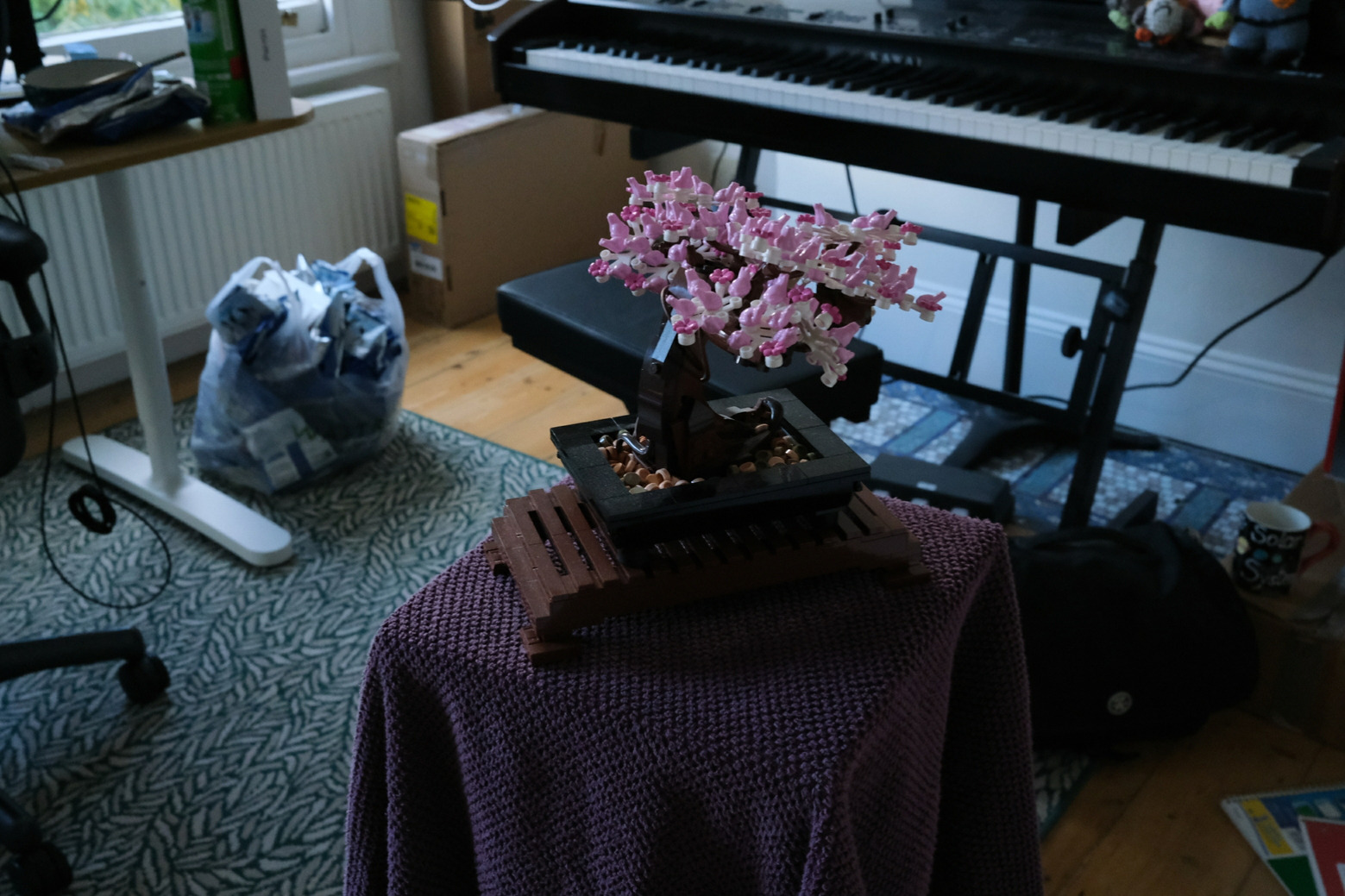}&
		\includegraphics[width=0.30\textwidth]{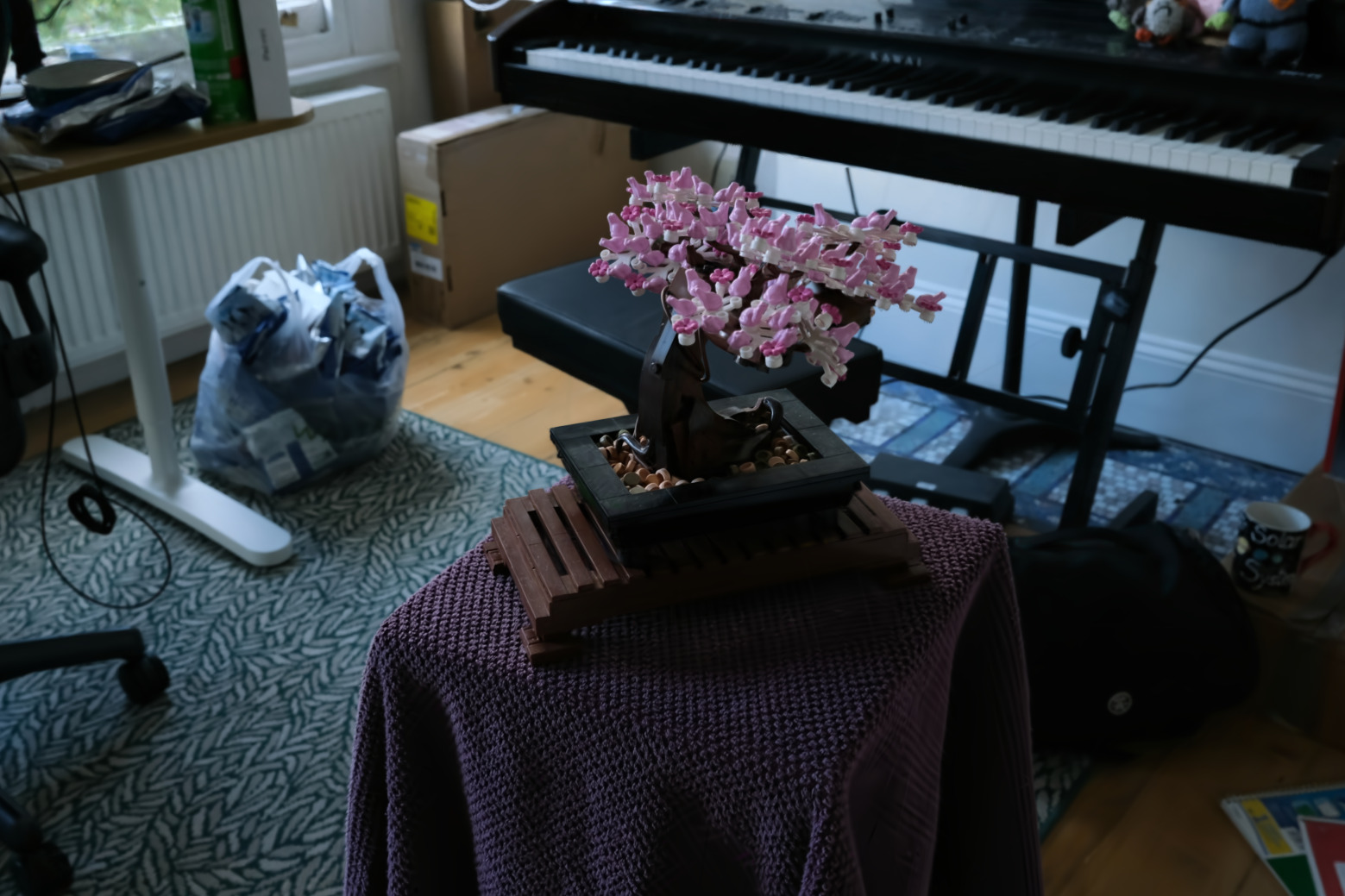}&
		\includegraphics[width=0.30\textwidth]{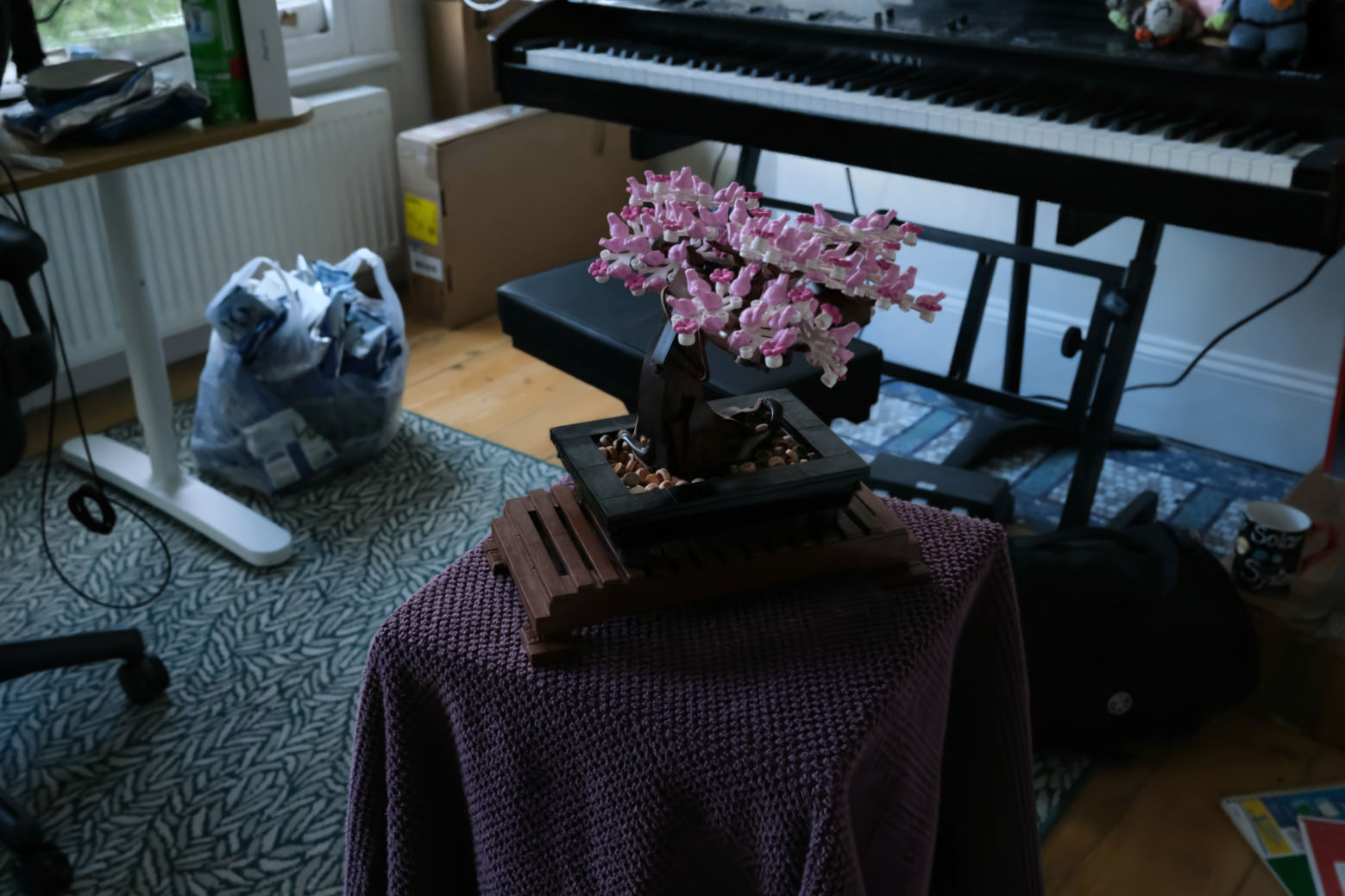}\\
		\raisebox{1.5cm}[0pt][0pt]{\rotatebox[origin=c]{90}{Bicycle}} &
		\includegraphics[width=0.30\textwidth]{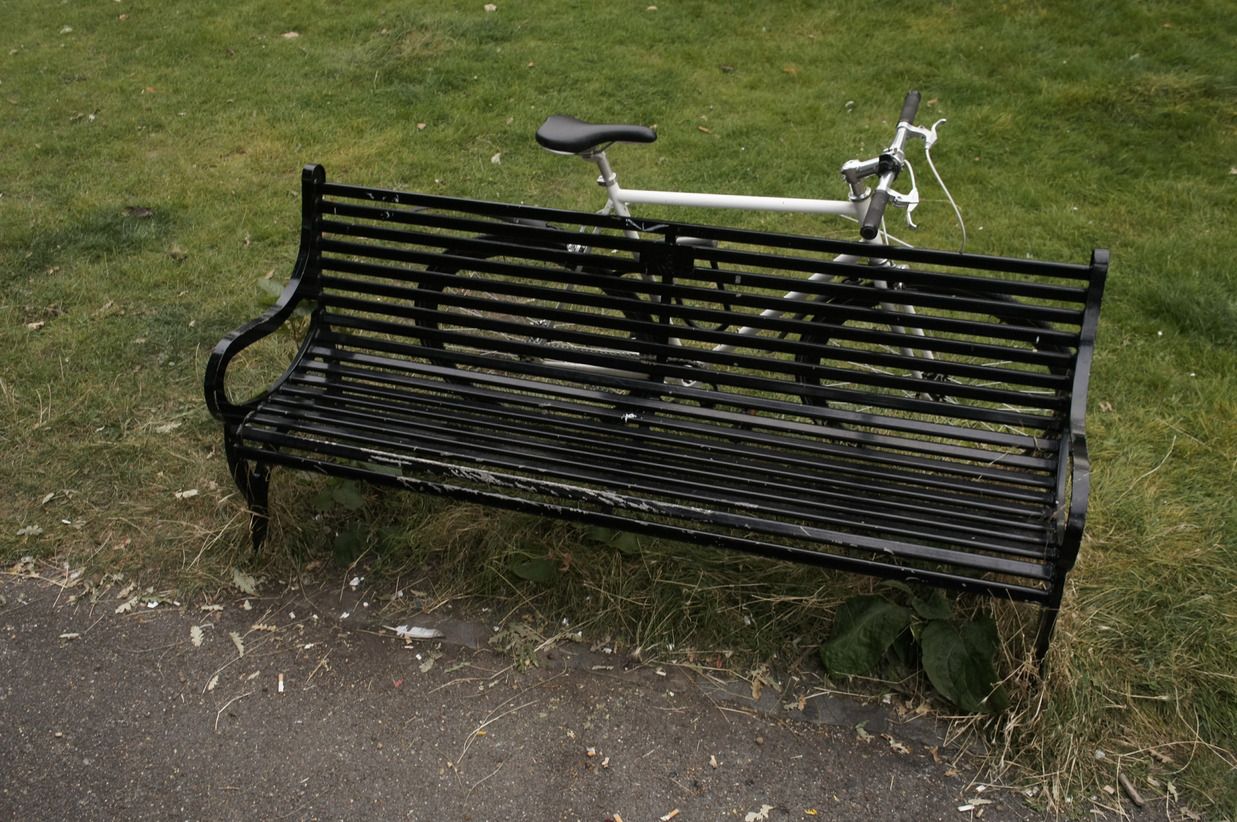}&
		\includegraphics[width=0.30\textwidth]{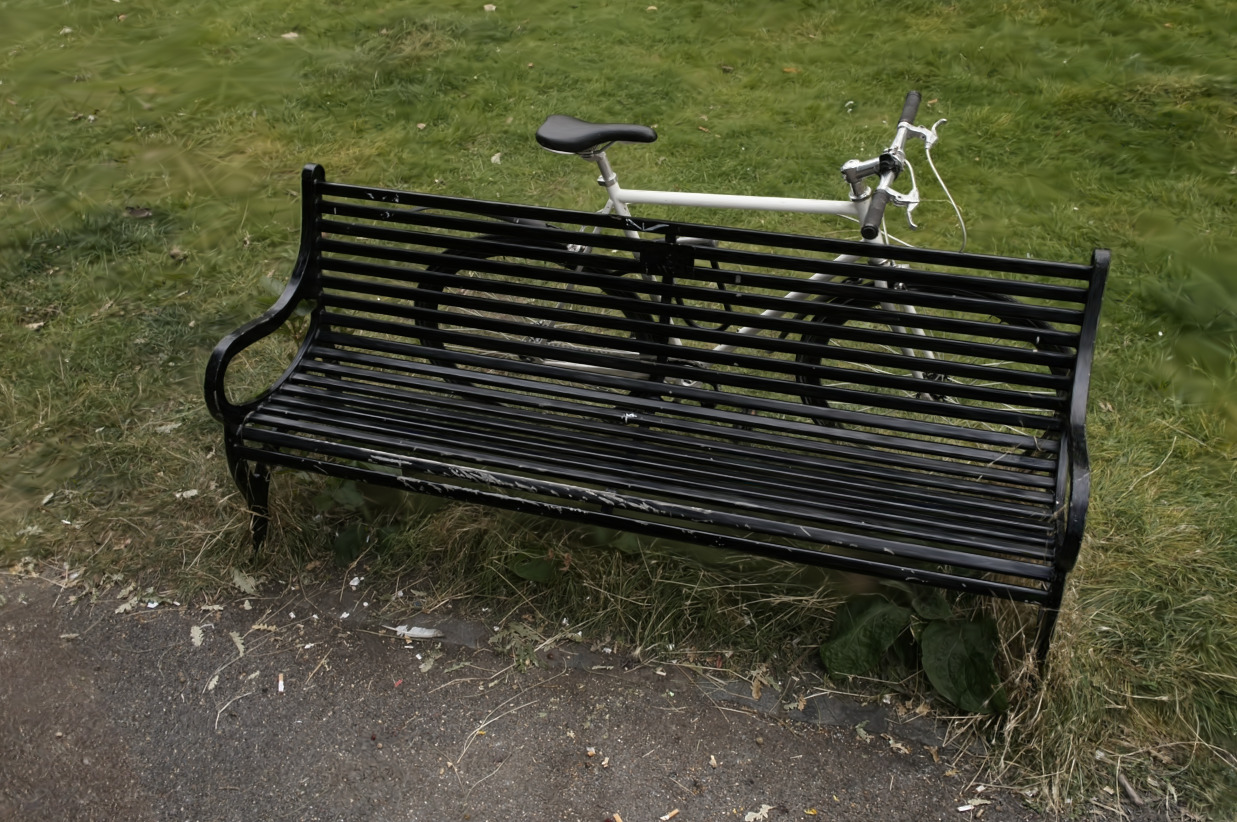}&
		\includegraphics[width=0.30\textwidth]{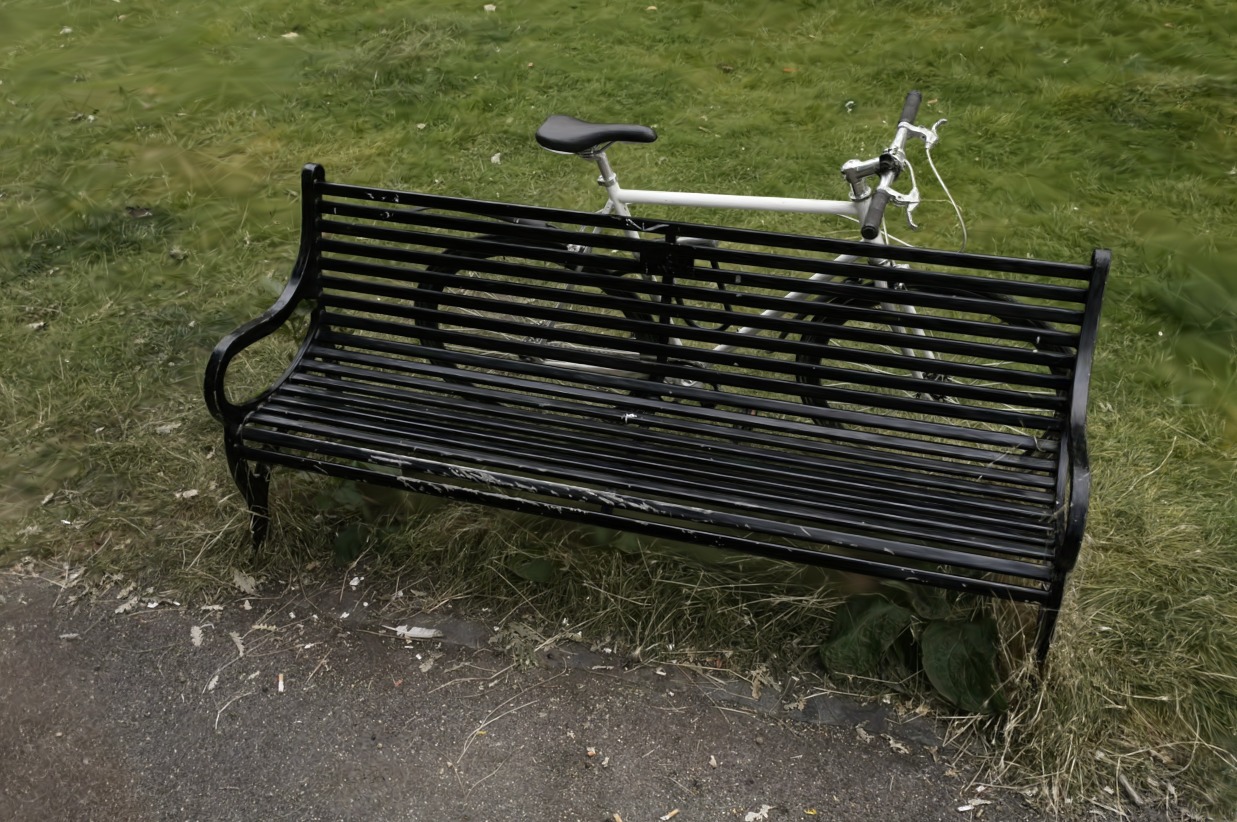}\\
        &Ground Truth & Baseline & Compressed
\end{tabular}
\caption{Random test views for each scene from Mip-NeRF360~\cite{barron_mip-nerf_2022}}
\label{fig:example-mip-2}
\end{figure*}

\begin{figure*}
\centering
\begin{tabular}{rccc}
		  \raisebox{2.8cm}[0pt][0pt]{\rotatebox[origin=c]{90}{Chair}} &
		\includegraphics[width=0.30\textwidth]{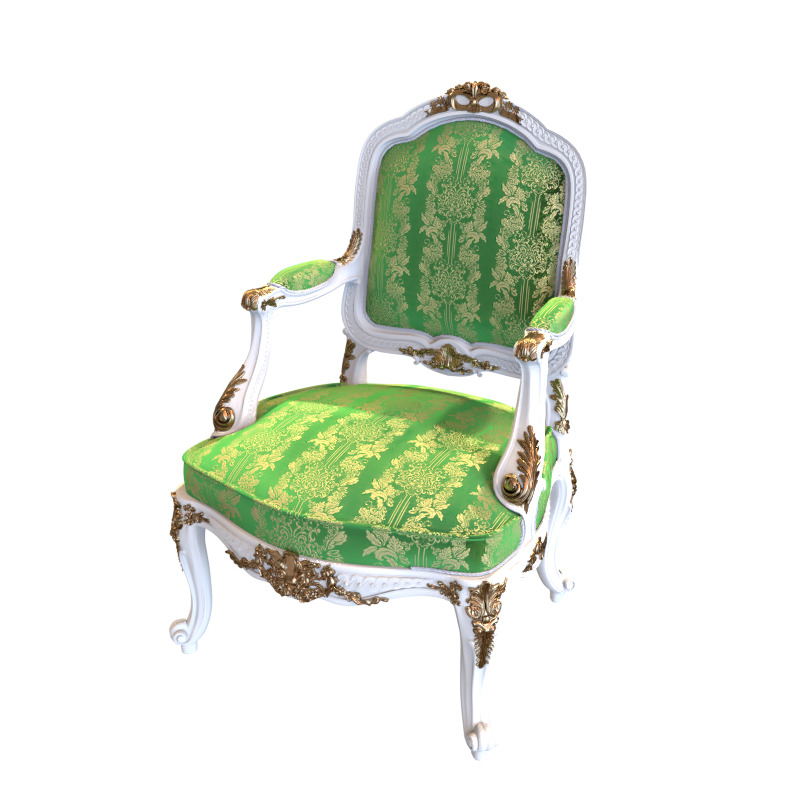}&
		\includegraphics[width=0.30\textwidth]{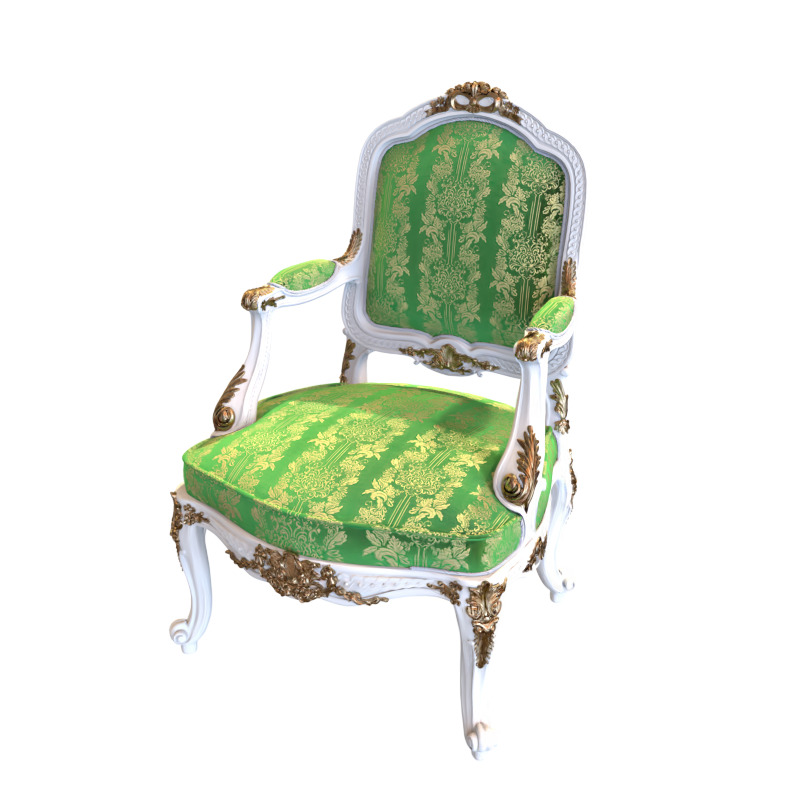}&
		\includegraphics[width=0.30\textwidth]{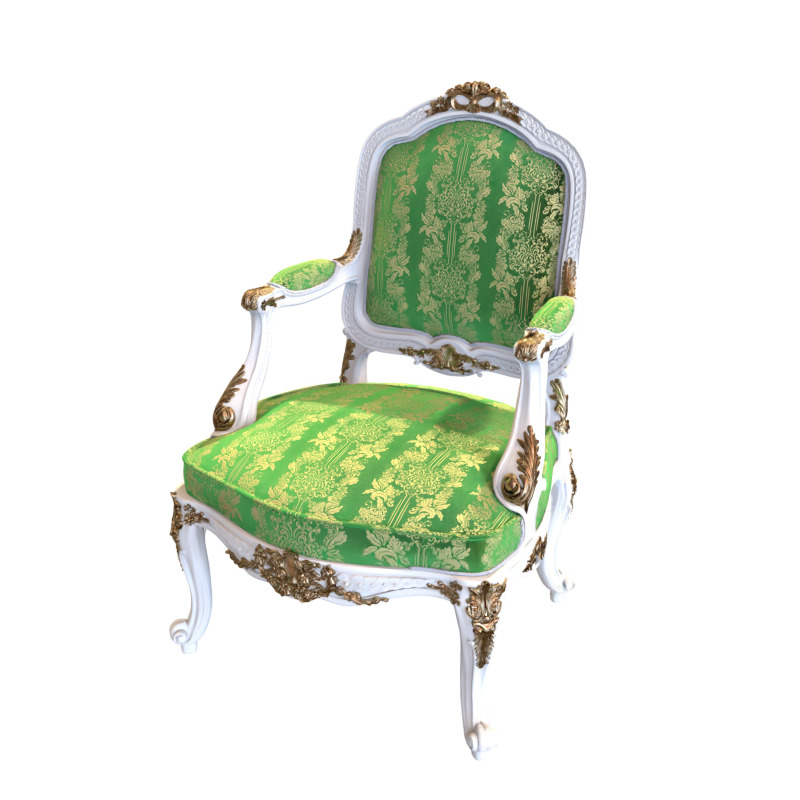}\\
		  \raisebox{2.8cm}[0pt][0pt]{\rotatebox[origin=c]{90}{Drums}} &
		\includegraphics[width=0.30\textwidth]{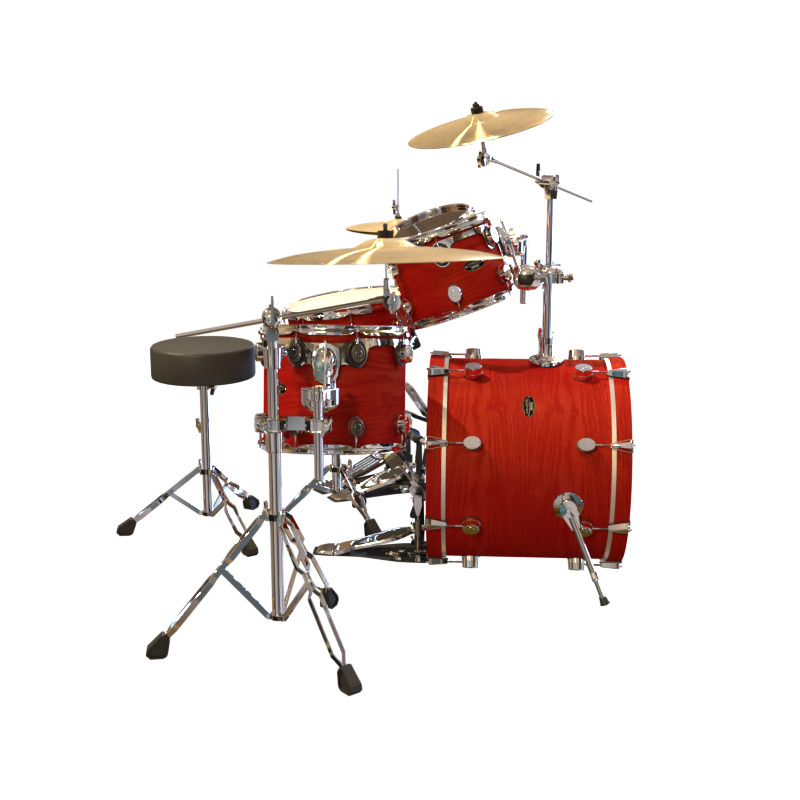}&
		\includegraphics[width=0.30\textwidth]{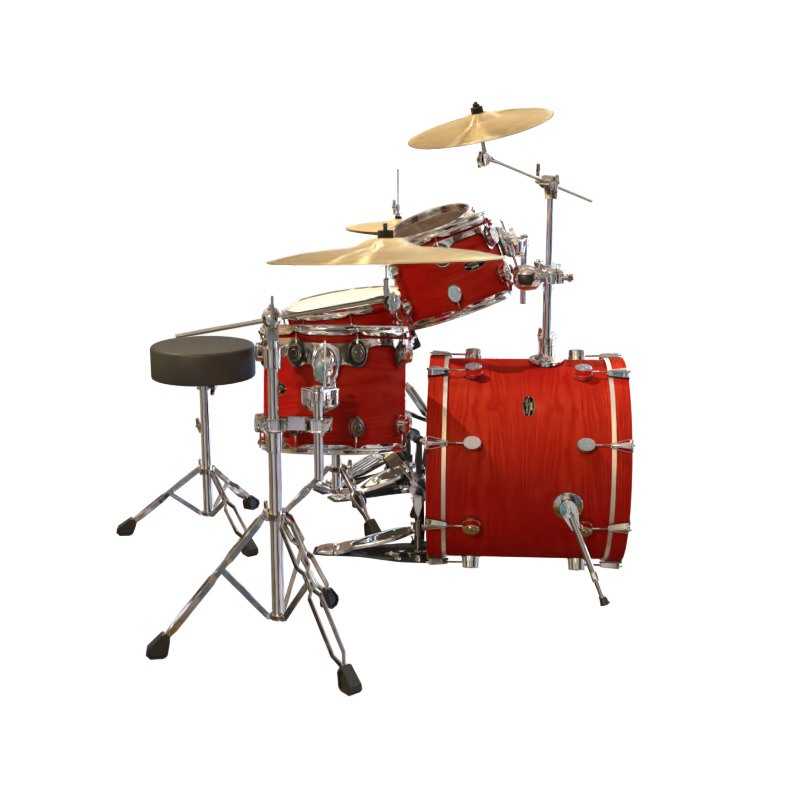}&
		\includegraphics[width=0.30\textwidth]{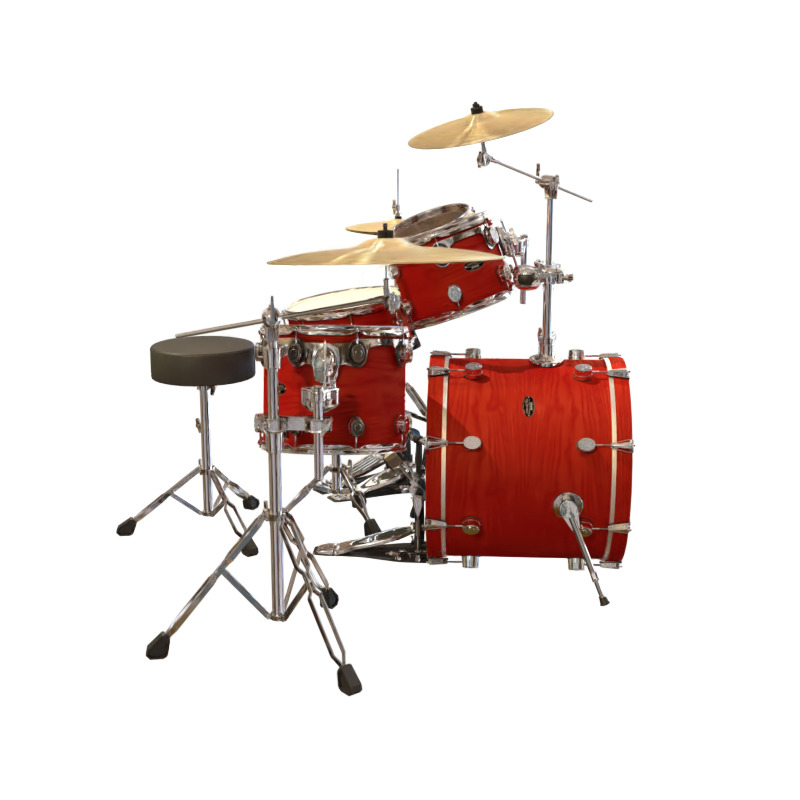}\\
		  \raisebox{2.8cm}[0pt][0pt]{\rotatebox[origin=c]{90}{Ficus}} &
		\includegraphics[width=0.30\textwidth]{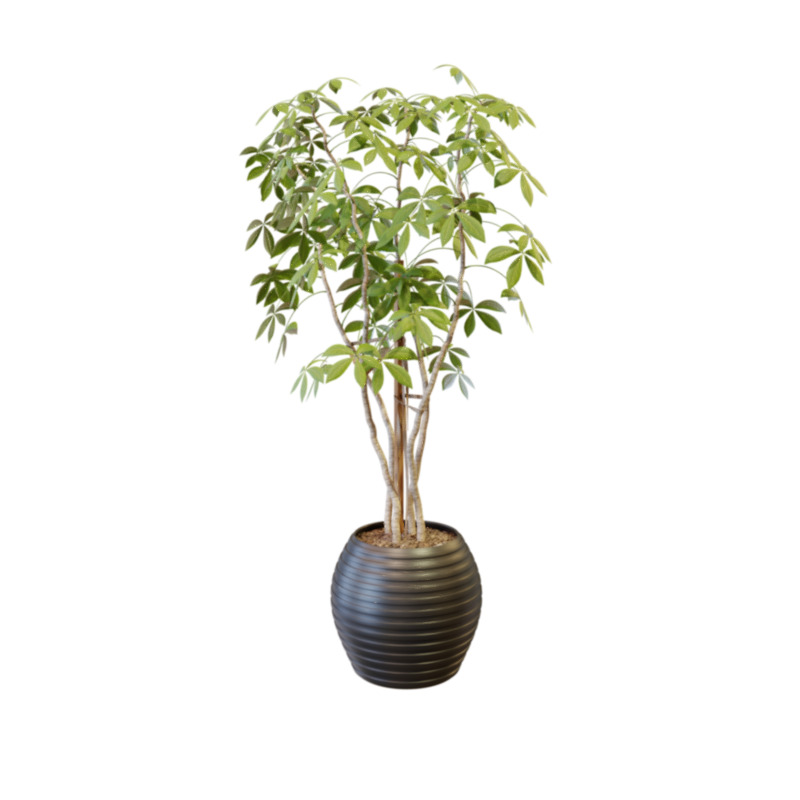}&
		\includegraphics[width=0.30\textwidth]{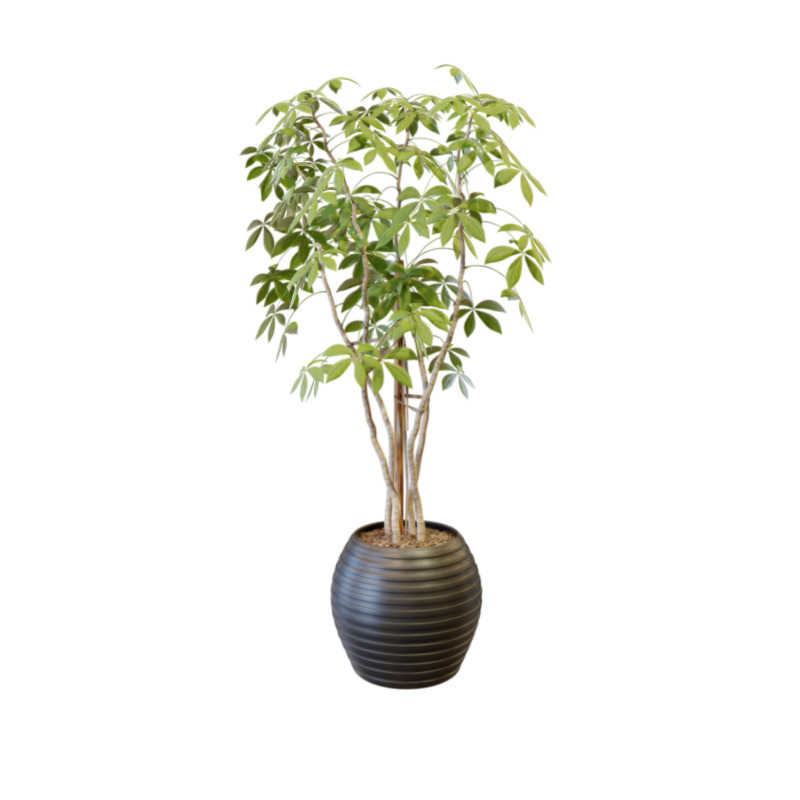}&
		\includegraphics[width=0.30\textwidth]{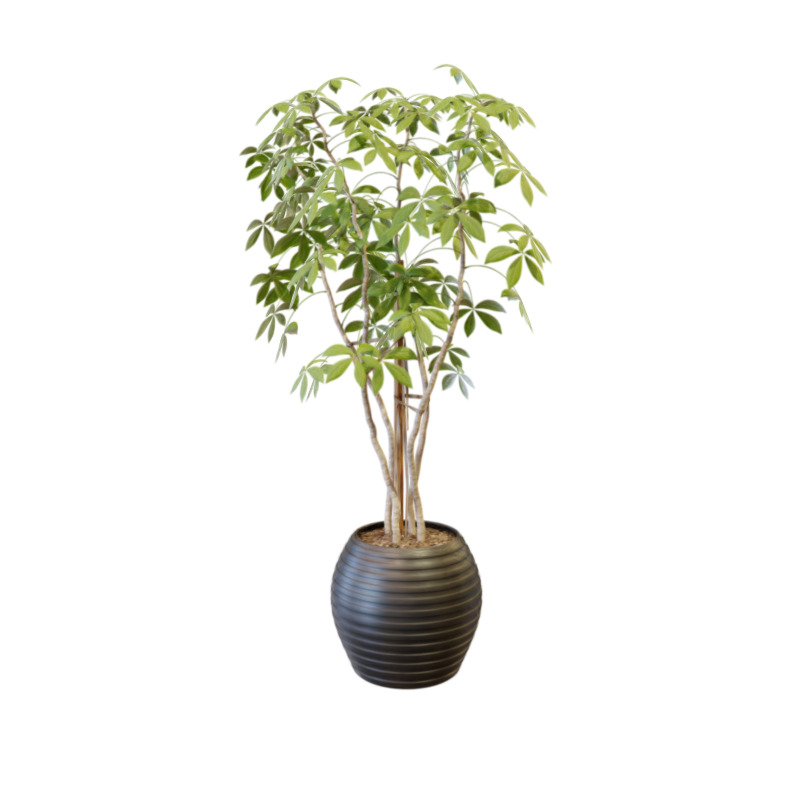}\\
		  \raisebox{2.8cm}[0pt][0pt]{\rotatebox[origin=c]{90}{Hotdog}} &
		\includegraphics[width=0.30\textwidth]{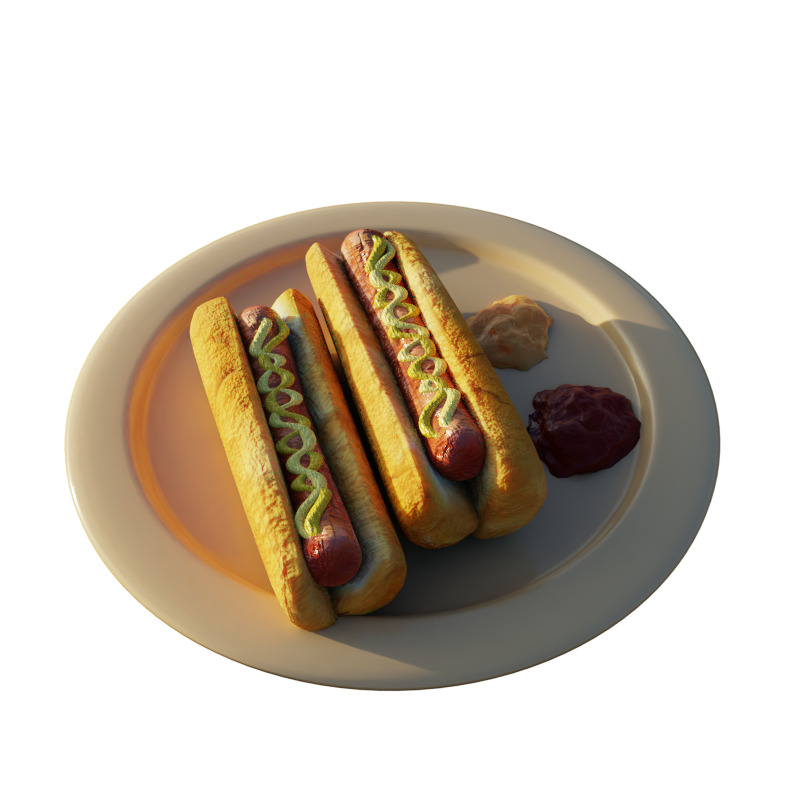}&
		\includegraphics[width=0.30\textwidth]{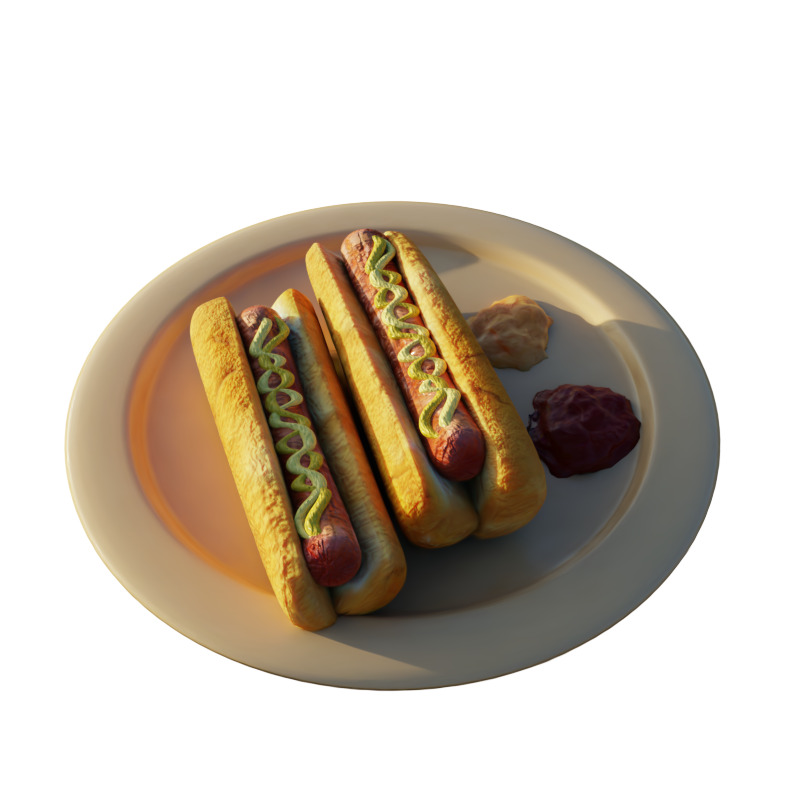}&
		\includegraphics[width=0.30\textwidth]{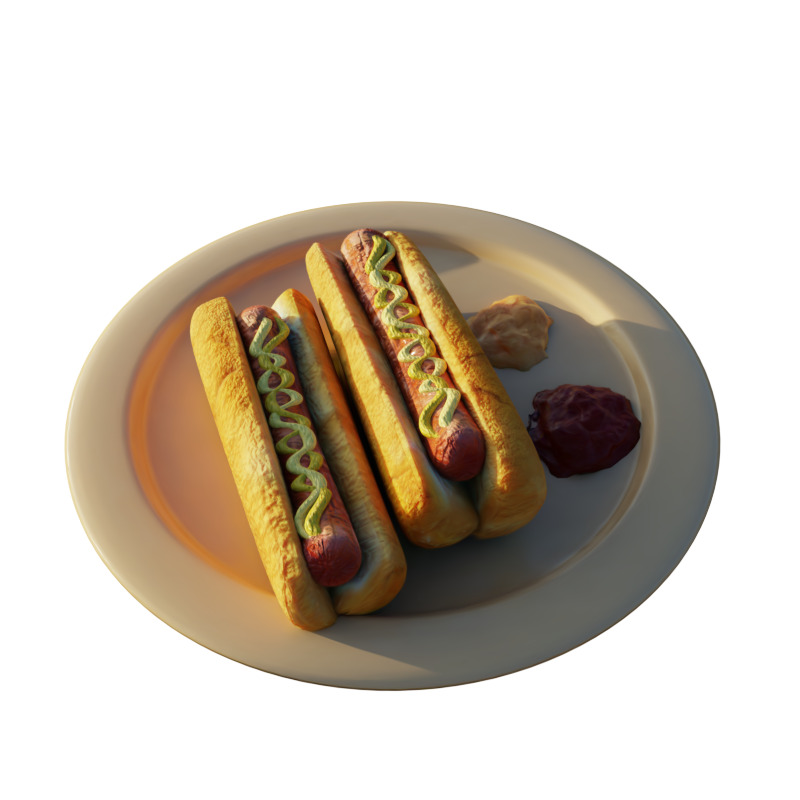}\\
        & Ground Truth & Baseline & Compressed
\end{tabular}
\caption{Random test views for each scene from NeRF Synthetic~\cite{mildenhall_nerf_2021}}
\label{fig:example-syn-1}
\end{figure*}

\begin{figure*}
\centering
    \begin{tabular}{rccc}
		\raisebox{2.8cm}[0pt][0pt]{\rotatebox[origin=c]{90}{Lego}} &
		\includegraphics[width=0.30\textwidth]{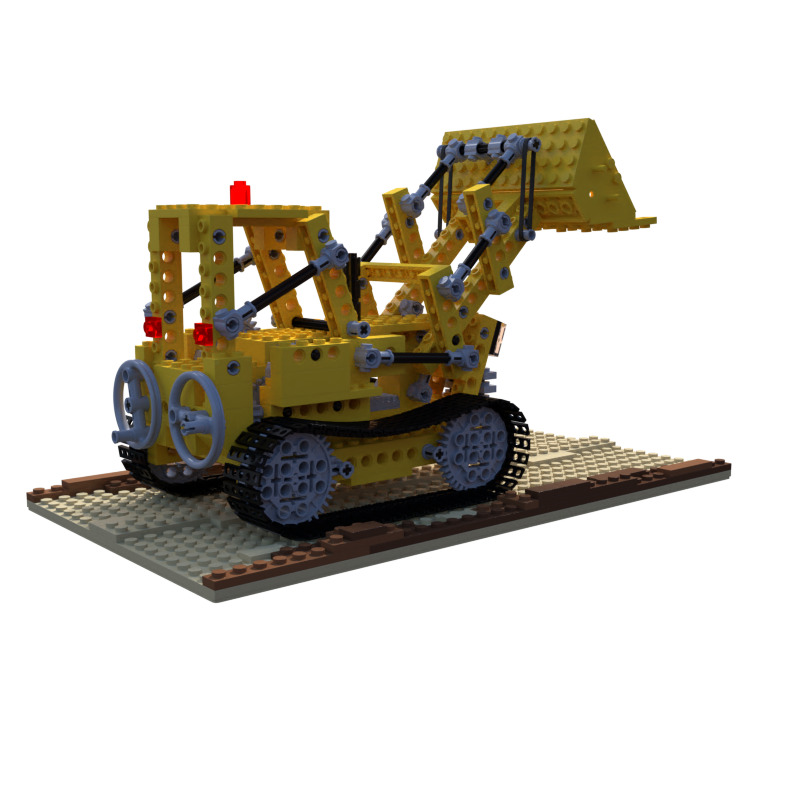}&
		\includegraphics[width=0.30\textwidth]{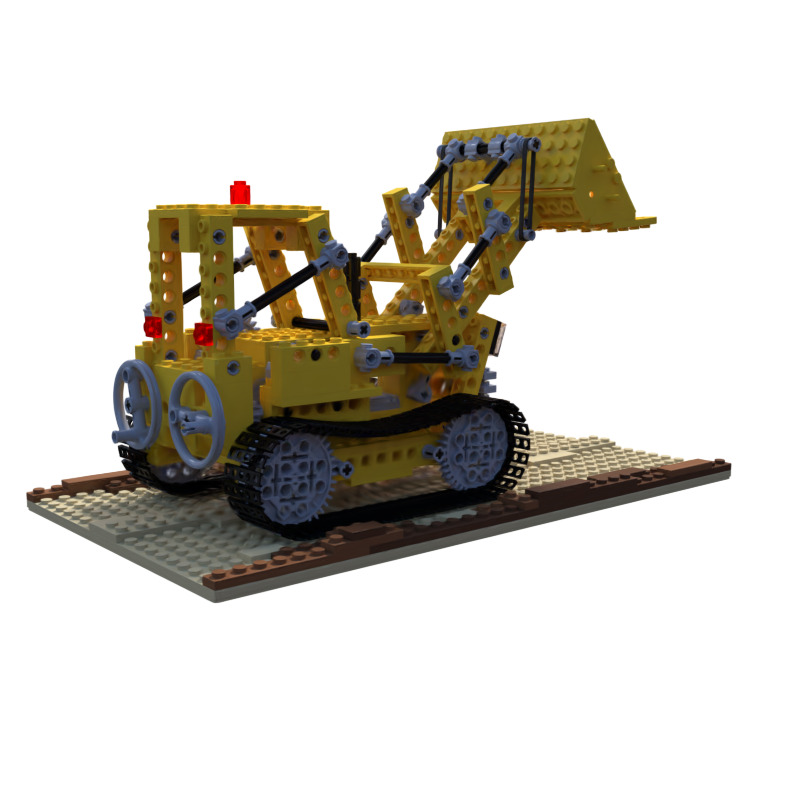}&
		\includegraphics[width=0.30\textwidth]{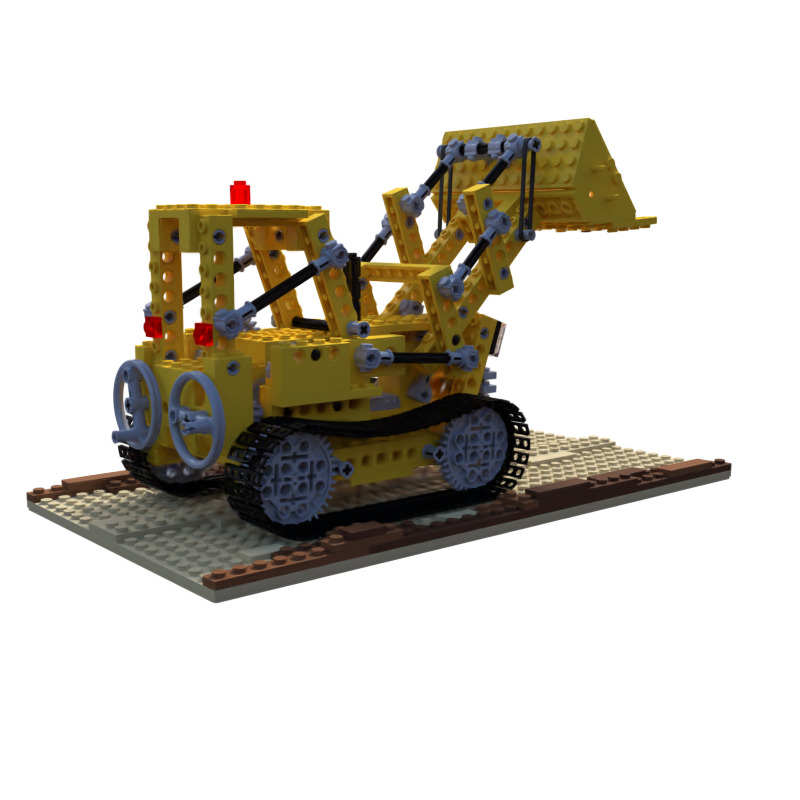}\\
		\raisebox{2.8cm}[0pt][0pt]{\rotatebox[origin=c]{90}{Materials}} &
		\includegraphics[width=0.30\textwidth]{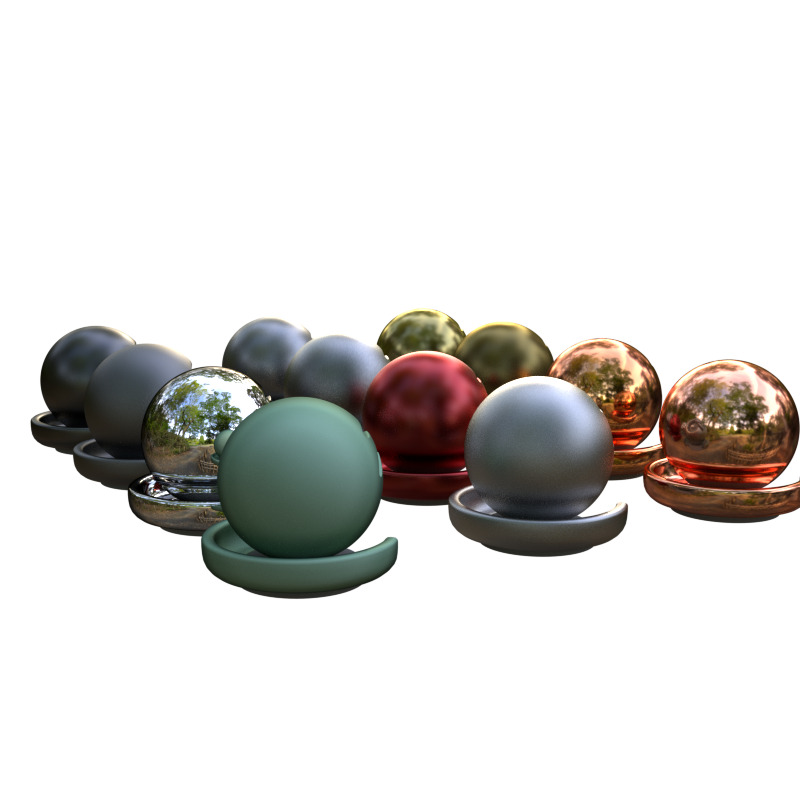}&
		\includegraphics[width=0.30\textwidth]{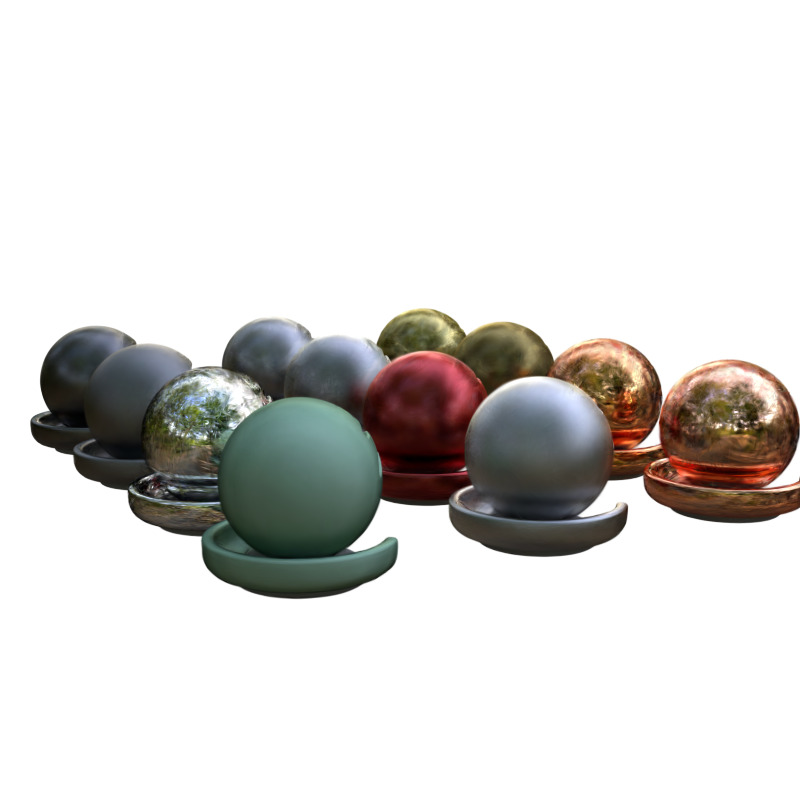}&
		\includegraphics[width=0.30\textwidth]{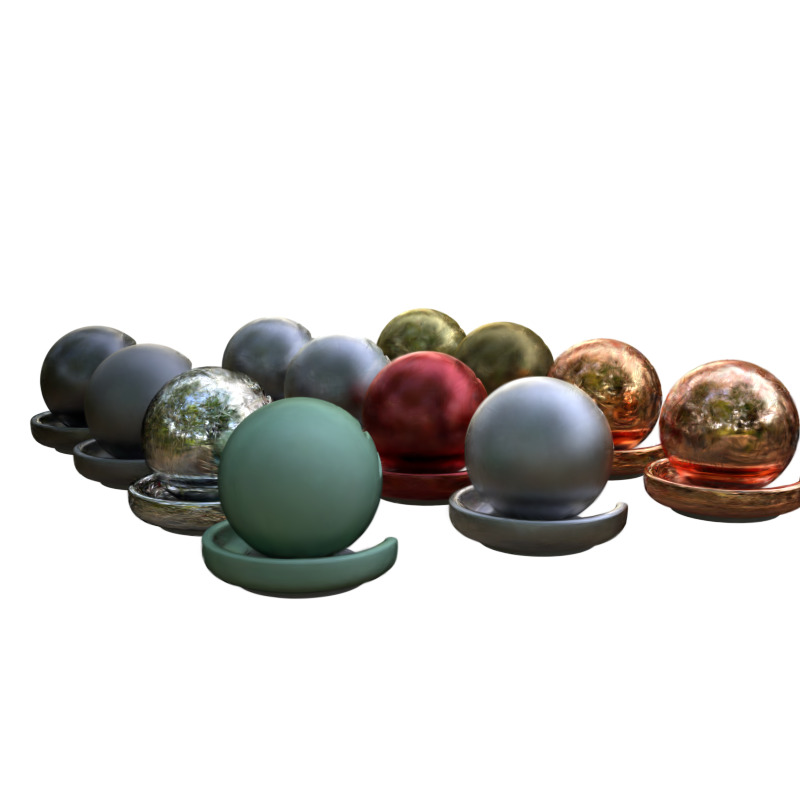}\\
		\raisebox{2.8cm}[0pt][0pt]{\rotatebox[origin=c]{90}{Mic}} &
		\includegraphics[width=0.30\textwidth]{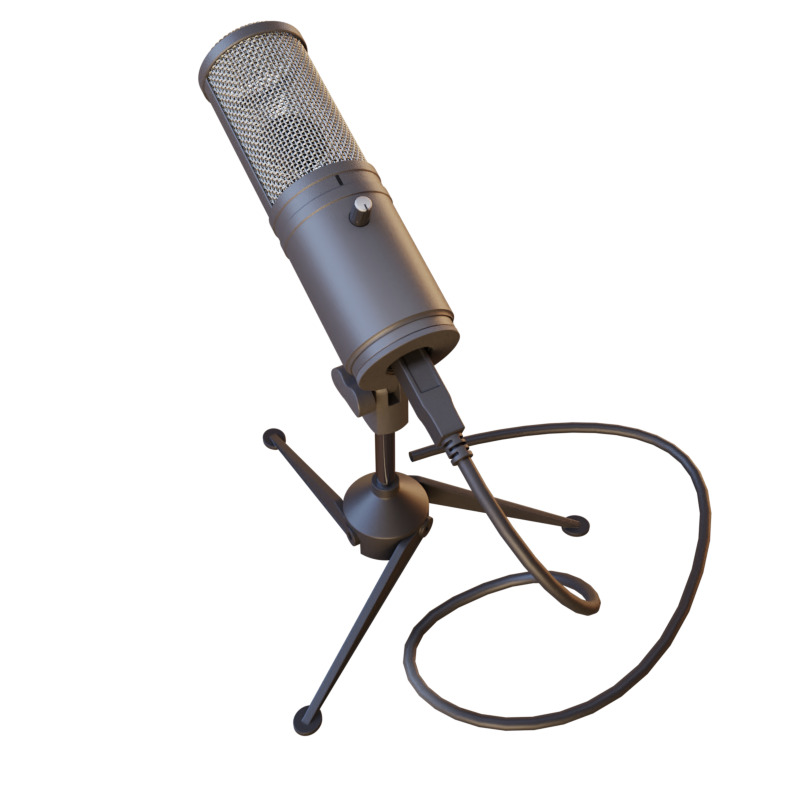}&
		\includegraphics[width=0.30\textwidth]{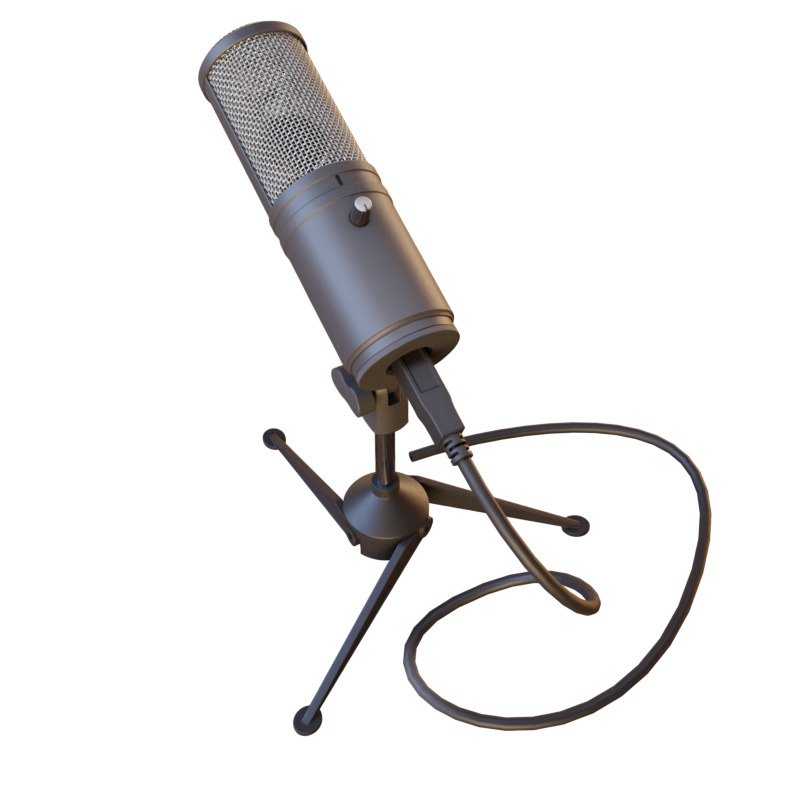}&
		\includegraphics[width=0.30\textwidth]{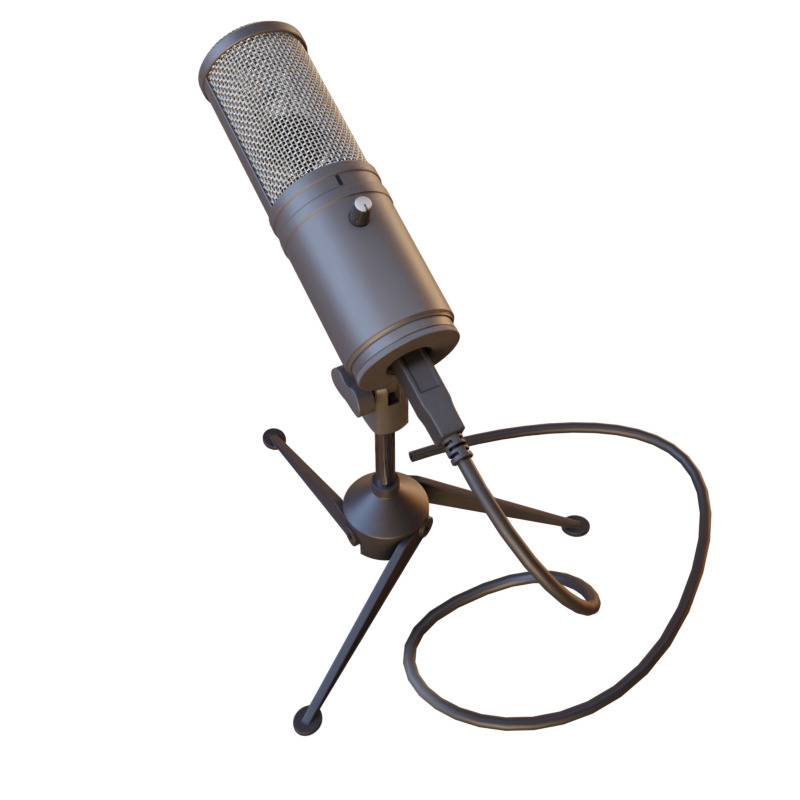}\\
		\raisebox{2.8cm}[0pt][0pt]{\rotatebox[origin=c]{90}{Ship}} &
		\includegraphics[width=0.30\textwidth]{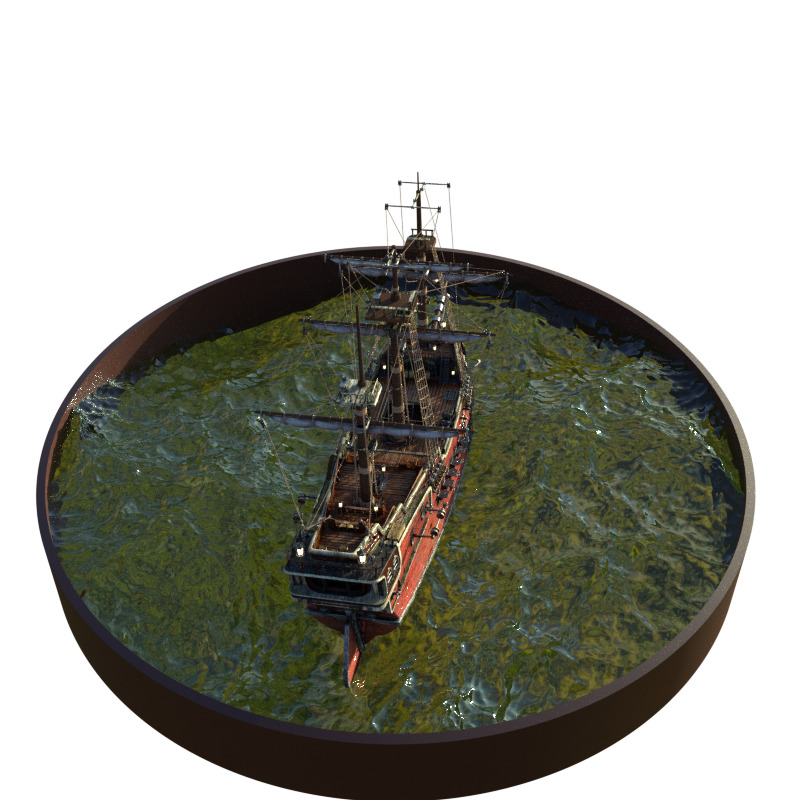}&
		\includegraphics[width=0.30\textwidth]{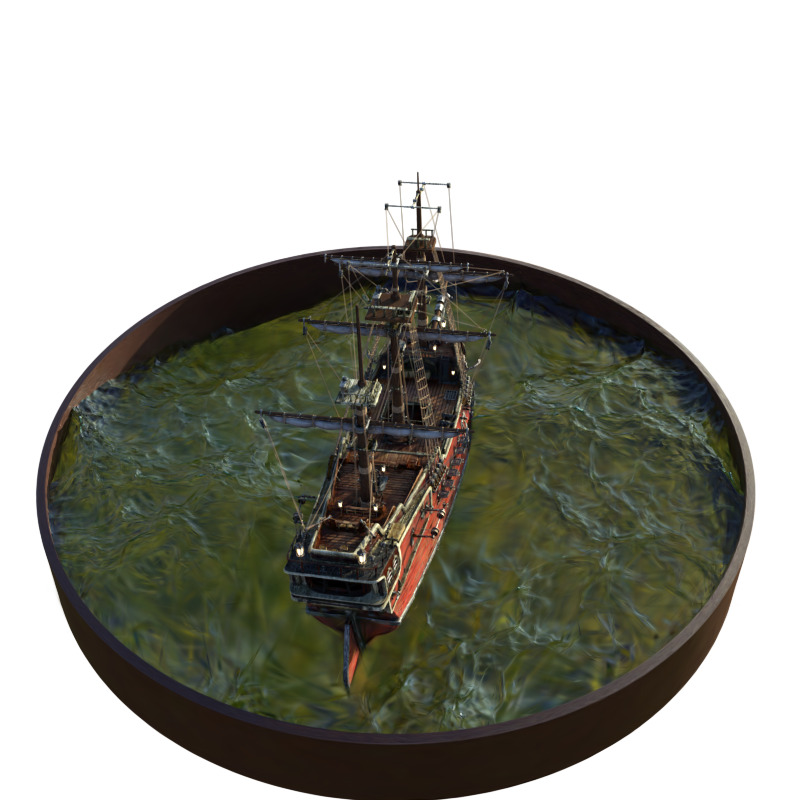}&
		\includegraphics[width=0.30\textwidth]{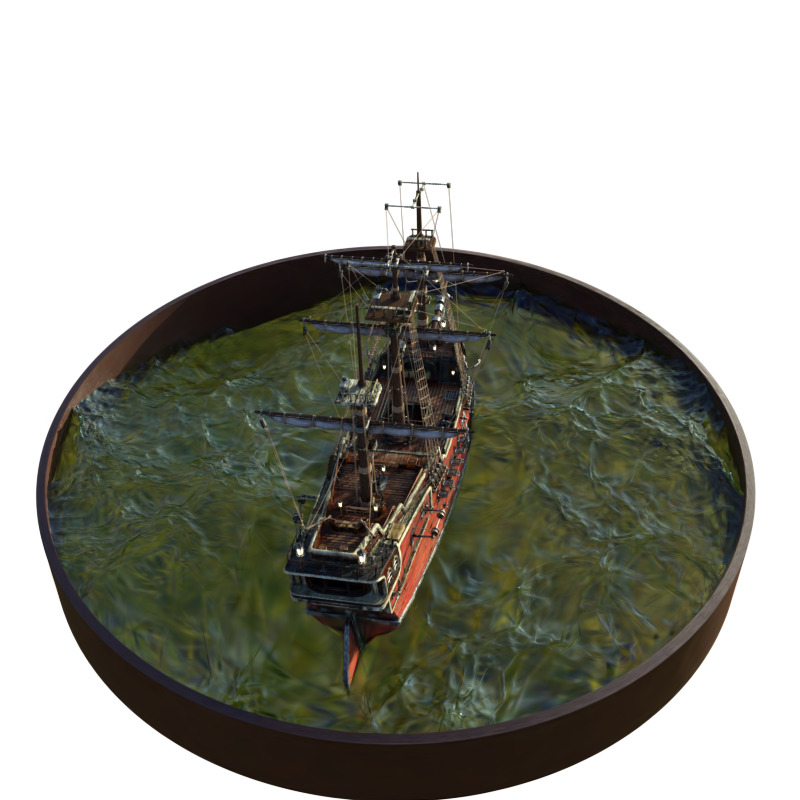}\\
        & Ground Truth & Baseline & Compressed
    \end{tabular}
\caption{Random test views for each scene from NeRF Synthetic~\cite{mildenhall_nerf_2021}}
\label{fig:example-syn-2}
\end{figure*}

For all scenes used in the paper, a random test view is selected. The ground truth images are compared to the renderings of the uncompressed (baseline) and our compressed (Compressed) scene representation. 
See \cref{fig:example-mip-1,fig:example-mip-2} for Mip-Nerf360 ~\cite{barron_mip-nerf_2022},
\cref{fig:example-deep} for Deep Blending ~\cite{hedman_deep_2018}, \cref{fig:example-tnt} for Tanks \& Temples ~\cite{knapitsch_tanks_2017}, \cref{fig:example-syn-1,fig:example-syn-2} for the synthetic scenes from ~\cite{mildenhall_nerf_2021}.

\subsection{Memory Requirements}

\begin{figure}[h]
    \centering
    \includegraphics[width=\linewidth]{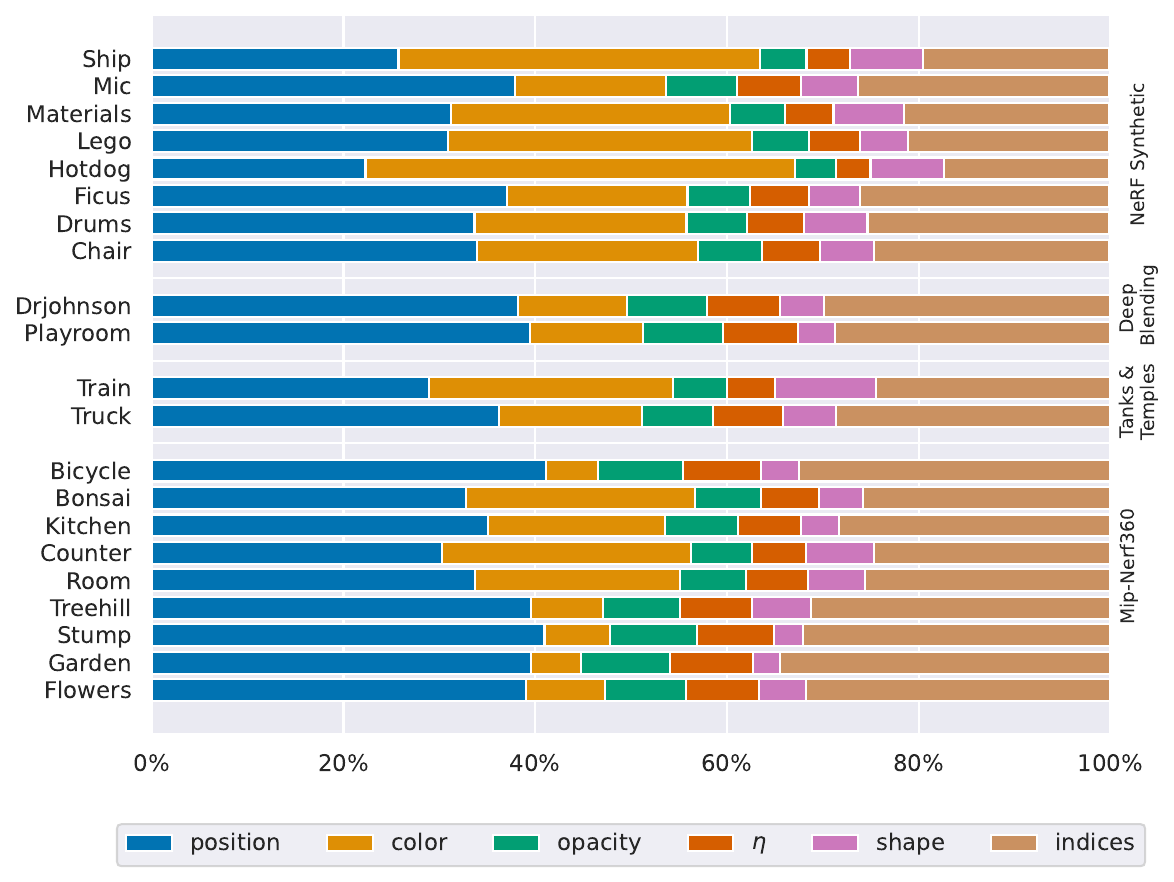}
    \caption{Storage size of different scene parameters in the compressed representation. Color is the codebook with all SH coefficients. Shape is the codebook with the Gaussian parameters and $\eta$ is the scaling factor. }
    \label{fig:file-composition}
\end{figure}

\cref{fig:file-composition} illustrates the memory requirements of different scene parameters.
The coordinates of the 3D Gaussian center points and the codebook indices take up the most memory in general.
The amount of memory required by the color codebook varies significantly between different scenes. 

\subsection{Timing Statistics}

We provide timings for the different stages of our compression pipeline.
\cref{tab:timing} shows the average and maximum time required by each stage.
It can be seen that the fine-tuning stage takes up $70\%$ of the total time.

\begin{table}[h]
\resizebox{\linewidth}{!}{
\begin{tabular}{l|SS}
\hline
                        & {Average Time  $\downarrow$} & {Maximum Time $\downarrow$}\\ \hline
Sensitivity Calculation                                                   &   8.05      &  11.38      \\
Clustering                                                                &   75.11     &  78.41       \\
QA Fine-tuning                                                            &   213.30     &  278.05       \\
Encoding                                                                  &   2.69      &  5.13      \\ \hline
Total                                                                     &   299.15     &  365.94      \\ \hline
\end{tabular}
}
\caption{Time requirements of the individual stages of the compression pipeline. We report the average and maximum time of each stage in seconds. The entropy and run-length encoding are grouped into the Encoding stage. Measurements were taken with an NVIDIA RTX A5000 graphics card.}
\label{tab:timing}
\end{table}

Additionally, we report timings for each stage of the novel view renderer. 
\cref{tab:render-time} shows the average times for two different scenes.
It can be seen that the preprocessing stage is accelerated by a factor of $5\times$ when using the compressed scene representation.

\begin{table}[h]
\resizebox{\linewidth}{!}{%
\begin{tabular}{ll|SSS|S}
\hline
                         &              & {Preprocess $\downarrow$} & {Sorting $\downarrow$} & {Rasterization $\downarrow$} & {Total $\downarrow$} \\ \hline
\vbox{\hbox{\multirow{3}{*}{\rotatebox[origin=c]{90}{Bicycle}}}}& Uncompressed & 1.46       & 0.55    & 2.81          & 4.82  \\[2.5ex]
                         & Compressed   & 0.28       & 0.48    & 2.45          & 3.22  \\ \hline
\vbox{\hbox{\multirow{3}{*}{\rotatebox[origin=c]{90}{Bonsai}}}}  & Uncompressed & 0.44       & 0.20    & 1.81          & 2.44  \\[2.5ex]
                         & Compressed   & 0.09       & 0.19    & 1.67          & 1.95   \\\hline
\end{tabular}%
}
\caption{Timings in milliseconds for the different stages of our renderer. Evaluated on an NVIDIA RTX A5000 with scenes from Mip-Nerf360~\cite{barron_mip-nerf_2022}}
\label{tab:render-time}
\end{table}

\newpage

\subsection{Sensitivity Calculation and Pruning}

The sensitivity of a parameter is calculated using the gradient of the total image energy wrt. this parameter (see \cref{eq:sensitivity}).
Kerbl~\etal~\cite{kerbl_3d_2023} clamp negative direction-dependent colors (i.e., resulting from the evaluation of the SH coefficients) to zero. 
For the clamped values, the partial derivatives are set to zero in the backward pass. 
This results in a sensitivity of zero for the respective SH coefficients, which is not desired since they possibly contribute to the training images. 
Therefore, we do not clamp colors when calculating the sensitivity.

We observe that a notable number of Gaussians (up to $15\%$) do not have any impact on the training images.
These particular splats exhibit zero sensitivity in the color parameters. 
Consequently, we opt to eliminate these splats from the scene (called Pruning in \cref{tab:ablation}).

\begin{figure}
    \centering
    \begin{subfigure}{0.45\linewidth}
        \includegraphics[width=\linewidth]{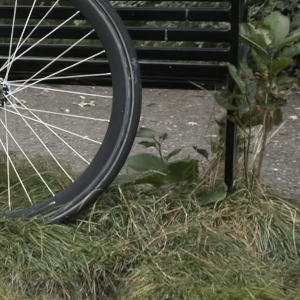}
        \caption{Baseline}
    \end{subfigure}
    \hfill
    \begin{subfigure}{0.45\linewidth}
        \includegraphics[width=\linewidth]{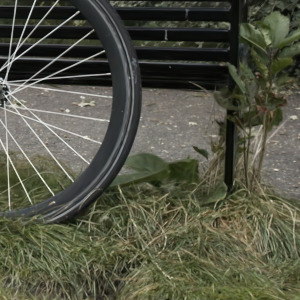}
        \caption{Compressed}
    \end{subfigure}
    \caption{Pruning failure case. Compared to the baseline reconstruction, some leaves have been removed in the compressed version due to pruning. }
    \label{fig:pruning-fail}
\end{figure}

Experiments with higher pruning thresholds have shown that more Gaussians can be removed with minimal loss in PSNR. 
However, this can lead to fine details in the scene being removed, which we consider undesirable.
An example of this can be seen in \cref{fig:pruning-fail}, where small leaves were removed from the reconstruction due to pruning.

\subsection{Covariance Matrix Clustering}

\label{sec:cov-update}

Given a rotation matrix $R\in\mathbb{R}^{3\times3}$ and a scaling vector $\mathbf{s}\in\mathbb{R}^3_{>0}$. The covariance matrix $\Sigma$ is defined as~\cite{kerbl_3d_2023}

\begin{align}
\Sigma = RSSR^T = RS^2R^T,
\end{align}
with $S = \mathrm{diag}(\mathbf{s})$ .
Since $\Sigma$ is real and symmetric it holds that  
\begin{align}
S^2=\mathrm{diag}([\lambda_1,\lambda_2,\lambda_3]^T) = \mathrm{diag}([s_1^2,s_2^2,s_3^2]^T),
\end{align}
where $\lambda_i$ are the eigenvalues of $\Sigma$.
By using the trace of $\Sigma$, the squared length of $\mathbf{s}$ can be calculated as
\begin{align}
\label{eq:trace}
    \mathrm{Tr}(\Sigma) = \sum_{i=1}^3 \lambda_i = \sum_{i=1}^3 s_i^2 = \lVert \mathbf{s}\rVert_2^2
\end{align}

\paragraph{Clustering Update Step}
In the following, we show that the clustering update step results in normalized covariance matrices as cluster centroids.
Given $N$ normalized covariance matrices $\hat{\Sigma}_i$ with $\lVert \mathbf{s_i}\rVert_2=1$ and respective weighting factors $w_i\in\mathbb{R}_{>0}$. Their centroid $\hat{\Sigma}_c$ is calculated as 
\begin{align}
    \hat{\Sigma}_c = \frac{1}{\sum_{i=1}^N w_i} \sum_{i=1}^N w_i \hat{\Sigma}_i
\end{align}.
By using \cref{eq:trace} it holds that

\begin{align}
    \lVert \mathbf{s_c}\rVert_2^2 &= \mathrm{Tr}(\hat{\Sigma}_c) \\
    &= \mathrm{Tr}(\frac{1}{\sum_{i=1}^N w_i} \sum_{i=1}^N w_i \hat{\Sigma}_i) \\
    &= \frac{1}{\sum_{i=1}^N w_i} \sum_{i=1}^N w_i \mathrm{Tr}(\hat{\Sigma}_i) \\
    &= \frac{1}{\sum_{i=1}^N w_i} \sum_{i=1}^N w_i \lVert \mathbf{s_i}\rVert_2^2 \\
    &= 1
\end{align}
This proves that the covariance matrix $\hat{\Sigma}_c$ has a normalized scaling vector and thus iteself is in a normalized form.

\paragraph{Covariance Matrix Normalization}
The following derivation proofs that a covariance matrix $\Sigma$ can be transformed into its normalized form $\hat{\Sigma}$ by dividing it by its trace, i.e.,  

\begin{align}
    \frac{\Sigma}{\mathrm{Tr}(\Sigma)} &= R\frac{S^2}{\mathrm{Tr}(\Sigma)}R^T = R \frac{S}{\lVert \mathbf{s}\rVert_2}\frac{S}{\lVert \mathbf{s}\rVert_2} R^T \\
    &= R \hat{S}^2 R^T = \hat{\Sigma}
\end{align}

\begin{table*}[]
\begin{tabular}{r|SSSS|SSSS|S}
     & \multicolumn{4}{c|}{3D Gaussian Splatting}  & \multicolumn{4}{c|}{Ours}& \\ \hline
Scene    & {PSNR $\uparrow$} & {SSIM $\uparrow$} & {LPIPS $\downarrow$} & {SIZE $\downarrow$} & {PSNR $\uparrow$} & {SSIM $\uparrow$} & {LPIPS $\downarrow$} & {SIZE $\downarrow$} & { Ratio $\uparrow$} \\ \hline
bicycle  &  25.171 & 0.762 & 0.216 & 1450.277 & 24.970 & 0.751 & 0.240 & 47.147 & 30.761 \\
bonsai  &  31.979 & 0.938 & 0.208 & 294.415 & 31.347 & 0.930 & 0.217 & 12.794 & 23.011 \\
counter  &  28.888 & 0.905 & 0.204 & 289.244 & 28.671 & 0.896 & 0.215 & 13.789 & 20.977 \\
flowers  &  21.448 & 0.602 & 0.341 & 860.062 & 21.152 & 0.584 & 0.358 & 31.140 & 27.619 \\
garden  &  27.179 & 0.861 & 0.115 & 1379.993 & 26.746 & 0.844 & 0.144 & 46.565 & 29.636 \\
kitchen  &  30.713 & 0.923 & 0.130 & 438.099 & 30.262 & 0.914 & 0.140 & 18.874 & 23.211 \\
room  &  31.341 & 0.916 & 0.223 & 376.853 & 31.138 & 0.911 & 0.231 & 15.033 & 25.068 \\
stump  &  26.562 & 0.770 & 0.219 & 1173.522 & 26.285 & 0.757 & 0.250 & 40.569 & 28.926 \\
treehill  &  22.303 & 0.631 & 0.328 & 894.903 & 22.256 & 0.620 & 0.351 & 33.318 & 26.859 \\ \hline
average  &  27.287 & 0.812 & 0.220 & 795.263 & 26.981 & 0.801 & 0.238 & 28.803 & 26.230 \\ \hline
\end{tabular}
\caption{Mip-Nerf360~\cite{barron_mip-nerf_2022} results.}
\label{tab:results-mip}
\end{table*}

\begin{table*}[]
\begin{tabular}{r|SSSS|SSSS|S}
     & \multicolumn{4}{c|}{3D Gaussian Splatting}  & \multicolumn{4}{c|}{Ours}& \\ \hline
Scene    & {PSNR $\uparrow$} & {SSIM $\uparrow$} & {LPIPS $\downarrow$} & {SIZE $\downarrow$} & {PSNR $\uparrow$} & {SSIM $\uparrow$} & {LPIPS $\downarrow$} & {SIZE $\downarrow$} & { Ratio $\uparrow$} \\ \hline
train  &  21.770 & 0.805 & 0.217 & 242.782 & 21.863 & 0.798 & 0.226 & 13.249 & 18.324 \\
truck  &  24.940 & 0.871 & 0.155 & 601.030 & 24.823 & 0.867 & 0.161 & 21.316 & 28.196 \\ \hline
average  &  23.355 & 0.838 & 0.186 & 421.906 & 23.343 & 0.832 & 0.194 & 17.282 & 23.260 \\ \hline
\end{tabular}
\caption{Tanks\&Temples~\cite{knapitsch_tanks_2017} results}
\label{tab:results-tnt}
\end{table*}

\begin{table*}[]
\begin{tabular}{r|SSSS|SSSS|S}
     & \multicolumn{4}{c|}{3D Gaussian Splatting}  & \multicolumn{4}{c|}{Ours}& \\ \hline
Scene    & {PSNR $\uparrow$} & {SSIM $\uparrow$} & {LPIPS $\downarrow$} & {SIZE $\downarrow$} & {PSNR $\uparrow$} & {SSIM $\uparrow$} & {LPIPS $\downarrow$} & {SIZE $\downarrow$} & { Ratio $\uparrow$} \\ \hline
drjohnson                  & 28.938                   & 0.896                    & 0.248                     & 805.358                        & 28.871                   & 0.895                    & 0.254                     & 28.938                         & 27.830                                                                          \\
playroom                   & 29.926                   & 0.901                    & 0.244                     & 602.186                        & 29.891                   & 0.900                    & 0.252                     & 21.660                         & 27.802                                                                          \\ \hline
average                    & 29.432                   & 0.898                    & 0.246                     & 703.772                        & 29.381                   & 0.898                    & 0.253                     & 25.299                         & 27.816                                                                          \\ \hline
\end{tabular}
\caption{Deep Blending~\cite{hedman_deep_2018} results}
\label{tab:results-deep}
\end{table*}

\begin{table*}[]
\begin{tabular}{r|SSSS|SSSS|S}
     & \multicolumn{4}{c|}{3D Gaussian Splatting}  & \multicolumn{4}{c|}{Ours}& \\ \hline
Scene    & {PSNR $\uparrow$} & {SSIM $\uparrow$} & {LPIPS $\downarrow$} & {SIZE $\downarrow$} & {PSNR $\uparrow$} & {SSIM $\uparrow$} & {LPIPS $\downarrow$} & {SIZE $\downarrow$} & { Ratio $\uparrow$} \\ \hline
chair                      & 35.864                   & 0.987                    & 0.012                     & 70.105                         & 35.297                   & 0.985                    & 0.014                     & 3.575                          & 19.609                                                                          \\
drums                      & 26.072                   & 0.954                    & 0.038                     & 83.665                         & 25.941                   & 0.952                    & 0.040                     & 3.829                          & 21.848                                                                          \\
ficus                      & 34.736                   & 0.987                    & 0.012                     & 70.177                         & 34.559                   & 0.986                    & 0.013                     & 3.059                          & 22.937                                                                          \\
hotdog                     & 37.646                   & 0.985                    & 0.021                     & 34.079                         & 37.367                   & 0.984                    & 0.022                     & 2.725                          & 12.505                                                                          \\
lego                       & 35.399                   & 0.981                    & 0.017                     & 76.071                         & 34.802                   & 0.979                    & 0.020                     & 4.314                          & 17.633                                                                          \\
materials                  & 29.861                   & 0.959                    & 0.035                     & 71.833                         & 29.602                   & 0.957                    & 0.038                     & 4.021                          & 17.862                                                                          \\
mic                        & 35.155                   & 0.991                    & 0.006                     & 77.563                         & 34.913                   & 0.991                    & 0.007                     & 3.025                          & 25.640                                                                          \\
ship                       & 30.954                   & 0.905                    & 0.111                     & 75.659                         & 31.005                   & 0.905                    & 0.111                     & 4.938                          & 15.322                                                                          \\ \hline
average                    & 33.211                   & 0.969                    & 0.031                     & 69.894                         & 32.936                   & 0.967                    & 0.033                     & 3.686                          & 19.170                                                                          \\ \hline
\end{tabular}
\caption{NeRF Synthetic~\cite{mildenhall_nerf_2021} results}
\label{tab:results-syn}
\end{table*}

\graphicspath{{img/examples/tnt}}

\graphicspath{{img/examples/deep}}

\graphicspath{{img/examples/mip}}

\graphicspath{{img/examples/syn}}

\end{document}